%% file: main.tex
\documentclass{article} % For LaTeX2e
\usepackage[]{iclr2025_conference,times}
\usepackage{setspace}
\usepackage{graphicx}
\usepackage{subcaption}
\usepackage{placeins}
\usepackage{wrapfig}

\usepackage{hyperref}
\usepackage{multirow}
\usepackage{paralist}
\usepackage{url}
\usepackage{algorithm}
\usepackage[noend]{algpseudocode}
\usepackage{setspace}
\usepackage{color}
\usepackage{colortbl}
\usepackage[most]{tcolorbox}
\usepackage{soul}
\newtcolorbox{white}{colback=white!10!white,boxrule=0pt, top=0pt,bottom=0pt, left=0pt}
\newtcolorbox{blue}{colback=blue!10!white,boxrule=0pt, top=0pt,bottom=0pt, left=0pt}
\newtcolorbox{red}{colback=red!10!white,boxrule=0pt,top=0pt,bottom=0pt, left=0pt}
\newtcolorbox{green}{colback=green!10!white,boxrule=0pt, top=0pt,bottom=0pt, left=0pt}
\colorlet{bluec}{blue!20!white}
\colorlet{greenc}{green!20!white}
\colorlet{redc}{red!20!white}
\usepackage{xspace}
\usepackage{booktabs} % fancy tabs
\usepackage{amsmath}
\usepackage{amssymb}
\usepackage{enumitem} % for compact enumerate
\usepackage{pgfplots}
\pgfplotsset{compat=1.18} % for addressing a warning

\usepackage{csvsimple} % For reading CSV files
\usepackage{booktabs}  % For nicer tables (optional)
\usepackage{longtable} % For longer tables that may span multiple pages (optional)
\usepackage{siunitx}
\usepackage{nicefrac}

\newcommand{\scolorbox}[2]{{%
  \setlength{\fboxsep}{0.5\fboxsep}%
  \colorbox{#1}{#2}%
}}

\newcommand{\trim}{\texttt{TRIM}\xspace}

\newcommand{\glob}{\texttt{GLOB}\xspace}
\newcommand{\spec}{\texttt{SPEC}\xspace}

\newcommand{\dept}{\texttt{DEPT}\xspace}

\newcommand{\act}{\texttt{ACT}\xspace}
\usepackage{pgfplotstable} % Required for table processing
\usepackage{booktabs} % For better table formatting
\usepackage{colortbl} % For row coloring
\usepackage{array}    % For better column alignment
\usepackage{siunitx}  % For number formatting
\algdef{SE}[SUBALG]{Indent}{EndIndent}{}{\algorithmicend\ }%
\algtext*{Indent}
\algtext*{EndIndent}

\usepackage[capitalise]{cleveref}
\title{{\huge DEPT: Decoupled Embeddings for Pre-training Language Models}}

\makeatletter
\newcommand{\myfnsymbol}[1]{%
  \expandafter\@myfnsymbol\csname c@#1\endcsname
}
\newcommand{\@myfnsymbol}[1]{%
  \ifcase #1
  \or 1% 1
  \or 2% 2
  \or 3% 2
  \or \TextOrMath{\textasteriskcentered}{*}% 3
  \or \TextOrMath{\textasteriskcentered}{*}\TextOrMath{\textasteriskcentered}{*}% 4
  \or \TextOrMath{\textdagger}{\dagger}% 5
  \or \TextOrMath{\textasteriskcentered}{*},\TextOrMath{\textasteriskcentered}{*}\TextOrMath{\textasteriskcentered}{*}% 6
  \fi
}
\newcommand{\affiliationA}{\@myfnsymbol{1}}
\newcommand{\affiliationB}{\@myfnsymbol{2}}
\newcommand{\affiliationC}{\@myfnsymbol{3}}
\newcommand{\equalcontributor}{\@myfnsymbol{4}}
\newcommand{\biequalcontributor}{\@myfnsymbol{5}}
\newcommand{\correspondingA}{\@myfnsymbol{6}}

\author{
Alex Iacob\textsuperscript{\correspondingA,\affiliationA,\affiliationB, \equalcontributor}
\And  
Lorenzo Sani\textsuperscript{\affiliationA,\affiliationB,\equalcontributor}
\And Meghdad Kurmanji\textsuperscript{\affiliationA}
\And 
William F. Shen\textsuperscript{\affiliationA,\biequalcontributor}
\And  
Xinchi Qiu\textsuperscript{\affiliationA,\biequalcontributor}
\And  
Dongqi Cai\textsuperscript{\affiliationA, \affiliationC, \biequalcontributor}
\And  
Yan Gao\textsuperscript{\affiliationA,\affiliationB, \biequalcontributor}
\And 
Nicholas D. Lane\textsuperscript{\affiliationA,\affiliationB, \biequalcontributor}
}

\iclrfinalcopy % Uncomment for camera-ready version, but NOT for submission.
\begin{document}

% {
% \begingroup
% \begin{figure}[t]
%     \quad
%     \begin{subfigure}{0.1275\textwidth}
%         \includegraphics[width=\textwidth]{plots/flower_logo.png}
%     \end{subfigure}
%     \hfill
%     \begin{subfigure}{0.1\textwidth}
%         \includegraphics[width=\textwidth]{plots/cambridge_logo.jpg}
%     \end{subfigure}
%     \hfill
%     \begin{subfigure}{0.1275\textwidth}
%         \includegraphics[width=\textwidth]{plots/dodo.png}
%     \end{subfigure}
% \end{figure}
% \endgroup
% }
% \setcounter{figure}{0}
% \setlength{\parskip}{1.8pt plus0pt minus3.0pt}
{
\begingroup
\begin{figure}[t]
\vspace{-0.85cm}
    \quad
    \begin{subfigure}{0.1275\textwidth}
        \includegraphics[width=\textwidth]{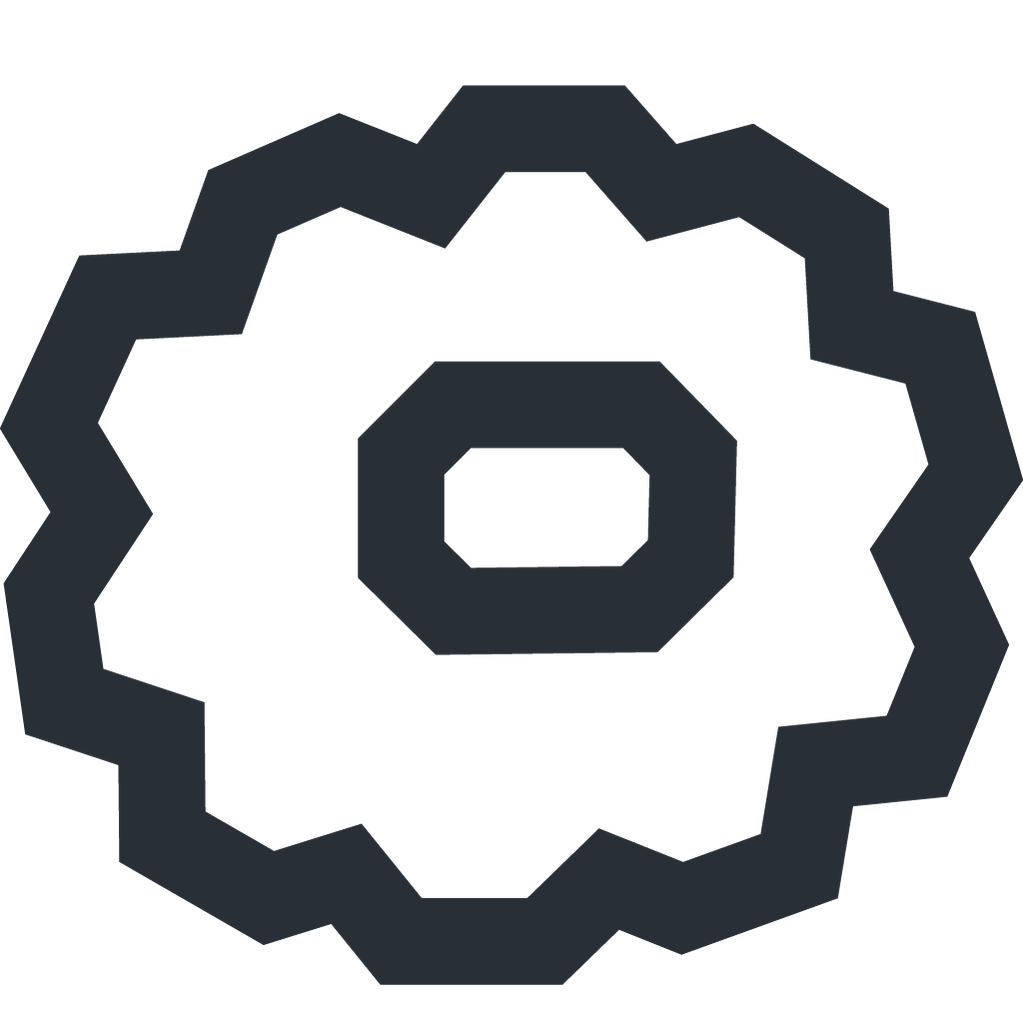}
    \end{subfigure}
    \hfill
    \begin{subfigure}{0.1\textwidth}
        \includegraphics[width=\textwidth]{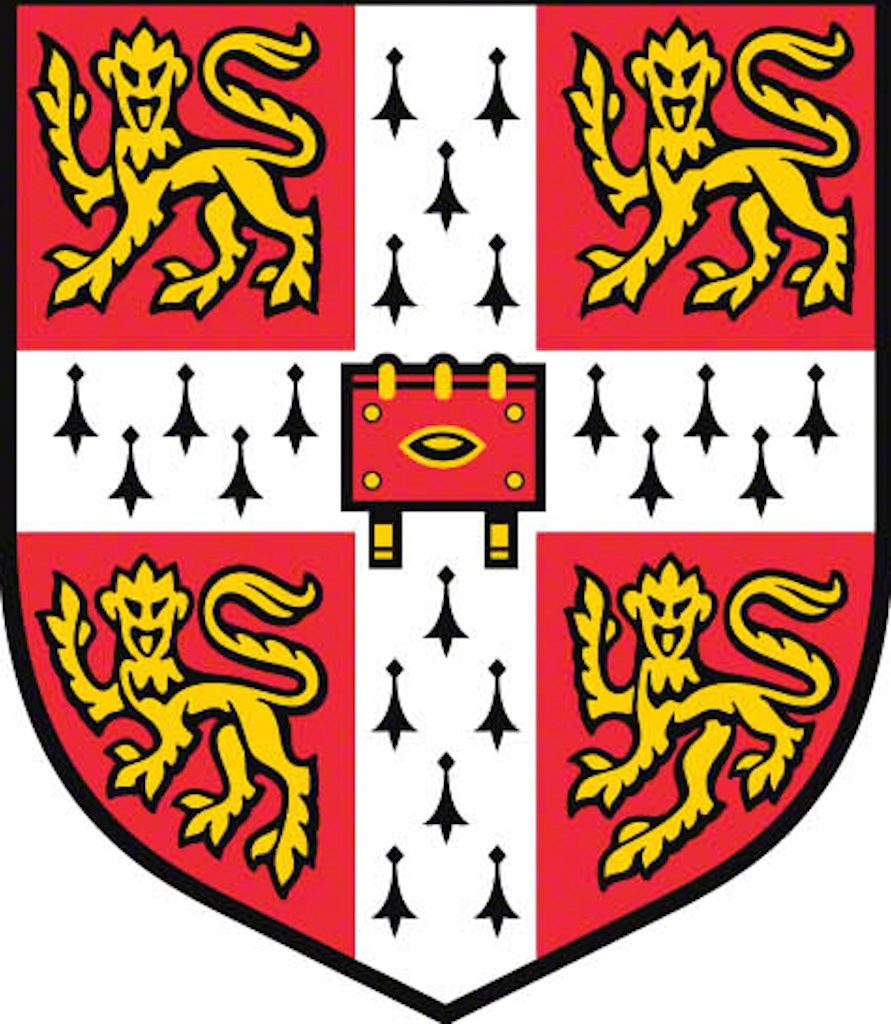}
    \end{subfigure}
    \hfill
    \begin{subfigure}{0.1275\textwidth}
        \includegraphics[width=\textwidth]{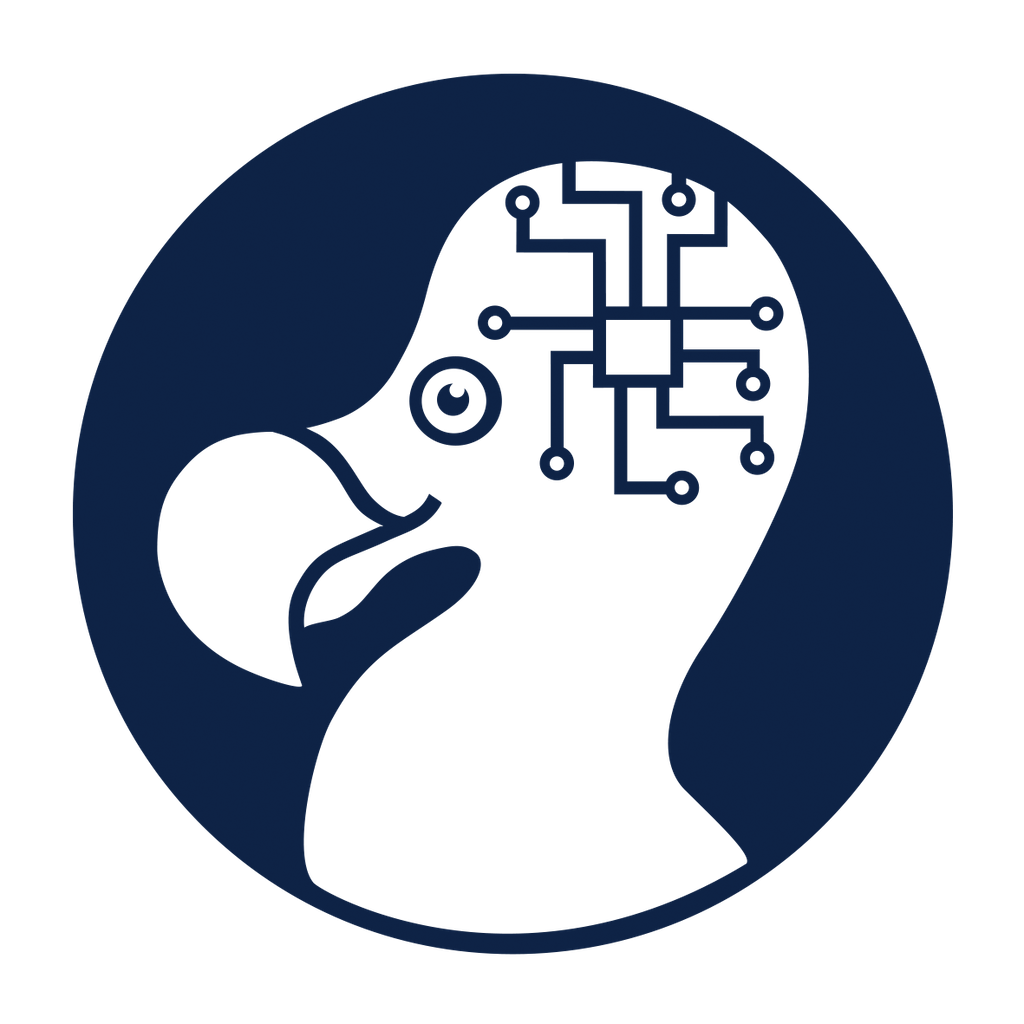}
    \end{subfigure}
    % \\
    % \vspace{-0.5cm}
    % \hrulefill
    % \vspace{0.5cm}
\end{figure}

\endgroup
}

\maketitle

% \renewcommand{\thefootnote}{\myfnsymbol{footnote}}
% % Layout the \thanks notes in the order you want
% \footnotetext[6]{Corresponding author: Alex Iacob \href{mailto:aai30@cam.ac.uk}{\nolinkurl{aai30@cam.ac.uk}}}
% \footnotetext[7]{Equal contribution}
% \footnotetext[1]{Department of Computer Science and Technology, University of Cambridge}
% \footnotetext[2]{Flower Labs}
% \footnotetext[3]{Beijing University of Posts and Telecommunications}

\renewcommand{\thefootnote}{\myfnsymbol{footnote}}
\footnotetext{\textsuperscript{\textdagger}Corresponding author: Alex Iacob \href{mailto:aai30@cam.ac.uk}{\nolinkurl{aai30@cam.ac.uk}}; \textsuperscript{*,**} Equal contribution; \textsuperscript{1}University of Cambridge; \textsuperscript{2}Flower Labs; \textsuperscript{3}Beijing University of Posts and Telecommunications.}

% \setcounter{footnote}{0}% Restart footnote counter
% % Footnotes for rest of document uses \fnsymbol (or whatever you choose)
% \renewcommand{\thefootnote}{\arabic{footnote}}

\begin{abstract}
\input{files/abstract}
\end{abstract}
\input{files/intro}

\input{files/methods}

\input{files/eval}

\input{files/related_work}
\input{files/conclusion}

\FloatBarrier
% \pagebreak
\subsubsection*{Acknowledgments}
All costs for the computation used for this work was funded by Flower Labs, and the research conducted by a team of researchers from Flower Labs and The University of Cambridge. Support for university-based researchers came from a variety of sources, but in particular, the following funding organizations are acknowledged: the European Research Council (REDIAL), the Royal Academy of Engineering (DANTE), and the Ministry of Education of Romania through the Credit and Scholarship Agency.

\bibliography{iclr2025_conference, ref-cdq}
\bibliographystyle{iclr2025_conference}
\newpage
\appendix
\input{appendix/experimental_details}

\input{appendix/additional_results}
\input{appendix/applications}
\input{appendix/narrative_sections}

\end{document}

%% file: files/abstract.tex
Language Model pre-training uses broad data mixtures to enhance performance across domains and languages. However, training on such heterogeneous text corpora requires extensive and expensive efforts. Since these data sources vary significantly in lexical, syntactic, and semantic aspects, they cause negative interference or the ``curse of multilinguality''. To address these challenges we propose a communication-efficient pre-training framework, \dept. Our method decouples embeddings from the transformer body while simultaneously training the latter on multiple data sources without requiring a shared vocabulary. \dept can: (1) train robustly and effectively under significant data heterogeneity, (2) minimize token embedding parameters to only what the data source vocabulary requires, while cutting communication costs in direct proportion to both the communication frequency and the reduction in parameters, (3) enhance transformer body plasticity and generalization, improving both average perplexity (up to $\mathbf{20\%}$) and downstream task performance, and (4) enable training with custom optimized vocabularies per data source. We demonstrate \dept's potential via the first vocabulary-agnostic federated pre-training of billion-scale models, reducing communication costs by orders of magnitude and embedding memory by $\mathbf{4-5\times}$.

%% file: files/intro.tex
\section{Introduction}

Language models (LMs) rely on sizable pre-training datasets to generalize across tasks~\citep{gpt2,gpt3}, and languages~\citep{HowGoodIsMultilingualBert, MonolingualTransferArtetxe,zhao2024breaking}. More data boosts generalization and language acquisition~\citep{TrainingComputeOptimalLLMs}. However, scaling data creates a heterogeneous mix of \textbf{data sources}—different domains and languages—that challenges LMs. Issues like \textit{Negative interference}~\citep{NegativeInterferenceMetaLearning}, where diverse sources compete for capacity, and the \textit{Curse of Multilinguality}~\citep{CurseOfMultilingualityUnsupervisedCrossLingual}, where adding languages yields diminishing returns, especially on low-resource languages~\citep{LowResourceLanguagesSurvey}, persist.

Existing methods for pre-training on heterogeneous data are costly and complex. Multilingual models like \texttt{BERT}~\citep{BERT}, \texttt{XLM}~\citep{CurseOfMultilingualityUnsupervisedCrossLingual}, and \texttt{mT5}~\citep{mC4} require temperature-tuning of language sampling ratios for each model-tokenizer pair, involving expensive model selection to optimize perplexity~\citep{CurseOfMultilingualityUnsupervisedCrossLingual}. Large Language Models~(LLMs) such as LLaMA handle heterogeneous data with intensive “language-specific heuristics and model-based filters”~\citep{llama3}. However, these methods still face challenges such as vocabulary dilution~\citep{HowGoodIsYourTokenizer} and sub-optimal cross-lingual/domain performance~\citep{WhenIsMultilingualityCurse}.

\begin{figure}[t]
    \centering
    \noindent\includegraphics[width=\textwidth]{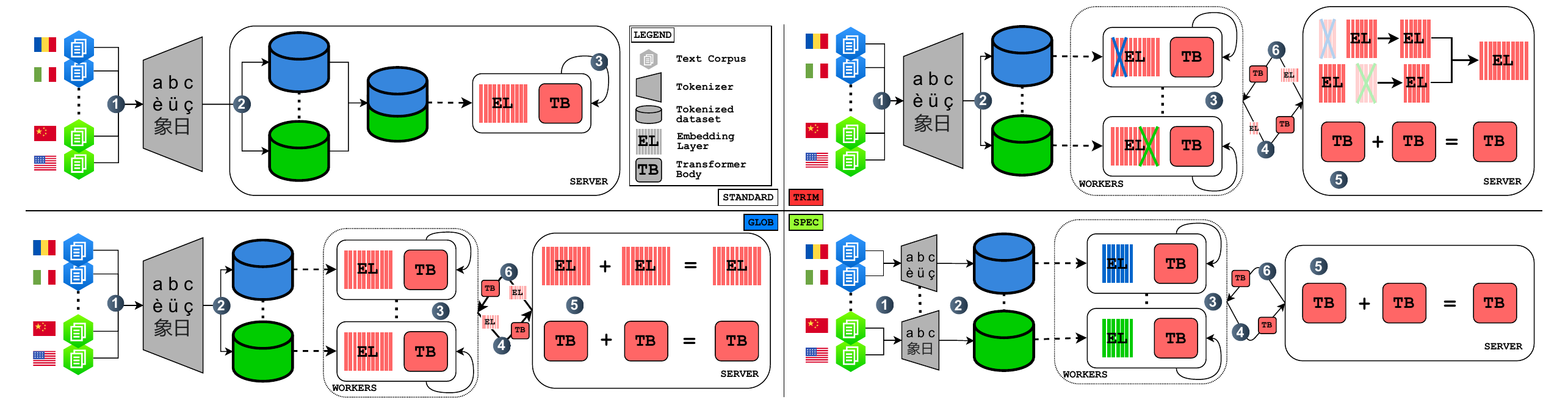}
    \caption{
    Pipeline for \dept variants: \trim (top-right), \glob (bottom-left), \spec (bottom-right), with the \texttt{STANDARD} approach (top-left). The numbered pipeline steps proceed as follows: (1) text corpora are processed into a vocabulary and tokenizer (global for \texttt{STANDARD}, \glob, and \trim; global or personalized for \spec); (2) corpora are tokenized into a pre-tokenized dataset; (3) \texttt{WORKERS} train the model on their pre-tokenized data; (4) partial training results are collected; (5) results are aggregated; (6) the new model is sent to \texttt{WORKERS}. Steps 3–6 repeat to convergence.}
    
    \label{fig:summary_diagram}
    \vspace{-0.3cm}
\end{figure}

This paper proposes a communication-efficient pre-training pipeline to address heterogeneous data challenges. Observing that custom vocabularies boost performance across languages~\citep{HowGoodIsYourTokenizer} and domains~\citep{TransfomersCanDoArithmetic}, we propose partially or fully \emph{decoupling} the \emph{embedding space} from transformer bodies. This approach optimizes embeddings for specific data sources while the transformer learns abstract representations. We introduce \textbf{D}ecoupled \textbf{E}mbeddings for \textbf{P}re-\textbf{T}raining (\dept) in three variants, \glob, \trim, and \spec (see \cref{fig:summary_diagram}), each increasingly leveraging specialized representations to allow pre-training with distinct domains/languages, embedding matrices, and vocabularies. For example, our \spec variant scales the vocabulary size linearly with the number of data sources without increasing memory requirements.

\dept enables pre-training on heterogeneous data sources with unique vocabularies and linguistic features. In the \dept pipeline, data sources are isolated as silos, akin to clients in cross-silo Federated Learning (FL)~\citep{mcmahan2017communication}. \dept trains on each silo and aggregates contributions like FL clients. This work examines whether an LM can converge on data mixtures without a shared (1) output vocabulary, (2) embedding matrices, or (3) tokenization.

In summary, our work brings the following scientific contributions:
\begin{compactenum}
    \item \dept offers a solution to train an effective transformer body without shared global embeddings, avoiding the time, electricity, and carbon-intensive HPO tuning.
    \item \dept reduces the memory requirements of models by $ {\scriptstyle \mathcal{O} ((|\mathcal{V}| - \overline{|\mathcal{V}_k|}) d_{\mathrm{model}})}$ where ${\scriptstyle \overline{|\mathcal{V}_k|}}$ is the average data source's vocabulary size, ${\scriptstyle |\mathcal{V}|}$ the global vocabulary size, and ${\scriptstyle d_{\mathrm{model}}}$ the embedding dimension. For multilingual models, this can save up to $\mathbf{80\%}$ of the embedding-matrix size, reducing $\mathbf{409}$M parameters for our billion-scale multilingual model.
    \item \dept-based transformer bodies show better generalization, achieving lower validation perplexities, with improvements upward of $15.3-20\%$ to average perplexity. \dept models also excel in model plasticity, quickly adapting to new languages/domains. Finally, \dept improves downstream fine-tuning performance on Natural Language Understanding tasks.
    \item \dept is communication-efficient in distributed settings, reducing communication costs compared to standard distributed data parallelism~\citep{FSDP_Pytorch} proportionally to its communication frequency. Compared to communication-efficient SGD~\citep{LocalSGD}, it obtains further reductions proportional to the size of the model embeddings. Additionally, \dept enables vocabulary-agnostic federated pre-training for the first time.
\end{compactenum}
\vspace{-0.25cm}

% While constructing a global embedding matrix efficiently remains the primary difficulty in training such models, our early experiments indicate that this becomes increasingly less problematic with model scale.

%We have not fully explored the properties and capabilities of models trained using \dept; however, we are eager to offer this new paradigm to the community and to develop it further as they see fit.

%% file: files/methods.tex
\section{Decoupled Embeddings For Pre-training~(\dept)} \label{sec:methodology}

Prior work attributes the \textit{Curse of Multilinguality} to capacity contention, vocabulary dilution~\citep{CurseOfMultilingualityUnsupervisedCrossLingual}, and suboptimal tokenization~\citep{HowGoodIsYourTokenizer}. These issues affect embeddings—even though the transformer body is vocabulary-independent~\citep{VocabMatching}. For instance, while English may need $150\,000$ tokens~\citep{VocabularyScalingLaws}, multilingual models allocate $250\,000$ tokens across hundreds of languages, leading to dilution, contention, and under-representation~\citep{LowResourceLanguagesSurvey}. We \textbf{propose} decoupling embeddings during training to enable custom parameters that reduce contention and vocabularies that avoid dilution and suboptimal tokenization.

We argue that training the transformer body without shared embeddings is feasible. Our \textbf{intuition} is based on evidence that: (a) transformers adapt to new languages by re-learning embeddings~\citep{MonolingualTransferArtetxe}; (b) syntactic similarity matters more than subword sharing for performance~\citep{HowGoodIsMultilingualBert}; and (c) periodically re-initializing embeddings enhances plasticity~\citep{ActiveForgetting}. This suggests that transformer body performance is partly embedding-independent, allowing decoupling.
Our method, \dept, achieves this decoupling by: (1) tokenizing data sources independently, using a global or custom vocabulary; (2) randomly initializing LM parameters; and (3) training iteratively over random source subsets (see \cref{sec:methodology}). This contrasts with standard pre-training, which uses shared embeddings and draws random samples from a distribution of all sources.

\subsection{Method}\label{subsec:method}

Akin to federated and meta-learning, \dept optimizes a global parameter set $\theta$ (the transformer body) along with optional embeddings $\phi,\psi$ across data sources $S$. It trains iteratively by selecting a subset $S_t\subset S$ each round $t$. For each data source ($k\in S_t$), \dept independently performs inner-loop optimization (\texttt{InnerOPT}, e.g., SGD) and then aggregates the transformer bodies using an outer-loop optimizer (\texttt{OuterOPT}, e.g., FedAvg). We present three variants for managing $\phi$ and $\psi$, offering progressively stronger specialization, and compare them in \cref{subsec:variant_assumptions_and_benefits}.

\input{algorithms/mclr}
\begin{itemize}[noitemsep,topsep=0pt,parsep=2pt,partopsep=0pt]
    \item[\scolorbox{blue!30}{\glob}] \textbf{Shared Embeddings}: Based on FedAvg-like methods, \glob sends a global transformer and embeddings to each data source, which then trains locally. The updated models are aggregated via \texttt{OuterOPT}, making \glob suitable for federated and centralized settings.

    \item[\scolorbox{red!30}{\trim}] \textbf{Partially-decoupled}: Each data source gets a global transformer and embeddings but trims the token embeddings to its local vocabulary $\mathcal{V}_k$, reducing the input/output space. During \texttt{OuterOPT} aggregation, trimmed embeddings are projected to the global vocabulary.
    
    \item[\scolorbox{green!30}{\spec}] \textbf{Fully-decoupled}: Each data source gets a global transformer and, when first sampled, randomly initializes specialized token/position embeddings. These remain local (never aggregated), supporting any vocabulary, including those from specialized tokenizers.
\end{itemize}

\dept replaces the standard pre-training pipeline (\cref{fig:summary_diagram}) for broad pre-training before adaptation~\citep{llama3}. \cref{alg:1} runs in parallel, scales with hardware, and reduces communication. Reduced communication makes it ideal for low-bandwidth settings like cross-silo FL.

\subsection{Trimmed Embedding Aggregation (\trim)}

For data source \(k\), trimmed embeddings $\phi_k\!\in\!\mathbb{R}^{|\mathcal{V}_k|\!\times\!d_{\mathrm{model}}}$ are derived from global ones $\phi\!\in\!\mathbb{R}^{|\mathcal{V}|\!\times\!d_{\mathrm{model}}}$ as $\phi_k\!=\!\mathcal{I}_k \phi$, where $|\mathcal{V}|$ is the global vocabulary size, $|\mathcal{V}_k|$ the source-specific size, and $d_{\mathrm{model}}$ the embedding dimension.
The indicator function $\mathcal{I}_k(i, j)\!=\!\mathbb{I}[\text{the }\!j\text{-th token in $\mathcal{V}$ corresponds to the }\!i\text{-th local token in }\!\mathcal{V}_k]$ selects tokens from $\phi$.
After \texttt{InnerOPT} we create $\hat{\phi}_k\!\in\!\mathbb{R}^{|\mathcal{V}|\!\times d_{\mathrm{model}}}$, using zero-padding for tokens in $\mathcal{V}\!\setminus\!\mathcal{V}_k$, and use $\mathcal{I}_k^\top\!\in\!\mathbb{R}^{|\mathcal{V}|\!\times\!|\mathcal{V}_k|}$ to project $\phi_k$ back, $\hat{\phi}_k\!=\!\mathcal{I}_k^\top\!\phi_k$. Aggregation~(\texttt{OuterOPT}) is then applied to $\{\hat{\phi}_k\}_{k\!\in\!S_t}$ with zero-padding ignored to avoid interference between tokens not shared across sources.

\subsection{Positional Embedding Specialization (\spec)}

Unlike other variants, \spec specializes both token embeddings $\phi$ and positional embeddings $\psi$, as evidence shows syntactic order-dependent properties matter more than subword sharing~\citep{HowGoodIsMultilingualBert}. Thus, \spec is agnostic to vocabulary and sequence length, enabling federated learning without shared tokenization. Without positional specialization, \spec resembles \trim, but with the embedding matrix split across sources and disjoint vocabularies $\{\mathcal{V}_k\}_{k=1}^{K}$ such that $\mathcal{V} = \cup_{k=1}^{K} \mathcal{V}_k$.

\subsection{Variant Characteristics}\label{subsec:variant_assumptions_and_benefits}

\begin{table}[H]
\caption{Memory and communication costs of \dept, where: $\mathcal{M}$ is the number of model parameters; $|\mathcal{V}|$ is the global vocabulary size; $\overline{|\mathcal{V}_k|}$ is the mean data source vocabulary size; $d_{\mathrm{model}}$ is the embedding dimension; $N_{\mathrm{local}} = N/T$ is the number of local steps done per iteration for a total number steps $N$; $\mathcal{L}$ is the sequence length. \glob reduces comms by only communicating every $N_{\mathrm{local}}$ steps while \trim also reduces embedding size. \spec brings further reductions over \trim by not sharing token or position embeddings. The standard baseline is assumed to be distributed~training with per-step synchronization. Concrete numbers for our models~(see \cref{tab:model_architectures}) are shown in \cref{tab:practical_memory_cost_short}.}
\label{tab:memory_comms_costs}
\centering
\begin{tabular}{@{}lccc@{}}
\toprule
\textbf{Method} & \textbf{Memory Cost} & \textbf{Per-step Comms Cost} & \textbf{Vocab Agnostic} \\ \midrule
\textbf{\texttt{STD}} & {\scriptsize $\mathcal{O}(\mathcal{M}$})& {\scriptsize $\mathcal{O}(\mathcal{M})$} & {\scriptsize $\times$} \\ \midrule
\textbf{\glob}     & {\scriptsize $\mathcal{O}(\mathcal{M}$}) & {\scriptsize$\mathcal{O}$}$(\frac{\mathcal{M}}{N_\mathrm{local}})$ & {\scriptsize $\times$} \\ \midrule
\textbf{\trim}     & {\scriptsize $\mathcal{O}(\mathcal{M} - (|\mathcal{V}| - \overline{|\mathcal{V}_k|}) d_{\mathrm{model}})$} & {\scriptsize $\mathcal{O}$}$(\frac{\mathcal{M} - (|\mathcal{V}| - \overline{|\mathcal{V}_k|}) d_{\mathrm{model}}}{N_\mathrm{local}})$ & {\scriptsize $\times$} \\ \midrule
\textbf{\spec}     & {\scriptsize $\mathcal{O}(\mathcal{M} - (|\mathcal{V}| - \overline{|\mathcal{V}_k|}) d_{\mathrm{model}})$} &  {\scriptsize $\mathcal{O}$}$(\frac{\mathcal{M} - (|\mathcal{V}| + \mathcal{L}) d_{\mathrm{model}}}{N_\mathrm{local}})$ & {\scriptsize \checkmark} \\ \bottomrule
\end{tabular}
\end{table}

In most scenarios, practitioners can deploy any of our proposals, obtaining reduced communication and memory costs as shown in \cref{tab:memory_comms_costs}. However, some settings are appropriate for a given variant.

\textbf{\glob} resembles a standard pre-training pipeline. Although it does not explicitly decouple embeddings from the transformer, they decouple over the course of an inner-loop iteration since only local tokens influence them. As a communication-efficient form of SGD, \glob reduces communication costs compared to distributed algorithms such as DDP~\citep{PyTorchDistributed} or FSDP~\citep{FSDP_ZeRO}, which synchronize gradients at every step. However, constructing a global vocabulary requires sufficient knowledge of the dataset and may risk vocabulary dilution and capacity contention.

\textbf{\trim} shares the same assumptions as \glob and can be deployed similarly. It further reduces memory requirements for embeddings to match the data source’s needs ($d_{\mathrm{model}} \times \mathcal{V}_k$), also lowering communication costs. These savings are substantial for multilingual models with large vocabularies\citep{VocabularyTrimming}, for instance, \texttt{mT5} and \texttt{mBART}~\citep{mC4,BART} allocate $40\%-80\%$ of parameters to embeddings. Since our models use \textit{tied weights}~\citep{TiedWeights}, \trim restricts their output space, unlike \glob, bringing a slight impact to perplexity.

\textbf{\spec} enables pre-training across data sources without a shared vocabulary, providing \trim's benefits plus local specialization. Communication costs are minimized by transferring only the transformer body to the outer optimizer and decoupling embeddings, enabling vocabulary-agnostic training. This makes \spec ideal for training a transformer body with unknown or private data. To enable inference, \spec requires a global embedding matrix. While several methods exist~(\cref{subsec:limitations,app:extending_spec}), we use the straightforward approach of multi-phase adaptive pre-training~\citep{DontStopPreTraining}, or continued pre-training with a randomly initialized matrix. This approach follows other techniques for enhancing model capabilities, e.g., long-context pre-training stages~\citep{BERT,llama3} and domain adaptation~\citep{DontStopPreTraining}.

%% file: algorithms/mclr.tex
\begin{algorithm}[t]
\footnotesize
\caption{Decoupled Embedding for Pre-Training (\dept) variants: \scolorbox{blue!30}{\glob} \scolorbox{red!30}{\trim} \scolorbox{green!30}{\spec}}
\begin{singlespace}
\begin{algorithmic}[1]
\vspace{-0.04cm}
\Require{$S$: set of $K$ data sources, $T$: number of rounds}
\Require{$\theta_0$: initial transformer blocks, $\phi_0$, $\psi_0$: optional token/positional embeddings}
\Require{$\{\mathcal{D}_k\}_{k=1}^K$: source-specific datasets, $\{\mathcal{V}_k\}_{k=1}^K$: source-specific vocabularies}
\Require{\texttt{InnerOPT}: inner optimizer, \texttt{OuterOPT}: outer optimizer, e.g., AdamW and FedAvg}
\For{each update round $t = 1, 2, \ldots, T$}
    \State{Randomly select a subset $S_t\subseteq S$ of data sources for round $t$}
    \For{each data source $k \in S_t$ \textbf{in parallel}}
        \State{\scolorbox{blue!30}{$\theta_t^k, \phi_t^k, \psi_t^k \gets \texttt{InnerOPT}(\theta_{t-1}, \phi_{t-1}, \psi_{t-1}, \mathcal{D}_k)$}} \Comment{ \scolorbox{blue!30}{\glob:} Global embeddings}
        \State{\scolorbox{red!30}{$\phi_{t-1}\vert_{\mathcal{V}_k} = \texttt{Trim}(\phi_{t-1},\mathcal{V}_k)$}} \Comment{\scolorbox{red!30}{\trim:} Trim global token embeddings}
        \State{\scolorbox{red!30}{$\theta_t^k, \phi_t\vert_{\mathcal{V}_k}, \psi_t^k  \gets \texttt{InnerOPT}(\theta_{t-1}, \phi_{t-1}\vert_{\mathcal{V}_k},  \psi_{t-1}, \mathcal{D}_k)$}} \Comment{\scolorbox{red!30}{\trim}}
        \State{\scolorbox{green!30}{$\theta_t^k, \phi_t^k, \psi_t^k \gets \texttt{InnerOPT}(\theta_{t-1}, \phi^k_{t-1}, \psi^k_{t-1}, \mathcal{D}_k)$}} \Comment{ \scolorbox{green!30}{\spec:} specialized embeddings}
        
        \State{\scolorbox{white!30}{$\Delta \theta_t^k \gets \theta_t^k - \theta_{t-1}$}}  \Comment{Compute parameter update}
        \State{\hspace{+0.19\fboxsep}\scolorbox{blue!30}{$\Delta \phi_t^k \gets \phi_t^k - \phi_{t-1}$}} \Comment{\scolorbox{blue!30}{\glob:} Compute global token embedding update}
        \State{\scolorbox{red!30}{$\Delta \phi_t\vert_{\mathcal{V}_k} \gets \phi_t\vert_{\mathcal{V}_k} - \phi_{t-1}\vert_{\mathcal{V}_k}$}}  \Comment{\scolorbox{red!30}{\trim:} Compute Trimmed embeddings update}
        \State{\scolorbox{purple!30}{$\Delta \psi_t^k \gets \psi_t^k - \psi_{t-1}$}}  \Comment{\scolorbox{blue!30}{\glob}+\scolorbox{red!30}{\trim:} global positional embedding update}

    \EndFor
    \State{\scolorbox{white!30}{$\theta_t \gets \texttt{OuterOPT}(\theta_{t-1}, \{\Delta \theta_t^k\}_{k \in S_t})$}} \Comment{Apply the updates for the transformer body}
    \State{\scolorbox{blue!30}{$\phi_t \gets \texttt{OuterOPT}(\phi_{t-1}, \{\Delta \phi_t^k\}_{k \in S_t})$}}  \Comment{\scolorbox{blue!30}{\glob:} Apply token updates}
    \State{\scolorbox{red!30}{$\phi_t \gets \texttt{OuterOPT}(\phi_{t-1}, \{\Delta \phi_t\vert_{\mathcal{V}_k}\}_{k \in S_t})$}} \Comment{\scolorbox{red!30}{\trim:} Apply token updates} 
    \State{\scolorbox{purple!30}{$\psi_t \gets \texttt{OuterOPT}(\psi_{t-1}, \{\Delta \psi_t^k\}_{k \in S_t})$}}  \Comment{\scolorbox{blue!30}{\glob}+\scolorbox{red!30}{\trim:} Apply position updates}
\EndFor
\State{\Return{$\theta_T, \phi_T, \psi_T$}}
\end{algorithmic}
\end{singlespace}
\label{alg:1}

\end{algorithm}

%% file: files/eval.tex
\section{Experimental Design}
We propose DEPT as an efficient alternative to standard pre-training to address the \textit{Curse of Multilinguality} and \textit{Negative interference}. In this section, we conduct experiments to evaluate DEPT's performance, focusing on the following research questions:

\begin{itemize}[noitemsep,topsep=0pt,parsep=2pt,partopsep=0pt]
    \item[\textbf{RQ1}] Does \dept allow us to increase the number of training tokens from heterogeneous data? 
    \item[\textbf{RQ2}] Does \dept improve efficiency, in terms of memory and communication costs?
    \item[\textbf{RQ3}] Does \dept improve \textbf{zero-shot generalization} to out-of-distribution data?
    \item[\textbf{RQ4}] Does \dept improve model \textbf{plasticity} when learning new distributions?
\end{itemize}

\subsection{Experimental Setup}
For our experiments, we train decoder-only transformers—currently the most relevant architectures—ranging from $125$M to $1.3$B parameters with $12$ to $24$ blocks (\cref{tab:model_architectures,tab:practical_memory_cost_short}). We use parameter averaging~\citep{fedavg,LocalSGD} as our \texttt{OuterOpt} optimizer, and \texttt{AdamW}~\citep{AdamW} for \texttt{InnerOpt}. Full experimental details on our architecture, training hyperparameters (\cref{tab:model_architectures,tab:practical_memory_cost_short}), dataset, and baseline implementation are in \cref{app:experimental_details}.

\subsection{Multi-domain and Multilingual Methodology}

To evaluate \dept on \textbf{multi-domain} data, we use \texttt{The Pile}~\citep{ThePile}, which includes $22$ subsets. We select $16$ non-copyrighted subsets as our $K$ data sources in \cref{alg:1}: GitHub~(\texttt{GH}), DeepMind Mathematics~(\texttt{DM}), Wikipedia~(\texttt{WK}), Common Crawl~(\texttt{CC}), PubMed Abstracts~(\texttt{PA}), PubMed Central~(\texttt{PC}), USPTO Backgrounds~(\texttt{UB}), NIH Exporter~(\texttt{NH}), FreeLaw~(\texttt{FL}), Enron Emails~(\texttt{EE}), EuroParl~(\texttt{EP}), Stack Exchange~(\texttt{SE}), Philosophy Papers~(\texttt{PP}), ArXiv~(\texttt{AX}), Project Gutenberg~(\texttt{GU}), and Hacker News~(\texttt{HN}). Ubuntu IRC~(\texttt{UI}) is the out-of-distribution dataset. 

For \textbf{multilingual} data, we use MC4~\citep{mC4} with a mix of high, medium, and low-resource languages: English (\texttt{EN}), Italian (\texttt{IT}), and Chinese (\texttt{ZH}) as high-resource; Serbian (\texttt{SR}) and Malay (\texttt{MS}) as medium-resource; and Swahili (\texttt{SW}), Urdu (\texttt{UR}), and Latin (\texttt{LA}) as low-resource. Following \citep{HowGoodIsYourTokenizer}, we train unigram \texttt{SentencePiece}~\citep{SentencePiece} tokenizers with a $50\,257$ vocabulary per data source. \spec variants with optimized per-source vocabularies have the $\texttt{OPT}$ suffix; otherwise, they use a global vocabulary with specialized embeddings.

\subsection{Baselines}

We compare \dept with standard pre-training methods from prior works~\citep{CurseOfMultilingualityUnsupervisedCrossLingual}.
General distributed SGD methods~\citep{PyTorchDistributed,FSDP_ZeRO}, which synchronize gradients at each step and sample from all data sources simultaneously, are labeled as \texttt{STD}.
For multilingual data, we apply temperature-weighted sampling~\citep{BERT} with $\tau = 0.3$, denoted as \texttt{STD} ($\tau=0.3$), as well as uniform, \texttt{STD} ($\tau=0$), and proportional, \texttt{STD} ($\tau=1$), sampling.\footnote{$\tau = 0.3$ was tuned and found effective in \cite{BERT,CurseOfMultilingualityUnsupervisedCrossLingual,mC4}.}
For multi-domain data, we use uniform and proportional sampling.
Given our data sources random sampling (\cref{alg:1}), baselines with uniform sampling are closest to \dept.

Additionally, we compare against the ``pre-training with active forgetting''~(\act) method~\citep{ActiveForgetting}, which enhances plasticity and generalization by periodically randomly resetting embeddings. While \citet{ActiveForgetting} transfer monolingual models between languages, we only utilize their pre-training phase due to our different settings. Like \spec, \act does not produce a fully trained embedding matrix and we employ the same multi-phase adaptive pre-training to create a new embedding matrix from a random initialization. Despite this similarity, \spec is significantly more compute efficient than \act, as it avoids extensive retraining of embeddings. Full details for how we implemented and adapted \act can be found in \cref{app:adapting_actiev_forgetting}. 

\subsection{Metrics} \label{subsec:metrics}

The key characteristics for multi-domain and multilingual pre-training are model \textbf{generalization} and \textbf{plasticity}. \textbf{Generalization} refers to the model's ability to perform well on out-of-distribution (OOD) data, whether in-domain or out-of-domain. We assess in-domain generalization by evaluating the \textit{perplexity} of a model on the test set of each training data source, while OOD generalization is evaluated with unseen datasets. Furthermore, we evaluate \dept's efficacy in building foundation models through downstream tasks: Natural Language Inference via \texttt{MNLI}~\citep{MNLI}, Question Answering via \texttt{RACE}~\citep{RACE}, Sentence Similarity via \texttt{STSB}~\citep{STSB}, and Sentence Classification via \texttt{SST-2}~\citep{SST2} Since we use decoder-only models below the model-size threshold for in-context learning abilities~\citep{gpt3}, we follow \citet{gpt-1} for fine-tuning. The evaluation metrics are \textit{accuracy} (\texttt{MNLI, RACE, SST-2}) and \textit{Pearson correlation} (\texttt{STSB}). The full details are in \cref{app:fine_tuning}.

\textbf{Plasticity} refers to the model's ability to \textbf{quickly} and \textbf{effectively} adapt to a new domain, either to reach target performance with minimal steps or to achieve the highest possible performance. We evaluate the plasticity of \dept models by training them on new data, such as a different domain or language, as well as the most heterogeneous subset of the training data, determined by the size of its local vocabulary within the shared global vocabulary~(see \cref{app:data_sources}).

We assess training robustness and stability using the L2 norm of model parameters and activations. Model divergence in LLMs, as noted by the OPT~\citep{meta_opt} and PaLM~\citep{PALM} teams, correlates with rapid increases in activation norms, a trend also observed in vision transformers~\citep{ScalingVisionTransformers}. While more common at large scales, this issue can arise in smaller transformers depending on learning rate suitability~\citep{ReproducingDivergence}, which, like batch size, is influenced by the gradient noise scale for a given data distribution~\citep{LargeBatchTraining}. Notably, all performance comparisons use optimized baseline hyperparameters~(see \cref{app:experimental_details}).

\subsection{Continued Pre-training and Evaluation}\label{subsec:continued_pre_training}

Once pre-training is complete, some methods, including \spec and \act, lack a global embedding, while others, such as \texttt{STANDARD} pre-training, \glob, and \trim, include one. For \act and \spec (see \cref{subsec:continued_pre_training}), we enable a global (shared) embedding through multi-phase adaptive pre-training~\citep{DontStopPreTraining}. This involves broad \dept pre-training (\cref{alg:1}) followed by continued pre-training on another $15\text{-}19\%$ of the \textbf{total} steps on a non-private dataset using a randomly initialized embedding matrix with a global vocabulary tailored to the specific corpus. For this phase, we use the tokenizer of \citet{GptNeox20B} for English data and \citet{mC4} for multilingual data. These extra steps are applied to all models for fair comparison. While random initialization reveals the quality of the transformer body for all \dept variants, we are also concerned with the independent effectiveness of \glob and \trim in building high-quality global embeddings compared to \texttt{STANDARD} methods. We perform the same $15\text{-}19\%$ extra steps for this comparison, starting from pre-trained embeddings.

Unlike pre-training, this stage requires a sampling strategy. Since \texttt{The Pile} is curated for proportional sampling~\citep{ThePile}, we use it for multi-domain continued pre-training, while uniform sampling is applied to multilingual data to support low-resource languages.

\section{Results}

Our results show that \dept improves transformer body generalization~(\cref{tab:mc4_125M_r,tab:the_pile_350M_pr}), enhancing robustness~(\cref{fig:motivation:pre_train_Activations}), plasticity~(\cref{fig:fed:mc4:125M:balanced:perplexity_ratio}), and downstream performance~(\cref{tab:downstream}) while bringing communication and memory costs reduction~(\cref{tab:practical_memory_cost_short}).

\subsection{\dept Is Robust To Data Heterogeneity~(RQ1)}

\begin{figure}[ht]
    \centering
    \noindent\subfloat[\texttt{The Pile} pre-train, activation norms, 24-block]
    {\includegraphics[width=0.485\columnwidth]{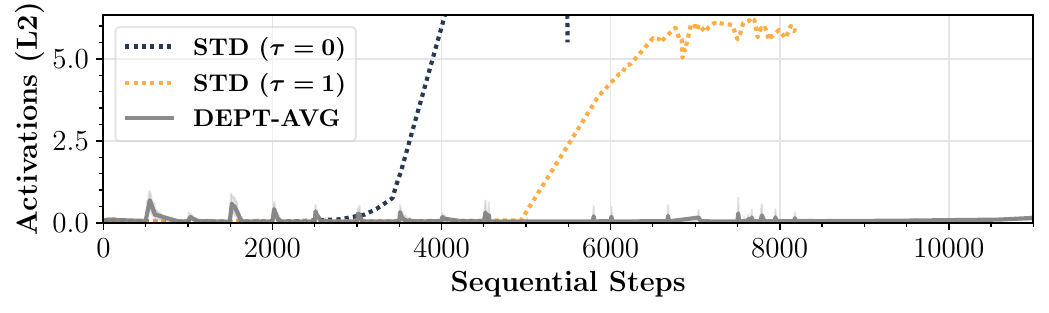}}  \hfill
    \noindent\subfloat[\texttt{The Pile} pre-train, parameter norms, 24-block]
    {\includegraphics[width=0.485\columnwidth]{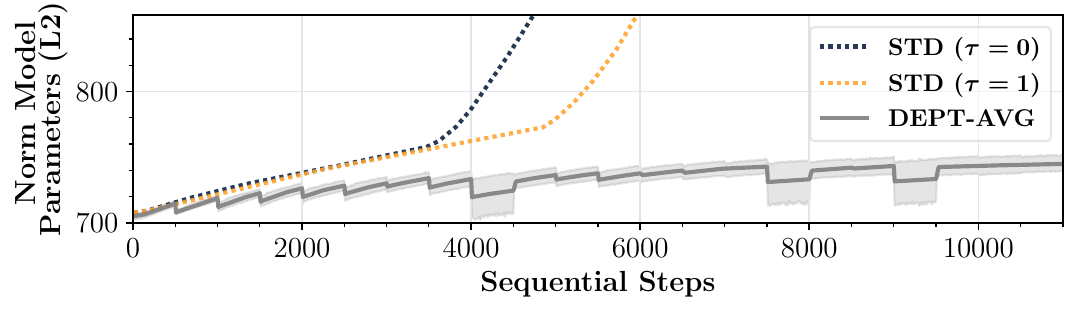}} \hfill
    \caption{Activations and model norms of \texttt{STANDARD}~(\texttt{STD}) training versus \dept~(avg $\pm$ min/max) for a $350$M model trained with identical local hyperparameters—prior to adjusting \texttt{STD} ($\tau=0$) and \texttt{STD} ($\tau=1$) (uniform and proportional sampling) to a lower learning rate. The \texttt{OuterOpt} of \dept introduces regularization effects due to noise-injection~\citep{DontUseLargeBatchesUseLocalSGD}, meta-learning~\citep{REPTILE} characteristics, which constrain these sources~\citep{meta_opt} of model divergence.}
    \label{fig:motivation:pre_train_Activations}
    \vspace{-0.2cm}
\end{figure}

Our experiments demonstrate \dept's robustness to multilingual and multi-domain data heterogeneity. As shown in \textbf{\cref{fig:motivation:pre_train_Activations}}, \dept resists activation divergence and model norm increases, which can halt perplexity improvements or cause divergence~\citep{meta_opt,PALM,ReproducingDivergence}. When using the same local hyperparameters as the baselines, models trained with all \dept variants maintain lower activation norms due to the regularization effects of \texttt{OuterOpt} (\cref{alg:1}). Learning rates for baselines are reduced for later comparisons to ensure convergence.

\subsection{\dept Improves Training Efficiency~(RQ2)}

\input{tables/practical_memory_cost_short}

\textbf{\cref{tab:memory_comms_costs,tab:practical_memory_cost_short}} show that \dept significantly reduces average GPU memory and per-step communication costs compared to DDP. The $500 \times$ memory cost reduction from \glob matches that of Local SGD, as it synchronizes gradients only every $N_{\mathrm{local}}$ steps, allowing GPUs to operate independently in between. \trim further improves memory and communication costs by reducing vocabulary size, shrinking the global embedding matrix by $8\%$ to $32\%$ for multilingual data and by $2\%$ to $78\%$ for \texttt{The Pile}, with the largest reduction ($78\%$) achieved for the mathematics subset (see \cref{app:data_sources} for precise vocab sizes). \spec eliminates embedding-related communication, reducing costs by an additional $13\%$ to $30\%$ for multi-domain data and $34\%$ for multilingual data. Finally, \dept enables efficient training of billion-scale models~(\cref{app:big_model}) on multilingual data, achieving a $714 \times$ reduction in communication costs~(\cref{tab:practical_memory_cost_short}) and a $24\%$ reduction in memory costs.

\subsection{\dept Improves Zero-shot Generalization~(RQ3)}

We show that \dept variants significantly enhance transformer body generalization, outperforming \texttt{STANDARD} pre-training and active-forgetting~(\act) in: (a) perplexity on pre-training validation data, (b) perplexity on OOD validation data, and (c) downstream fine-tuning on \texttt{MNLI, RACE, STSB}. As detailed in \cref{subsec:continued_pre_training}, \dept serves as the first stage of a multiphase adaptive pre-training pipeline, followed by continued pre-training on a non-private dataset. With pre-training data coalesced as in \texttt{STANDARD} training, Our results reflect performance after this phase is applied to baselines as well, ensuring embeddings process the same number of tokens. To gauge tokenizer effectiveness on a dataset, we report the unigram cross-entropy (\texttt{UNIGRAM-CE}) of the unigram model defined by the token frequencies, with higher values indicating a harder-to-model distribution~\citep{VocabularyScalingLaws}(see \cref{app:tokenization_considerations}). Overall, \dept variants win $82.2\% = \tfrac{51}{62}$ of our main comparisons across \texttt{The Pile}, \texttt{MC4} and downstream tasks, producing generalizable and performant transformer bodies. 

\subsubsection{Transformer Body Generalization}~\label{subsubsec:transformer_body_generalization}
\input{tables/the_pile_350M-pr}
\input{tables/mc4_125M-r}

\textbf{\cref{tab:mc4_125M_r,tab:the_pile_350M_pr}} present results where embedding matrices are initialized randomly. \dept variants significantly outperform all baselines across validation sets for multilingual and multi-domain data sources, including high- and low-resource subsets. Min and max improvements, shown in the last two rows of the tables, compare the worst and best \dept variants to the best-performing baseline. The best \dept variant achieves an average performance improvement of $17.3\%$ on \texttt{MC4} and $15.3\%$ on \texttt{The Pile}, while even the worst variant shows improvements of $14.4\%$ and $9.7\%$, respectively. \dept wins $100\%=\tfrac{17}{17}=\tfrac{11}{11}$ comparisons for \texttt{The Pile} and \texttt{MC4}, respectively.
For \texttt{OOD} data, \dept variants outperform by $10\text{-}20\%$ on average for \texttt{MC4} and $1.5\text{-}10.5\%$ on \texttt{The Pile}, despite the high \texttt{UNIGRAM-CE} of OOD data, which makes it more difficult. This demonstrates that \dept produces superior transformer bodies with better generalization. Notably, \trim performs comparably to \glob despite significant reductions in parameter counts and communication costs during pre-training, suggesting that out-of-vocabulary mistakes do not drastically impact performance. For downstream tasks, however, \trim surpasses \glob (\cref{tab:downstream}).
\spec performs similarly to \glob and \trim, even without sharing token embeddings across data sources. The \texttt{SPEC-OPT} variant, trained with unique vocabularies and parameters for each \texttt{The Pile} data source, outperforms \glob on datasets with high \texttt{UNIGRAM-CE} or those dissimilar to natural language, such as multilingual \texttt{EP}, math-heavy \texttt{DM}, code-based \texttt{GH}, and the high-\texttt{UNIGRAM-CE} dataset \texttt{UI}. For \texttt{MC4}, \spec consistently outperforms on OOD datasets with high \texttt{UNIGRAM-CE}.
These results hold across model sizes~(see \cref{tab:the_pile_125M_r}), and across sampling techniques~(\cref{tab:the_pile_350M_r}).

\subsubsection{Pre-trained Embedding Generalization}

\input{tables/the_pile_350M-pm}
\input{tables/mc4_125M-m}

\textbf{\cref{tab:mc4_125M_m,tab:the_pile_350M_pm}} represent cases where the global embedding is initialized using the final global embedding obtained during pre-training, applicable only to the \glob and \trim variants. For \texttt{The Pile}~(\cref{tab:the_pile_350M_pm}), both variants outperform their standard pre-training counterparts, achieving a $5.5\%$ improvement in average accuracy and winning $\tfrac{12}{17}$ comparisons. Two of the lost comparisons, the small subsets \texttt{EN} and \texttt{EP}, are instead won when using uniform sampling~(\cref{tab:the_pile_350M_m}).\input{tables/downstream_table}Furthermore, \dept consistently outperforms when starting from random embeddings due to its superior transformer body. Thus, we argue that differences in performance compared to results in \cref{subsubsec:transformer_body_generalization} are primarily driven by variations in embedding sampling ratios. For \texttt{MC4}~(\cref{tab:mc4_125M_m}), \dept wins $\tfrac{4}{8}$ comparisons for in-distribution data and $\tfrac{1}{3}$ for OOD data, providing disproportionate benefits for the low-resource \texttt{UR} and \texttt{SW} languages. These languages have very high \texttt{UNIGRAM-CE} values, indicating that the global shared tokenizer, trained with temperature-weighted sampling, underserve them. Switching to proportional sampling during continued pre-training improves performance on high-resource languages, winning \texttt{EN}. Similarly to \texttt{The Pile}, the other comparisons are all won when starting from random embeddings. Thus, while \dept may benefit the transformer body, care must be taken to design an appropriate continued pre-training pipeline to effectively fine-tune the embeddings.

\subsubsection{Downstream Generalization}

\textbf{\cref{tab:downstream}} presents the downstream performance of $24$-block \dept models pre-trained and continued pre-trained (with uniform sampling) on \texttt{The Pile}. \dept models consistently outperform the baselines, regardless of initialization, with \trim achieving the best results and \spec matching \glob in wins. Despite occasional losses to \glob in language modeling, we speculate that the restricted vocabulary of \trim forces it to adapt to language shifts, improving generalization, akin to \act's re-initialization but more effective. While \act performs better on downstream tasks than on language modeling \citep{ActiveForgetting}, it is outperformed by \dept. \dept leverages inherent aggregation noise to develop robust parameters without artificial re-initialization, ensuring that parameter updates are not discarded and avoiding the waste of compute cycles.

\subsection{\dept Improves Model Plasticity~(RQ4)}

\begin{figure}[ht]
    \centering
    \subfloat[\texttt{MC4-FULL}, $12$-block.]{\includegraphics[width=0.485\textwidth]{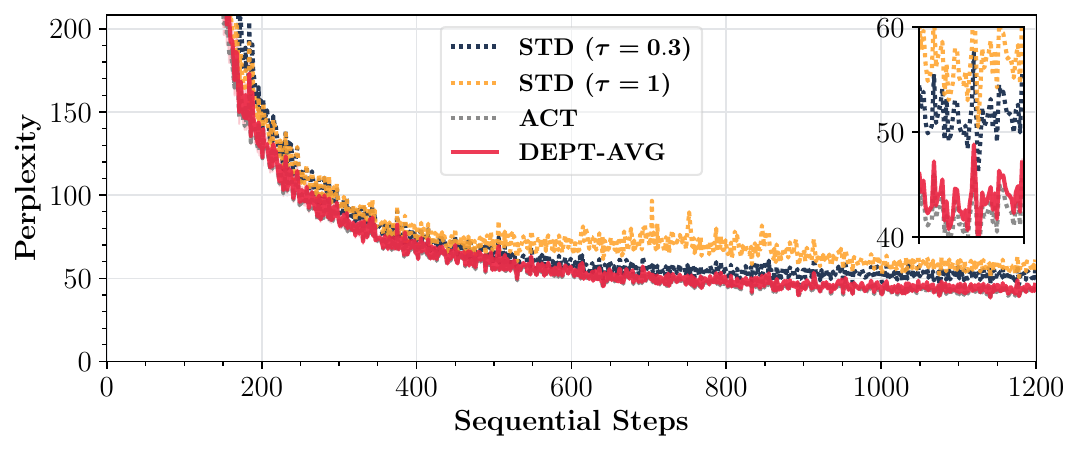}}\hfill
    \subfloat[\texttt{HI}, $12$-block.]{\includegraphics[width=0.485\textwidth]{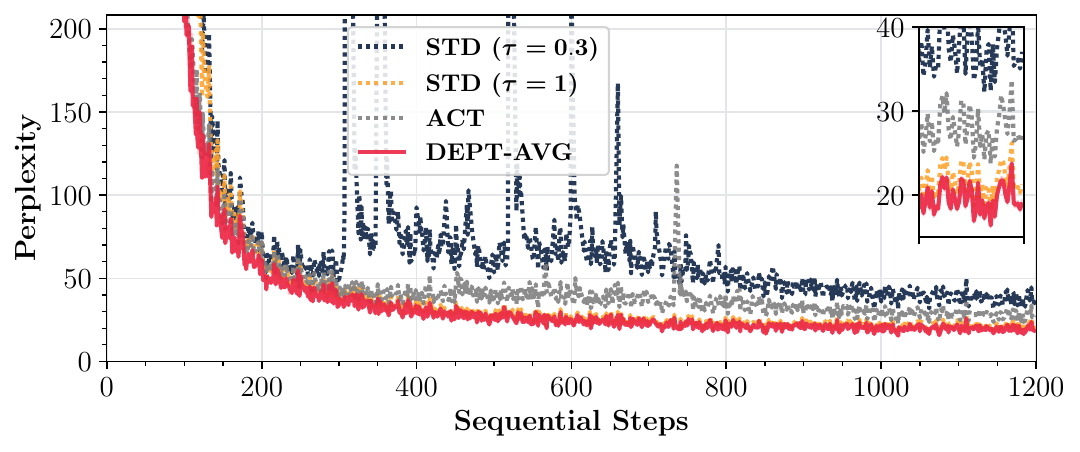}}\hfill
    \caption{Adaptation curves starting from a randomly initialized matrix. \dept variants are always stable in their convergence, reaching the \textbf{lowest perplexity} for the full dataset and the out-of-distribution language~(\texttt{HI}). It is also always the \textbf{fastest} to adapt, full results available in \Cref{fig:fed:mc4:125M:balanced:perplexity_ratio_full}}
    \label{fig:fed:mc4:125M:balanced:perplexity_ratio}
\end{figure}

Finally, we investigate how plastic \dept models are in adapting to either a new data source or to the most heterogeneous subset of the pre-training set. \Cref{fig:fed:mc4:125M:balanced:perplexity_ratio} shows the perplexity adaptation plots when starting from a random initialization on the full pre-training set~(serving as a baseline), the data source with the smallest vocabulary~(\texttt{SW}), or new languages (\texttt{HI,DE}). \dept variants are always the fastest to adapt to each data source and provide the lowest final perplexity; for the full pre-training set, we use perplexity taken over all language validation sets.

%% file: tables/practical_memory_cost_short.tex
\begin{table}[ht]
\vspace{-0.2cm}
\caption{Practical memory and communication costs for \dept, where the total number of steps is \( N = N_{\mathrm{local}} T \) with \( T \) the total number of iterations, and \(\overline{\mathcal{V}_k}\) as the average vocabulary size across data sources. Standard pre-training requires a full in-memory embedding matrix for the global vocabulary while synchronizing gradients every step rather than every \( N_{\mathrm{local}}\) steps. All \dept variants yield communication savings, with \glob as the baseline. \trim provides additional savings proportional to the gap between global and local vocabulary sizes, while \spec further reduces costs  by never communicating embeddings. For the full comparison, see \Cref{tab:practical_memory_cost}.}\label{tab:practical_memory_cost_short} 
\centering
\resizebox{\textwidth}{!}{%
\begin{tabular}{@{}lcccccccc@{}}
\toprule
\textbf{Type} & \textbf{\#Blocks} & \textbf{Method} & $\boldsymbol{N_{local}}$ & $\boldsymbol{T}$ & $\boldsymbol{\overline{|\mathcal{V}_k|}} \pm \sigma$ & $\boldsymbol{\overline{|\mathcal{V}_k|} \times d_{\mathrm{model}}}$ & $\boldsymbol{\overline{\mathcal{M}_k}}$ $\boldsymbol{(\downarrow)}$ & \textbf{Per-step Comms Cost} $\boldsymbol{(\downarrow)}$ \\ \midrule

\textbf{Multilingual}   & $12$ & \textbf{\texttt{STD}} & $5 \times 10^{3}$  & $1$  & $250\,112$             & $192$M   & $278$M $(1\times)$     & $278$M $(1\times)$ \\
\textbf{Multilingual}   & $12$ & \textbf{\glob}             & $500$  & $10$  & $250\,112$             & $192$M   & $278$M $(1\times)$     & $0.56$M $(0.002\times)$ \\
\textbf{Multilingual}   & $12$ & \textbf{\trim}             & $500$  & $10$  & $216\,135 \pm 27\,160$ & $166$M   & $252$M $(0.92\times)$  & $0.5$M $(0.002\times)$ \\
\textbf{Multilingual}   & $12$ & \textbf{\spec}             & $500$  & $10$  & $216\,135 \pm 27\,160$ & $166$M   & $252$M $(0.92\times)$  & $0.17$M $\boldsymbol{(0.0006\times)}$ \\
\textbf{Multilingual}   & $12$ & \textbf{\texttt{SPEC-OPT}} & $500$  & $10$  & $50\,257 \pm 0$        & $38.6$M  & $125$M $\boldsymbol{(0.45\times)}$  & $0.17$M $\boldsymbol{(0.0006\times)}$ \\
\midrule
\textbf{Multilingual ($\boldsymbol{1}$B)} & $24$ & \textbf{\texttt{STD}} & $7 \times 10^{3}$  & $1$  & $250\,112$             & $512.2$M & $1.71$B $(1\times)$    & $1.71$B $(1\times)$ \\
\textbf{Multilingual ($\boldsymbol{1}$B)} & $24$ & \textbf{\texttt{SPEC-OPT}} & $500$  & $14$ & $50\,257 \pm 0$        & $102.9$M & $1.3$B $\boldsymbol{(0.76\times)}$  & $2.4$M $\boldsymbol{(0.001\times)}$ \\
\bottomrule
\end{tabular}%
}
\end{table}

%% file: tables/the_pile_350M-pr.tex
\begin{table}[ht]
\vspace{-0.4cm}
\caption{Validation perplexity ($\downarrow$) for $24$-block models trained on \texttt{The Pile} after \textbf{continued pre-training} with \textbf{proportional} sampling from \textbf{randomly-initialized} embeddings shows that \dept improves performance across \textbf{all} data sources, outperforming baselines by $15.3\%$ on average. \texttt{SPEC-OPT}, using an optimized vocabulary, outperforms \glob on high \texttt{UNIGRAM-CE} sources.}\label{tab:the_pile_350M_pr} 
\centering
\resizebox{\textwidth}{!}{%
\begin{tabular}{@{}lcccccccccccccccccc@{}}
\toprule
\textbf{\begin{tabular}[c]{@{}l@{}}Name\\ ({\scriptsize UNIGRAM-CE})\end{tabular}} & \textbf{\begin{tabular}[c]{@{}c@{}}DM\\ ($6.9$)\end{tabular}} & \textbf{\begin{tabular}[c]{@{}c@{}}EN\\ ($7.9$)\end{tabular}} & \textbf{\begin{tabular}[c]{@{}c@{}}EP\\ ($10$)\end{tabular}} & \textbf{\begin{tabular}[c]{@{}c@{}}FL\\ ($7.8$)\end{tabular}} & \textbf{\begin{tabular}[c]{@{}c@{}}GH\\ ($7.9$)\end{tabular}} & \textbf{\begin{tabular}[c]{@{}c@{}}CC\\ ($7.9$)\end{tabular}} & \textbf{\begin{tabular}[c]{@{}c@{}}PA\\ ($8.2$)\end{tabular}} & \textbf{\begin{tabular}[c]{@{}c@{}}SE\\ ($7.7$)\end{tabular}} & \textbf{\begin{tabular}[c]{@{}c@{}}PP\\ ($9.1$)\end{tabular}} & \begin{tabular}[c]{@{}c@{}}WK\\ ($8.2$)\end{tabular} & \textbf{\begin{tabular}[c]{@{}c@{}}AX\\ ($7.7$)\end{tabular}} & \textbf{\begin{tabular}[c]{@{}c@{}}UB\\ ($7.8$)\end{tabular}} & \textbf{\begin{tabular}[c]{@{}c@{}}PC\\ ($8$)\end{tabular}} & \textbf{\begin{tabular}[c]{@{}c@{}}NH\\ ($8.1$)\end{tabular}} & \textbf{\begin{tabular}[c]{@{}c@{}}GU\\ ($7.7$)\end{tabular}} & \textbf{\begin{tabular}[c]{@{}c@{}}HN \\ ($7.7$) \end{tabular}} & \textbf{\begin{tabular}[c]{@{}c@{}}UI-OOD\\ ($10$)\end{tabular}} & \textbf{\begin{tabular}[c]{@{}c@{}}AVG\\ ($8.1$)\end{tabular}} \\ \midrule
\textbf{\texttt{STD}} ($\tau=0$) & $5.5$ & $44.8$ & $93.5$ & $30.9$ & $8.1$ & $79.6$ & $46.6$ & $23.4$ & $126.6$ & $58.2$ & $14.3$ & $34.1$ & $22.3$ & $58.9$ & $76.3$ & $65.2$ & $163.6$ & $56$ \\
\textbf{\texttt{STD}} ($\tau=1$) & $5$ & $30.6$ & $49.5$ & $20.6$ & $6$ & $56.2$ & $30.9$ & $16.8$ & $81.2$ & $39.1$ & $11$ & $23.7$ & $16.1$ & $39.3$ & $54.6$ & $46.9$ & $99$ & $36.9$ \\
\textbf{\act} & $-$ & $-$ & $-$ & $-$ & $-$ & $-$ & $-$ & $-$ & $-$ & $-$ & $-$ & $-$ & $-$ & $-$ & $-$ & $-$ & $-$ & $-$ \\ \midrule
\textbf{\glob} & $4.8$ & $\mathbf{25.7}$ & $38.2$ & $\mathbf{17.3}$ & $5.4$ & $\mathbf{47.7}$ & $\mathbf{25.7}$ & $\mathbf{14.7}$ & $68.3$ & $\mathbf{32.7}$ & $\mathbf{9.9}$ & $\mathbf{20}$ & $\mathbf{14}$ & $\mathbf{32.2}$ & $\mathbf{46.5}$ & $\mathbf{39.8}$ & $94.8$ & $31.6$ \\
\textbf{\trim} & $4.8$ & $27.3$ & $39.5$ & $18.5$ & $5.6$ & $51.2$ & $27.8$ & $15.4$ & $71.8$ & $35.1$ & $10.3$ & $21.7$ & $14.8$ & $35.1$ & $49.1$ & $42.2$ & $95.7$ & $33.3$ \\
\textbf{\spec} & $4.8$ & $26.7$ & $36.8$ & $18.2$ & $5.5$ & $50.1$ & $27.1$ & $15.1$ & $69.1$ & $34.2$ & $10.1$ & $21.1$ & $14.5$ & $34.3$ & $48.5$ & $41.7$ & $97.6$ & $32.7$ \\
\textbf{\texttt{SPEC-OPT}} & $\mathbf{4.7}$ & $25.9$ & $\mathbf{35}$ & $17.5$ & $\mathbf{5.4}$ & $48.3$ & $26.1$ & $14.7$ & $\mathbf{66.6}$ & $32.8$ & $9.9$ & $20.4$ & $14.1$ & $32.9$ & $47.3$ & $40.5$ & $\mathbf{88.6}$ & $\mathbf{31.2}$ \\ \midrule
\textbf{Min Imp (\%)} & $\mathbf{3.7}$ & $\mathbf{10.6}$ & $\mathbf{20.2}$ & $\mathbf{10.1}$ & $\mathbf{7.4}$ & $\mathbf{8.9}$ & $\mathbf{10.3}$ & $\mathbf{8.4}$ & $\mathbf{11.5}$ & $\mathbf{10.3}$ & $\mathbf{7}$ & $\mathbf{8.6}$ & $\mathbf{8.2}$ & $\mathbf{10.6}$ & $\mathbf{9.9}$ & $\mathbf{10}$ & $\mathbf{1.4}$ & $\mathbf{9.7}$ \\
\textbf{Max Imp (\%)} & $\mathbf{4.2}$ & $\mathbf{15.7}$ & $\mathbf{29.3}$ & $\mathbf{16.3}$ & $\mathbf{11}$ & $\mathbf{15.1}$ & $\mathbf{16.9}$ & $\mathbf{12.9}$ & $\mathbf{17.9}$ & $\mathbf{16.5}$ & $\mathbf{10.6}$ & $\mathbf{15.7}$ & $\mathbf{13.3}$ & $\mathbf{18}$ & $\mathbf{14.7}$ & $\mathbf{15.2}$ & $\mathbf{10.5}$ & $\mathbf{15.3}$ \\ \bottomrule
\end{tabular}%
}
% }
\vspace{-0.3cm}
\end{table}

%% file: tables/mc4_125M-r.tex
\begin{table}[ht]
\caption{ Validation perplexity ($\downarrow$) for 12-block models trained on \texttt{MC4} using \textbf{continued pre-training} with \textbf{uniform sampling} from \textbf{randomly-initialized} embeddings. \dept improves transformer performance across all languages, averaging a $17.3\%$ gain for pre-train data and $20.8\%$ on OOD sources. \spec outperforms \glob on high \texttt{UNIGRAM-CE} OOD data.} \label{tab:mc4_125M_r} 
\resizebox{\textwidth}{!}{%
\begin{tabular}{@{}lccccccccccccc@{}}
\toprule
 & \multicolumn{9}{c}{\textbf{In-Distribution}} & \multicolumn{4}{c}{\textbf{Out-of-Distribution}} \\ \cmidrule(lr){2-10} \cmidrule(l){11-14} 
\textbf{\begin{tabular}[c]{@{}l@{}}Name\\ ({\scriptsize UNIGRAM-CE})\end{tabular}} & \textbf{\begin{tabular}[c]{@{}c@{}}ZH\\ ($9.8$)\end{tabular}} & \textbf{\begin{tabular}[c]{@{}c@{}}UR\\ ($10.5$)\end{tabular}} & \textbf{\begin{tabular}[c]{@{}c@{}}MS\\ ($9.2$)\end{tabular}} & \textbf{\begin{tabular}[c]{@{}c@{}}IT\\ ($7.7$)\end{tabular}} & \textbf{\begin{tabular}[c]{@{}c@{}}SR\\ ($10.5$)\end{tabular}} & \textbf{\begin{tabular}[c]{@{}c@{}}LA\\ ($9$)\end{tabular}} & \textbf{\begin{tabular}[c]{@{}c@{}}EN\\ ($7.5$)\end{tabular}} & \textbf{\begin{tabular}[c]{@{}c@{}}SW\\ ($10$)\end{tabular}} & \textbf{\begin{tabular}[c]{@{}c@{}}Avg (In-D)\\ ($9.3$)\end{tabular}} & \textbf{\begin{tabular}[c]{@{}c@{}}EL\\ ($14.4$)\end{tabular}} & \textbf{\begin{tabular}[c]{@{}c@{}}HI\\ ($13.9$)\end{tabular}} & \textbf{\begin{tabular}[c]{@{}c@{}}DE\\ ($9.7$)\end{tabular}} & \textbf{\begin{tabular}[c]{@{}c@{}}Avg (OOD)\\ ($12.6$)\end{tabular}} \\ \cmidrule(r){1-10} \cmidrule(l){11-14} 
\textbf{\texttt{STD}} ($\tau=0$) & $154.8$ & $38.2$ & $96.8$ & $83.8$ & $73.3$ & $63$ & $112.7$ & $62.8$ & $85.7$ & $5660.8$ & $4600.3$ & $1339.2$ & $3866.8$ \\
\textbf{\texttt{STD}} ($\tau=0.3$) & $129.5$ & $34.5$ & $88$ & $75.4$ & $65.2$ & $56.3$ & $103.7$ & $56.8$ & $76.2$ & $4219.2$ & $3996$ & $1076.3$ & $3097.1$ \\
\textbf{\texttt{STD}} ($\tau=1$) & $84.6$ & $26.8$ & $64.8$ & $55.1$ & $47.1$ & $41.1$ & $77.6$ & $42.4$ & $54.9$ & $3340.3$ & $2514.7$ & $672.5$ & $2175.8$ \\
\textbf{\act} & $96.1$ & $28.8$ & $71.3$ & $60.4$ & $52.3$ & $44.9$ & $85.6$ & $46.3$ & $60.7$ & $2450.2$ & $2412.5$ & $715.9$ & $1859.5$ \\ \cmidrule(r){1-10} \cmidrule(l){11-14} 
\textbf{\glob} & $67.7$ & $\mathbf{22.4}$ & $\mathbf{53.7}$ & $\mathbf{46}$ & $\mathbf{38.6}$ & $\mathbf{33.9}$ & $\mathbf{65.4}$ & $\mathbf{35.2}$ & $\mathbf{45.4}$ & $2308.3$ & $1676.5$ & $559.5$ & $1514.7$ \\
\textbf{\trim} & $\mathbf{67.7}$ & $22.8$ & $55.2$ & $47.5$ & $39.7$ & $35.1$ & $67.2$ & $36.3$ & $46.4$ & $2547.7$ & $1911$ & $567.4$ & $1675.4$ \\
\textbf{\spec} & $69.5$ & $23$ & $55.4$ & $47.8$ & $40.3$ & $34.7$ & $68.1$ & $36.3$ & $46.9$ & $\mathbf{2232.1}$ & $\mathbf{1578.8}$ & $\mathbf{544.7}$ & $\mathbf{1451.9}$ \\ \cmidrule(r){1-10} \cmidrule(l){11-14} 
\textbf{Min Imp (\%)} & $\mathbf{17.8}$ & $\mathbf{14}$ & $\mathbf{14.5}$ & $\mathbf{13.4}$ & $\mathbf{14.6}$ & $\mathbf{14.6}$ & $\mathbf{12.2}$ & $\mathbf{14.3}$ & $\mathbf{14.4}$ & $-4$ & $\mathbf{20.8}$ & $\mathbf{15.6}$ & $\mathbf{10.8}$ \\
\textbf{Max Imp (\%)} & $\mathbf{20}$ & $\mathbf{16.4}$ & $\mathbf{17.1}$ & $\mathbf{16.6}$ & $\mathbf{18.1}$ & $\mathbf{17.4}$ & $\mathbf{15.7}$ & $\mathbf{16.9}$ & $\mathbf{17.3}$ & $\mathbf{8.9}$ & $\mathbf{34.6}$ & $\mathbf{19}$ & $\mathbf{20.8}$ \\ \bottomrule
\end{tabular}%
}
\vspace{-0.3cm}
\end{table}

%% file: tables/the_pile_350M-pm.tex
\begin{table}[ht]
\caption{Validation perplexity ($\downarrow$) for 24-block models trained on \texttt{The Pile} with \textbf{continued pre-training} using \textbf{proportional sampling} from \textbf{pre-trained embeddings}. \dept wins $70\%=\tfrac{12}{17}$ comparisons with \glob consistently outperforming \trim. In \cref{tab:the_pile_350M_pr}, \dept wins the remaining $5$ due to its superior transformer body. Likewise, the \texttt{EN} and \texttt{EP} comparisons are won when using uniform sampling~(\cref{tab:the_pile_350M_m}) as embeddings become more refined on these smaller datasets.}\label{tab:the_pile_350M_pm} 
\centering
\resizebox{\textwidth}{!}{%
\begin{tabular}{@{}lcccccccccccccccccc@{}}
\toprule
\textbf{\begin{tabular}[c]{@{}l@{}}Name\\ ({\scriptsize UNIGRAM-CE})\end{tabular}} & \textbf{\begin{tabular}[c]{@{}c@{}}DM\\ ($6.9$)\end{tabular}} & \textbf{\begin{tabular}[c]{@{}c@{}}EN\\ ($7.9$)\end{tabular}} & \textbf{\begin{tabular}[c]{@{}c@{}}EP\\ ($10$)\end{tabular}} & \textbf{\begin{tabular}[c]{@{}c@{}}FL\\ ($7.8$)\end{tabular}} & \textbf{\begin{tabular}[c]{@{}c@{}}GH\\ ($7.9$)\end{tabular}} & \textbf{\begin{tabular}[c]{@{}c@{}}CC\\ ($7.9$)\end{tabular}} & \textbf{\begin{tabular}[c]{@{}c@{}}PA\\ ($8.2$)\end{tabular}} & \textbf{\begin{tabular}[c]{@{}c@{}}SE\\ ($7.7$)\end{tabular}} & \textbf{\begin{tabular}[c]{@{}c@{}}PP\\ ($9.1$)\end{tabular}} & \textbf{\begin{tabular}[c]{@{}c@{}}WK\\ ($8.2$)\end{tabular}} & \textbf{\begin{tabular}[c]{@{}c@{}}AX\\ ($7.7$)\end{tabular}} & \textbf{\begin{tabular}[c]{@{}c@{}}UB\\ ($7.8$)\end{tabular}} & \textbf{\begin{tabular}[c]{@{}c@{}}PC\\ ($8$)\end{tabular}} & \textbf{\begin{tabular}[c]{@{}c@{}}NH\\ ($8.1$)\end{tabular}} & \textbf{\begin{tabular}[c]{@{}c@{}}GU\\ ($7.7$)\end{tabular}} & \textbf{\begin{tabular}[c]{@{}c@{}}HN \\ ($7.7$\end{tabular}} & \textbf{\begin{tabular}[c]{@{}c@{}}UI-OOD\\ ($10$)\end{tabular}} & \textbf{\begin{tabular}[c]{@{}c@{}}AVG\\ ($8.1$)\end{tabular}} \\ \midrule
\textbf{\texttt{STD}} ($\tau=0$) & $\mathbf{4.4}$ & $\mathbf{13.8}$ & $\mathbf{15.6}$ & $14.9$ & $5.1$ & $41.8$ & $20.7$ & $13$ & $38.3$ & $26.8$ & $9.5$ & $17.1$ & $12.7$ & $23.4$ & $37.2$ & $30.9$ & $\mathbf{54.1}$ & $22.3$ \\
\textbf{\texttt{STD}} ($\tau=1$) & $4.5$ & $19.9$ & $21.9$ & $13.3$ & $\mathbf{4.5}$ & $37$ & $19.7$ & $11.6$ & $47.8$ & $24.5$ & $8.5$ & $16.2$ & $11.5$ & $25$ & $36.4$ & $31.7$ & $54.3$ & $22.8$ \\ \midrule
\textbf{\glob} & $4.5$ & $17$ & $16.1$ & $\mathbf{13.2}$ & $4.5$ & $\mathbf{34.5}$ & $\mathbf{17.9}$ & $\mathbf{11.2}$ & $\mathbf{37.8}$ & $\mathbf{22.4}$ & $\mathbf{8.4}$ & $\mathbf{14.4}$ & $\mathbf{11}$ & $\mathbf{20.6}$ & $\mathbf{35.5}$ & $\mathbf{28.3}$ & $61.2$ & $\mathbf{21.1}$ \\
\textbf{\trim} & $4.6$ & $20.5$ & $23$ & $13.9$ & $4.6$ & $38$ & $20.2$ & $12$ & $49.9$ & $25.1$ & $8.7$ & $16.6$ & $11.8$ & $25.7$ & $38$ & $32.9$ & $56.8$ & $23.7$ \\ \midrule
\textbf{Min Imp (\%)} & $-3$ & $-48.7$ & $-46.9$ & $-3.9$ & $-3.5$ & $-2.7$ & $-2.7$ & $-3.4$ & $-30.1$ & $-2.6$ & $-2.9$ & $-2.7$ & $-2.6$ & $-9.6$ & $-4.3$ & $-6.4$ & $-13.1$ & $-6$ \\
\textbf{Max Imp (\%)} & $-1.2$ & $-23.6$ & $-3$ & $\mathbf{0.9}$ & $-0.8$ & $\mathbf{6.8}$ & $\mathbf{9}$ & $\mathbf{3.4}$ & $\mathbf{1.4}$ & $\mathbf{8.4}$ & $\mathbf{0.9}$ & $\mathbf{11}$ & $\mathbf{4}$ & $\mathbf{12.3}$ & $\mathbf{2.6}$ & $\mathbf{8.4}$ & $-5$ & $\mathbf{5.5}$ \\ \bottomrule
\end{tabular}%
}
\vspace{-0.3cm}
\end{table}

%% file: tables/mc4_125M-m.tex
\begin{table}[ht]
    \caption{ Validation perplexity ($\downarrow$) for 12-block models trained on \texttt{MC4} using \textbf{continued pre-training} with \textbf{uniform sampling} from pre-trained embeddings. \dept achieves a $6.4\%$ improvement in average perplexity for in-distribution data but slightly underperforms for OOD data, winning $50\%=\tfrac{4}{8}$ of in-distribution and $33\%=\tfrac{1}{3}$ of OOD comparisons. In \cref{tab:mc4_125M_r}, \dept wins the remaining cases due to a better transformer body.}\label{tab:mc4_125M_m} 
\resizebox{\textwidth}{!}{%
\begin{tabular}{@{}lccccccccccccc@{}}
\toprule
 & \multicolumn{9}{c}{\textbf{In-Distribution}} & \multicolumn{4}{c}{\textbf{Out-of-Distribution}} \\ \cmidrule(lr){2-10} \cmidrule(l){11-14} 
\textbf{\begin{tabular}[c]{@{}l@{}}Name\\ ({\scriptsize UNIGRAM-CE})\end{tabular}} & \textbf{\begin{tabular}[c]{@{}c@{}}ZH\\ ($9.8$)\end{tabular}} & \textbf{\begin{tabular}[c]{@{}c@{}}UR\\ ($10.5$)\end{tabular}} & \textbf{\begin{tabular}[c]{@{}c@{}}MS\\ ($9.2$)\end{tabular}} & \textbf{\begin{tabular}[c]{@{}c@{}}IT\\ ($7.7$)\end{tabular}} & \textbf{\begin{tabular}[c]{@{}c@{}}SR\\ ($10.5$)\end{tabular}} & \textbf{\begin{tabular}[c]{@{}c@{}}LA\\ ($9$)\end{tabular}} & \textbf{\begin{tabular}[c]{@{}c@{}}EN\\ ($7.5$)\end{tabular}} & \textbf{\begin{tabular}[c]{@{}c@{}}SW\\ ($10$)\end{tabular}} & \textbf{\begin{tabular}[c]{@{}c@{}}Avg (In-D)\\ ($9.3$)\end{tabular}} & \textbf{\begin{tabular}[c]{@{}c@{}}EL\\ ($14.4$)\end{tabular}} & \textbf{\begin{tabular}[c]{@{}c@{}}HI\\ ($13.9$)\end{tabular}} & \textbf{\begin{tabular}[c]{@{}c@{}}DE\\ ($9.7$)\end{tabular}} & \textbf{\begin{tabular}[c]{@{}c@{}}Avg (OOD)\\ ($12.6$)\end{tabular}} \\ \cmidrule(r){1-10} \cmidrule(l){11-14} 
\textbf{\texttt{STD}} ($\tau=0$) & $57.8$ & $21$ & $46.5$ & $40$ & $33.6$ & $29.4$ & $57.5$ & $30.3$ & $39.5$ & $1698.8$ & $1365.7$ & $385.5$ & $1150$ \\
\textbf{\texttt{STD}} ($\tau=0.3$) & $45.5$ & $20.6$ & $41.5$ & $31$ & $\mathbf{31.7}$ & $\mathbf{29.3}$ & $46.1$ & $31.1$ & $34.6$ & $\mathbf{1419.4}$ & $1087.6$ & $321.9$ & $\mathbf{943}$ \\
\textbf{\texttt{STD}} ($\tau=1$) & $44.4$ & $23.9$ & $44.3$ & $\mathbf{25.2}$ & $36.5$ & $33.4$ & $\mathbf{38.3}$ & $36.4$ & $35.3$ & $1583.6$ & $1299.5$ & $\mathbf{285.5}$ & $1056.2$ \\ \cmidrule(r){1-10} \cmidrule(l){11-14} 
\textbf{\glob} & $\mathbf{40.1}$ & $\mathbf{15.5}$ & $\mathbf{30.1}$ & $39.6$ & $39$ & $29.7$ & $40.5$ & $\mathbf{24.6}$ & $\mathbf{32.4}$ & $1737.3$ & $\mathbf{823.4}$ & $335.1$ & $965.3$ \\
\textbf{\trim} & $41.9$ & $16.2$ & $31.3$ & $41.3$ & $40.8$ & $30.8$ & $42$ & $25.6$ & $33.7$ & $1725$ & $855.2$ & $345.6$ & $975.3$ \\ \cmidrule(r){1-10} \cmidrule(l){11-14} 
\textbf{Min Imp (\%)} & $\mathbf{5.6}$ & $\mathbf{21.1}$ & $\mathbf{24.7}$ & $-64$ & $-28.7$ & $-5.1$ & $-9.7$ & $\mathbf{15.5}$ & $\mathbf{2.5}$ & $-22.4$ & $\mathbf{21.4}$ & $-21.1$ & $-3.4$ \\
\textbf{Max Imp (\%)} & $\mathbf{9.7}$ & $\mathbf{24.4}$ & $\mathbf{27.6}$ & $-57.4$ & $-22.8$ & $-1.2$ & $-5.8$ & $\mathbf{18.7}$ & $\mathbf{6.4}$ & $-21.5$ & $\mathbf{24.3}$ & $-17.4$ & $-2.4$ \\ \bottomrule
\end{tabular}%
}
\vspace{-0.4cm}
\end{table}

%% file: tables/downstream_table.tex
% \begin{wraptable}{r}{0.4\textwidth}

\begin{wraptable}{r}{0.5\textwidth}
\caption{The performance on downstream tasks ($\boldsymbol{\uparrow}$), following continued pre-training, shows that \dept models achieve $3\%-7.5\%$ relative improvements over the baselines, with \trim delivering the best results. \dept consistently outperforms baselines. For the full results see Table \ref{tab:downstreamfull}.}
\label{tab:downstream}
\centering
\resizebox{0.5\textwidth}{!}{
\begin{tabular}{@{}lcccc@{}}
\toprule
 & \multicolumn{4}{c}{\textbf{Random Init}} \\
\cmidrule(l){2-5}
\textbf{Name} & \textbf{RACE (ACC)} & \textbf{MNLI (ACC)} & \textbf{STSB (PC)} & \textbf{SST2 (ACC)} \\
\midrule
\textbf{\texttt{STD}} ($\tau=0$)    & $0.50$ & $0.60$ & $0.66$ & $0.79$ \\
\textbf{\texttt{STD}} ($\tau=1$)    & $0.46$ & $0.68$ & $0.73$ & $0.81$ \\
\textbf{\act}              & $0.45$ & $0.66$ & $0.73$ & $0.80$ \\
\midrule
\textbf{\glob}             & $0.51$ & $\mathbf{0.72}$ & $0.78$ & $0.83$ \\
\textbf{\trim}             & $\mathbf{0.53}$ & $0.71$ & $0.78$ & $0.83$ \\
\textbf{\spec}             & $0.52$ & $0.71$ & $\mathbf{0.79}$ & $0.81$ \\
\textbf{\texttt{SPEC-OPT}} & $0.51$ & $0.69$ & $0.77$ & $\mathbf{0.85}$ \\
\midrule
\textbf{Min Imp (\%)}      & $\mathbf{2.9\%}$ & $\mathbf{4.6\%}$ & $\mathbf{5.9\%}$ & $-0.7\%$ \\
\textbf{Max Imp (\%)}      & $\mathbf{5.8\%}$  & $\mathbf{6.1\%}$  & $\mathbf{7.5\%}$  & $\mathbf{4.1\%}$   \\
\bottomrule
\end{tabular}}
\vspace{-0.3cm}
\end{wraptable}

%% file: files/related_work.tex
\vspace{-0.0cm}
\section{Related Work}
Large language models (LLMs) exhibit cross-lingual alignment due to ``incidental bilingualism'' \citep{briakou2023searching} and cross-lingual data sharing \citep{choenni2023languages}. Expanding multilingual data during pre-training can enhance language diversity \citep{le2023bloom} but often results in uneven performance due to data imbalance and low-resource degradation \citep{ding2024data,lai2023chatgpt}. Supervised parallel data (e.g., XLM \citep{conneau2019cross}, PaLM2 \citep{anil2023palm}), Knowledge Transfer \citep{zhang2023bayling,wang2023openchat}, and Domain Adaptation \citep{huang2024survey} face challenges in low-resource settings \citep{chang2023multilinguality,li2024quantifying}, with risks like training instability and catastrophic forgetting \citep{kirkpatrick2017overcoming}. This motivates our novel pipeline, focusing on language heterogeneity, generalization, and plasticity. Vocabulary construction is crucial in multilingual pre-training. Techniques include tokenization with a temperature setting \citep{BERT} and language-clustered vocabularies \citep{LanguageClusteredVocabularies}, though the latter requires predefined clusters. Active forgetting \citep{ActiveForgetting}, a related approach, enhances model plasticity by periodically re-initializing embeddings, easing adaptation to new languages.

%% file: files/conclusion.tex
\section{Conclusion} We investigated pre-training Language Models (LMs) under data heterogeneity, proposing an efficient and robust pipeline, \dept, which supports training under diverse data sources while mitigating \textit{Negative Interference} and the \textit{Curse of Multilinguality}. The core of \dept is decoupling the embedding space from the transformer body during pre-training, offered in three variants with varying degrees of separation. Experiments showed that \dept (1) allows training across heterogeneous data efficiently, (2) reduces the memory footpring of token embedding matrices by $\mathbf{4-5 \times}$, (3) improves model generalization and plasticity with lower perplexity on validation and out-of-distribution test datasets, and (4) supports custom vocabularies per data source, enabling vocabulary agnostic federated pre-training, which we have tested up to billion-scale models and intend to push further.
\subsection{Limitations \& Future Work}\label{subsec:limitations} \dept offers a \textit{pre-training} framework intended to precede further adaptation or fine-tuning. However, \dept models require a final global embedding for practical use. The \glob and \trim variants provide this at the end of pre-training, while \spec does not, suggesting future work on embedding generation methods, such as zero-shot embedding transfer~\citep{mosin2023fine}, vocabulary matching~\citep{VocabMatching} and model stitching~\citep{RelativeRepresentations}. 

%% file: appendix/experimental_details.tex
\section{Experimental Details}\label{app:experimental_details}
\subsection{Model Architectures and Hyperparameters }

\Cref{tab:model_architectures} presents the vocabulary-agnostic hyperparameters of our decoder-only models, while \cref{tab:practical_memory_cost} details vocabulary sizes, \dept-specific parameters, memory costs, and communication costs. Standard pre-training pipeline parameters were chosen based on the recommendations of \citet{TrainingComputeOptimalLLMs} and MosaicML, except for the billion-scale model, where we aligned with the recent state-of-the-art (SOTA) for English federated pre-training by \citet{LLMFL}. We always use a gradient clipping norm of $1$ and $\texttt{ALiBi}$ \citep{alibi} positional embeddings.

During continued pre-training, for models initialized randomly, we begin with $\eta_{\mathrm{max}}$ and decay over $N_{\mathrm{CT}}$ learning steps, allowing quick embedding matrix learning without requiring another full training pass, as is common in language rewiring~\citep{MonolingualTransferArtetxe}. When using pre-initialized models, we start from $\eta_{\mathrm{max}}/2$ since both the model and embeddings are reasonably well-trained. 

Importantly, the only parameter changed between \dept models and baselines is the learning rate $\eta_{\mathrm{max}}$. We use the same learning rate to contrast convergence properties for comparisons in \cref{fig:motivation:pre_train_Activations}. We tune the baselines' learning rate for later comparisons to ensure they perform the same number of training steps, selecting the best checkpoint for a baseline across all experiments. Except for tuning the learning-rate, \dept models always use the same hyperparameters as the baselines during local training.

\input{tables/arch_details}

\input{tables/practical_memory_cost}

We had to select a particular sampling ratio for the continued pre-training using the full pre-training set rather than a single language or domain. Due to its high heterogeneity, we default to uniform sampling for \texttt{MC4} in these cases. In contrast, for \texttt{The Pile}, we preferred proportional sampling as the dataset is entirely in English and has already had its data sources upsampled/downsampled based on usefulness. We also provide results using the alternative sampling policy in \cref{app:additional_results}.

\subsubsection{Software and Hardware}

Our software is based on the MosaicML composer~\citep{mosaicml} library for LLM pre-training and the open-source Flower~\citep{Flower} framework for federated learning. Crucially, we heavily rely on the MosaicML hyperparameters and infrastructure for our \texttt{InnerOPT}, making no changes to it after our embedding-matrix manipulation from \cref{alg:1} has been performed. For the standard baselines, we ran them on a completely unmodified version of the MosaicML codebase~(beyond using our data), which has been independently verified by thousands of users and used to submit accepted conference publications~\citep{paperMosaicML}. 

In terms of hardware, the low communication properties of \dept allowed us to run experiments via a mixture of loaned resources from separate cloud providers. Over the course of our experimentation, we used various machines equipped with either 1 H100 or 1 A100 GPU in the USA, Canada, and Europe, which turned out to be more cost-effective. We rented machines with $4$-$8$ H100 GPUs for the centralized baselines since we could not use Distributed Data Parallelism techniques over low-bandwidth internet connections. When the standard training baseline has a sufficiently low learning rate to converge, the difference in training time is driven by three factors. 

First, the throughput achieved by individual workers: for \glob, this should be identical to standard pre-training as the model in memory remains unchanged. For \trim and \spec, the reduced memory requirements may allow increasing the device micro-batch size in certain scenarios~(but not the global batch size, which heavily influences optimization properties). This depends heavily on the hardware; for example, in \text{DeepMind Mathematics} workloads, \trim or \spec can double the device micro-batch size, and similarly for \texttt{SPEC-OPT} in the case of multilingual data. 

Second, the communication topology significantly impacts wall clock time. For instance, in a $10$ Gbps bandwidth connection using Ring AllReduce for aggregation across workers, \dept can reduce training time by $33\%$ for a $1$ billion parameter model. In cases with a very fast connection, such as InfiniBand, the training time difference is primarily determined by throughput differences.

Third, the number of local data sources and the number of available workers impact the total training time, for \dept we always scale the number of workers to match the number of data sources exactly.

\subsubsection{Hyperparameter Tuning Methodology}\label{app:hyperparameter_tuning}

Given that MosaicML provides hyperparameter-tuned models on the C4~\citep{C4} dataset, we use their learning rate schedule and number of training steps as a starting point. In the case of \dept, we find that we can always use the MosaicML parameters since the \texttt{OuterOpt} application of \dept acts as a regulariser via noise-injection~\citep{DontUseLargeBatchesUseLocalSGD} and meta-learning effects~\citep{REPTILE}. This makes \dept models highly unlikely to diverge, even under extreme data heterogeneity and without a shared input or output space. In the case of standard training baselines, we gradually lower the learning rate, starting from the one reported in \cref{tab:model_architectures}.

We begin with the maximum learning rate $\eta_{\mathrm{max}}$ and systematically reduce it on a coarse grid in intervals of $0.5 \times 10^{-5}$:

\[
\eta = \eta_{\mathrm{max}} - 0.5k \times 10^{-5}, \quad k \in \{0, 1, 2, \dots, K\},
\]

where $k$ represents the step index, and $K$ is chosen such that $\eta > 0$ at the final step. Given that the length of the cosine cycle is directly extrapolated from known scaling laws on the number of tokens that the model needs to train on for compute-optimality~\citep{TrainingComputeOptimalLLMs}, approximately $20$ tokens per parameter, we stop as early as we find a learning rate that can complete the entire cosine schedule. Then, we choose the best-performing checkpoint, according to validation perplexity, across all experiments. We report these values in \cref{tab:model_architectures}.

This hyperparameter search does not cover all possible relevant parameters; given enough resources, we would also tune the gradient clipping norm. Furthermore, we could tune the batch size using the empirical model of large-batch training proposed by \citet{LargeBatchTraining}. Given that the appropriate learning rate depends on the chosen batch size and the desired target loss, such an optimization would require hundreds of experiments across all baselines to find an optimal configuration. 

\subsubsection{Adapting Active Forgetting}\label{app:adapting_actiev_forgetting}

To implement the active forgetting baseline~\citep{ActiveForgetting}, \act, we had to adapt the methodology to decoder-only models, which train with far fewer steps. To achieve this, we use a forgetting frequency of $500$ steps, equal to \dept's $N_{\mathrm{local}}$. We also use a cosine scheduler for the body with the same parameters as shown in \cref{tab:model_architectures}; however, we schedule the embedding matrix independently across the $500$ steps using the same scheduler but setting $\eta^\prime_{\max} = 500$. Finally, we selected the checkpoint with the lowest validation perplexity for continued pre-training in a forgetting cycle.

\subsection{Data Sources}\label{app:data_sources}

We quantify the lexical heterogeneity of a dataset based on \textit{lexical similarity} between data sources. A simple similarity measure is the size of the intersection of subwords between vocabularies. The smaller the intersection, the more dissimilar the vocabularies, and thus, the more challenging it becomes to train a shared tokenizer effectively across different domains or languages. For this section, we use the size of local vocabulary as a subset of the global vocabulary as a proxy, with smaller local vocabulary indicating that global tokenization does not serve a particular data source well.

Our default global tokenizer for multilingual data is that proposed by \citet{mC4}, with $\mathcal{V}= \num{250112}$ tokens.
Owing to its diverse pre-training, the mT5~\citep{mC4} tokenizer is a robust default choice, employed in recent works such as project Aya~\citep{ProjectAya}. However, its coverage of hundreds of languages does come with many shortcomings relating to the capacity allocated to each language. To showcase these challenges, we carefully selected languages from distinct families in the \texttt{MC4 subset}, including \texttt{English (EN)}, \texttt{Italian (IT)}, \texttt{Serbian (SR)}, \texttt{Swahili (SW)}, \texttt{Urdu (UR)}, \texttt{Latin (LA)}, \texttt{Chinese (ZH)}, and \texttt{Malay (MS)}. The corresponding vocabulary sizes of our languages are as follows: $\{247\,720, 211\,332, 208\,391, 170\,984, 188\,002, 220\,757, 240\,566\}$. Among these, \texttt{Swahili (SW)} is the most heterogeneous, as determined by its small subset of $170\,984$ tokens.

Our global tokenizer for English data was trained on \texttt{The Pile} \citep{ThePile} and proposed by \citet{GptNeox20B} with $\mathcal{V} = 50\,257$ tokens. We selected \texttt{The Pile} as our multi-domain dataset for several reasons. \texttt{The Pile} is a diverse, large-scale dataset specifically designed for training large language models (LLMs). Its diversity spans domains such as scientific papers, news, books, and web content, providing a comprehensive foundation for capturing varied linguistic patterns. Among the various subsets of \texttt{The Pile}, \textit{DM Mathematics} stands out as the most heterogeneous. This subset contains only $11,090$ tokens from the global vocabulary, significantly fewer than other subsets. Here are the sizes of other subsets in terms of their unique tokens from the global vocabulary: $\{49\,362, 49\,783, 46\,766, 49\,469, 49\,700, 47\,865, 48\,720\}$ $\{11\,090, 44\,249, 42\,957, 44\,432, 49\,992, 49\,841, 47\,687, 49\,961, 46\,825\}$. While this indicates much lower heterogeneity than in multilingual settings, vocabulary choice may still impact highly specialized model capabilities such as mathematical reasoning. 

\subsubsection{Tokenization Considerations}\label{app:tokenization_considerations}

One of the major challenges when representing multiple data sources with a single tokenizer is \textit{vocabulary dilution}. To maximize coverage, a tokenizer that aims to cover multiple languages or domains often needs to adopt many short subwords. This increases the \textit{tokenizer fertility} (i.e., the number of tokens produced per unit of text)~\citep{HowGoodIsYourTokenizer} and also raises the overall \textit{description length} — the total number of tokens required to represent the same data. This trade-off negatively affects the \textit{compression ratio}, as the same amount of information requires more tokens, reducing the model’s sample efficiency~\citep{VocabularyScalingLaws}. When non-uniform sampling ratios are used during pre-training, high-resource languages tend to have better \textit{fertility} than low-resource languages. This means high-resource languages are better represented in the vocabulary, and their tokens are more likely to be shared across the model’s parameters, improving their performance.
In contrast, low-resource languages suffer from poor fertility, where their unique vocabulary tokens are underrepresented, leading to worse performance. For example, \texttt{Swahili (SW)} and \texttt{Urdu (UR)} are low-resource languages that face these challenges. Our \spec method allows us to avoid many of these issues by providing an optimized vocabulary to a data source at the cost of losing a shared vocabulary and updating several embedding matrices. An alternative approach is to cluster vocabularies~\citep{LanguageClusteredVocabularies} to obtain subword sharing between more relevant languages. However, this requires that participating data sources are known in advance, do not change significantly, and that the appropriate number of clusters is also known.

To account for the effectiveness of a tokenizer on a given language, we report unigram cross-entropy in our experiments, which represents how effective a simple unigram model based on the tokenizer is on that data source as a proxy for the effectiveness of the tokenization. If the unigram cross-entropy is high on a given data source, it is likely underserved by the tokenization. Thus, all improvements brought about by using a more complex language model must consider this baseline. It can also be used to compute unigram-normalized cross-entropy or perplexity, a language modeling performance metric that is comparable across different vocabulary sizes~\citep{VocabularyScalingLaws}.

%% file: tables/arch_details.tex
\begin{table}[ht]
\caption{Architectural details and vocabulary-independent hyperparameters of our models.  The number of transformer blocks is denoted by \#Blocks, the number of attention heads by \#Heads, and the expansion ratio refers to the ratio of the hidden dimension in the feedforward layers. The total number of model parameters is $\mathcal{M}$, the vocabulary size is $|\mathcal{V}|$, and the model embedding dimension is $d_\mathrm{model}$. We train standard decoder-only transformers whose body ranges in size from $86.4$M to $1.2$B independent of embeddings. As we see in \cref{tab:practical_memory_cost}, the size of the embedding matrix can change the model size drastically. Our batch size is $|\mathcal{B}|$ while $|S_t|/|S|$ is our sampling ratios for the various data sources. The $\beta_1,\beta_2$ pair are AdamW parameters while the $S_c$ tuple represents the parameters of the cosine scheduler that we use, including the decay alpha $\alpha$, the decay period $\eta_{\mathrm{max}}$, and the total number of sequential steps $N$. Finally, we show the number of continued pre-training steps $N_{\mathrm{CT}}$ that we use, representing $15\%$ of total steps for the $298$M model and $19.3\%$ for the $86.4$M model. All of our models use a sequence length of $2048$. We followed the hyperparameters of \citet{LLMFL} for the billion-scale federated pre-training. We report the tuned $\eta_{\max}$,  for each baseline according to \cref{app:hyperparameter_tuning}, $\eta^{\mathrm{STD}(\tau=0)}_{\max}$, $\eta^{\mathrm{STD}(\tau=0.3)}_{\max}$,  $\eta^{\mathrm{STD}(\tau=1)}_{\max}$, we find that the embedding resting allows \act to use the same $\eta_{\max}$ as \dept. }\label{tab:model_architectures} 
\centering
\resizebox{\textwidth}{!}{%
\begin{tabular}{@{}lcccccccccclll@{}}
\toprule
\textbf{Type} & \textbf{\#Blocks} & $\boldsymbol{d_{\mathrm{model}}}$ & $\boldsymbol{\mathcal{M} - |\mathcal{V}| \times d_{\mathrm{model}}}$ & \textbf{\#Heads} & \textbf{Exp.~Ratio} & $\boldsymbol{|\mathcal{B}|}$ & $\boldsymbol{|S_t|/|S|}$ & $\boldsymbol{(\beta_1,~\beta_2)}$ & $\boldsymbol{S_C(\alpha,~\eta_{max},~N)}$ & $\boldsymbol{N_{CT}}$ & $\boldsymbol{\eta^{\mathrm{STD}(\tau=0)}_{\max}}$ & $\boldsymbol{\eta^{\mathrm{STD}(\tau=0.3)}_{\max}}$ & $\boldsymbol{\eta^{\mathrm{STD}(\tau=1)}_{\max}}$ \\ \midrule
\textbf{Multi-domain} & 12 & 768 & $86.4$M & 12 & 4 & 256 & $4/16$ & $(0.9,~0.95)$ & $(10^{-1},~6.0 \times 10^{-4},~5\times10^3)$ & $1.2 \times 10^3$ & $~4.5 \times 10^{-4}$ & $~4.5 \times 10^{-4}$ & $~5.0 \times 10^{-4}$ \\
\textbf{Multi-domain} & 24 & 1024 & $298.5$M & 16 & 4 & 256 & $4/16$ & $(0.9,~0.95)$ & $(10^{-1},~3\times10^{-4},~13.5\times10^3)$ & $2.4 \times 10^3$ & $~1.5 \times10^{-4}$ & $~2 \times10^{-4}$ & $~2 \times10^{-4}$ \\
\textbf{Multilingual} & 12 & 768 & $86.4$M & 12 & 4 & 256 & $3/8$ & $(0.9,~0.95)$ & $(10^{-1},~6\times10^{-4},~5\times10^3)$ & $1.2 \times 10^3$ & $4 \times10^{-4}$ & $4 \times10^{-4}$ & $4.5\times10^{-4}$ \\
\textbf{Multilingual} & 24 & 2048 & $1.2$B & 16 & 4 & 512 & $3/8$ & $(0.9,~0.95)$ & $(10^{-1},~2\times10^{-4},~7\times10^4)$ & - & $1\times10^{-4}$ & $1\times10^{-4}$ & $1.5\times10^{-4}$ \\ \bottomrule
\end{tabular}%
}
\end{table}

%% file: tables/practical_memory_cost.tex
\begin{table}[ht]
\caption{ Practical memory and communication costs for \dept, where the total number of steps is \( N = N_{\mathrm{local}} T \) with \( T \) the total number of iterations, and \(\overline{\mathcal{V}_k}\) as the average vocabulary size across data sources. Standard pre-training requires a full in-memory embedding matrix for the global vocabulary while synchronizing gradients every step rather than every \( N_{\mathrm{local}}\) steps. All \dept variants yield communication savings, with \glob as the baseline. \trim provides additional savings proportional to the gap between global and local vocabulary sizes, while \spec further reduces costs with or without optimized vocabularies by never communicating the token or positional matrices. }\label{tab:practical_memory_cost} 
\centering
\resizebox{\textwidth}{!}{%
\begin{tabular}{@{}lcccccccc@{}}
\toprule
\textbf{Type} & \textbf{\#Blocks} & \textbf{Method} & $\boldsymbol{N_{local}}$ & $\boldsymbol{T}$ & $\boldsymbol{\overline{|\mathcal{V}_k|}} \pm \sigma$ & $\boldsymbol{\overline{|\mathcal{V}_k|} \times d_{\mathrm{model}}}$ & $\boldsymbol{\overline{\mathcal{M}_k}}$ $\boldsymbol{(\downarrow)}$ & \textbf{Per-step Comms Cost} $\boldsymbol{(\downarrow)}$ \\ \midrule

\textbf{Multilingual}   & $12$ & \textbf{\texttt{STD}} & $5 \times 10^{3}$  & $1$  & $250\,112$             & $192$M   & $278$M $(1\times)$     & $278$M $(1\times)$ \\
\textbf{Multilingual}   & $12$ & \textbf{\glob}             & $500$  & $10$  & $250\,112$             & $192$M   & $278$M $(1\times)$     & $0.56$M $(0.002\times)$ \\
\textbf{Multilingual}   & $12$ & \textbf{\trim}             & $500$  & $10$  & $216\,135 \pm 27\,160$ & $166$M   & $252$M $(0.92\times)$  & $0.5$M $(0.002\times)$ \\
\textbf{Multilingual}   & $12$ & \textbf{\spec}             & $500$  & $10$  & $216\,135 \pm 27\,160$ & $166$M   & $252$M $(0.92\times)$  & $0.17$M $\boldsymbol{(0.0006\times)}$ \\
\textbf{Multilingual}   & $12$ & \textbf{\texttt{SPEC-OPT}} & $500$  & $10$  & $50\,257 \pm 0$        & $38.6$M  & $125$M $\boldsymbol{(0.45\times)}$  & $0.17$M $\boldsymbol{(0.0006\times)}$ \\
\midrule
\textbf{Multilingual-B} & $24$ & \textbf{\texttt{STD}} & $7 \times 10^{3}$  & $1$  & $250\,112$             & $512.2$M & $1.71$B $(1\times)$    & $1.71$B $(1\times)$ \\
\textbf{Multilingual-B} & $24$ & \textbf{\texttt{SPEC-OPT}} & $500$  & $14$ & $50\,257 \pm 0$        & $102.9$M & $1.3$B $\boldsymbol{(0.76\times)}$  & $2.4$M $\boldsymbol{(0.001\times)}$ \\
\midrule
\textbf{Multi-domain}   & $12$ & \textbf{\texttt{STD}} & $5 \times 10^{3}$  & $1$  & $50\,257$              & $38.6$M  & $125$M $(1\times)$     & $125$M $(1\times)$ \\
\textbf{Multi-domain}   & $12$ & \textbf{\glob}             & $500$              & $10$ & $50\,257$              & $38.6$M  & $125$M $(1\times)$     & $0.25$M $(0.002\times)$ \\
\textbf{Multi-domain}   & $12$ & \textbf{\trim}             & $500$              & $10$ & $45\,554 \pm 9462$     & $35$M    & $121$M $\boldsymbol{(0.97\times)}$  & $0.24$M $(0.002\times)$ \\
\textbf{Multi-domain}   & $12$ & \textbf{\spec}             & $500$              & $10$ & $45\,554 \pm 9462$     & $35$M    & $121$M $\boldsymbol{(0.97\times)}$  & $0.17$M $\boldsymbol{(0.001\times)}$ \\
\midrule
\textbf{Multi-domain}   & $24$ & \textbf{\texttt{STD}} & $13.5 \times 10^3$ & $1$  & $50\,257$              & $51.4$M  & $350$M $(1\times)$     & $350$M $(1\times)$ \\
\textbf{Multi-domain}   & $24$ & \textbf{\glob}             & $500$              & $27$ & $50\,257$              & $51.4$M  & $350$M $(1\times)$     & $0.7$M $(0.002\times)$ \\
\textbf{Multi-domain}   & $24$ & \textbf{\trim}             & $500$              & $27$ & $45\,554 \pm 9462$     & $46.6$M  & $345.2$M $\boldsymbol{(0.97\times)}$& $0.69$M $\boldsymbol{(0.002\times)}$ \\
\textbf{Multi-domain}   & $24$ & \textbf{\spec}             & $500$              & $27$ & $45\,554 \pm 9462$     & $46.6$M  & $345.2$M $\boldsymbol{(0.97\times)}$& $0.6$M $\boldsymbol{(0.002\times)}$ \\
\bottomrule
\end{tabular}%
}
\end{table}

%% file: appendix/additional_results.tex
\FloatBarrier
\section{Additional Results}\label{app:additional_results}

\subsection{Larger Model}
\begin{figure}[H]
    \centering
    \includegraphics[width=\columnwidth]{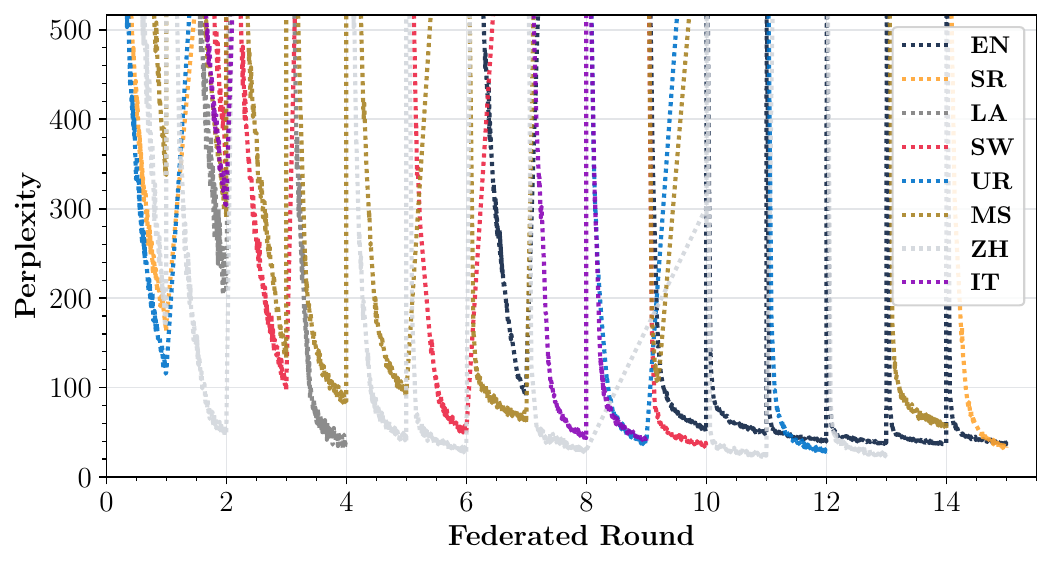}
    \caption{Convergence plot of our $1.3$ billion model trained in a vocabulary agnostic federated fashion. For the initial rounds, we sample $4$ data sources out of $8$; after seeing most of the clients, we reduce the number to $2$. We make sure only to introduce \texttt{EN} later into the experiment. }
    \label{app:big_model}
\end{figure}

Figure~\ref{app:big_model} provides further insights into the performance of \dept on a larger-scale experiment with a $1.3$ billion-parameter model. In this setting, the model is trained in a vocabulary-agnostic, federated fashion with dynamic client subsampling. During the initial rounds, $4$ out of $8$ data sources are sampled, which is reduced to $2$ after most clients have been processed. Importantly, \texttt{EN} is introduced later in the training process to evaluate the model's cross-lingual transfer capabilities to this high-resource language. The plot illustrates that the transformer body, enabled by \dept, effectively transfers knowledge across languages and domains, allowing newly introduced or previously stale data sources to converge to perplexity levels similar to their peers within one or two sampling rounds. This experiment underscores the feasibility and scalability of using \dept for collaborative large-scale language model pre-training, even under extreme client subsampling and without prior knowledge of the underlying data distribution.

\subsection{Plasticity}

\begin{figure}[H]
    \centering
    \subfloat[\texttt{MC4-FULL}, $12$-block.]{\includegraphics[width=0.485\textwidth]{plots/DEPT_IMPROVE_125M_MC4-B_ppl}}\hfill
    \subfloat[\texttt{SW}, $12$-block.]{\includegraphics[width=0.485\textwidth]{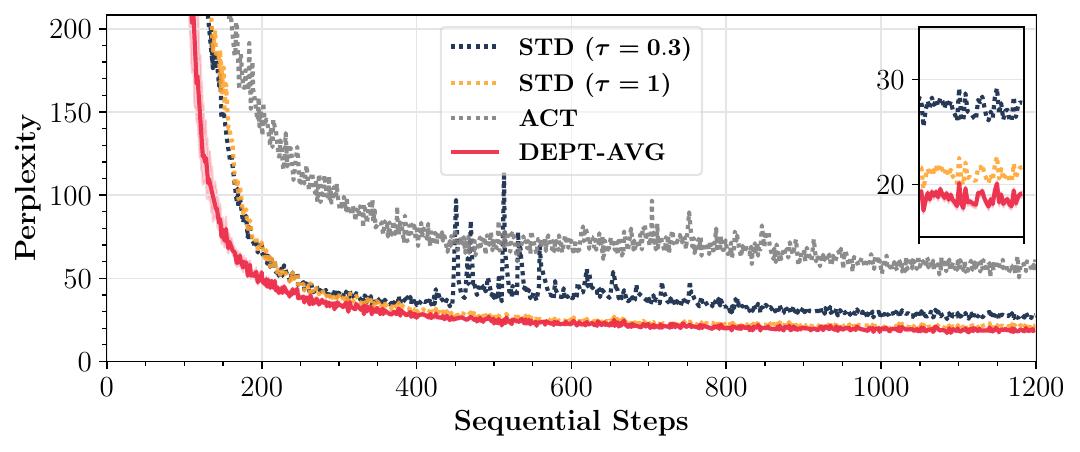}}\hfill \\
      \subfloat[\texttt{HI}, $12$-block.]{\includegraphics[width=0.485\textwidth]{plots/DEPT_IMPROVE_HI_125M_MC4-HI_ppl.pdf}}\hfill
    \subfloat[\texttt{DE}, $12$-block.]{\includegraphics[width=0.485\textwidth]{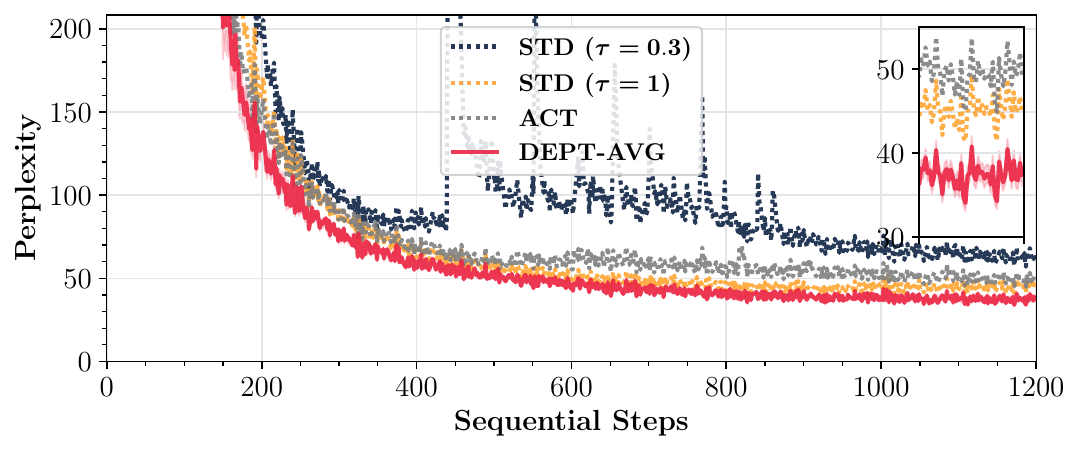}}\hfill
    \caption{Adaptation curves starting from a randomly initialized matrix. \dept is always stable in its convergence, reaching the \textbf{lowest perplexity} for the pre-training distribution~(\texttt{MC4-FULL}), for the lowest-resource languages in the distribution~(\texttt{SW}), and for the two out-of-distribution languages~(\texttt{HI}, \texttt{DE}). It is also always the \textbf{fastest} to adapt.}
    \label{fig:fed:mc4:125M:balanced:perplexity_ratio_full}
\end{figure} 

The results presented in Figure~\ref{fig:fed:mc4:125M:balanced:perplexity_ratio_full} demonstrate the robustness and adaptability of \dept across various settings, completing the plot shown in \cref{fig:fed:mc4:125M:balanced:perplexity_ratio}. Specifically, \dept consistently achieves the lowest perplexity across all scenarios: (1) the full pre-training distribution (\texttt{MC4-FULL}), (2) the lowest-resource language within the dataset (\texttt{SW}), and (3) two out-of-distribution languages (\texttt{HI} and \texttt{DE}). Furthermore, \dept is not only effective in reaching convergence but also does so at a faster rate compared to other approaches. These results showcase its utility in a wide range of multilingual and domain-adaptive pre-training tasks; for example, if a new client were to be introduced in a federated setting, they show that the \dept trained model could quickly adapt to its data distribution. Alternatively, multi-phase adaptive pre-training represents a distinct advantage in terms of data efficiency.

\subsection{IID Data Performance}\label{app:IID}

\begin{figure}[H]
    \centering

    \subfloat[\texttt{C4}, 12-block, Perplexity]{\includegraphics[width=0.485\textwidth]{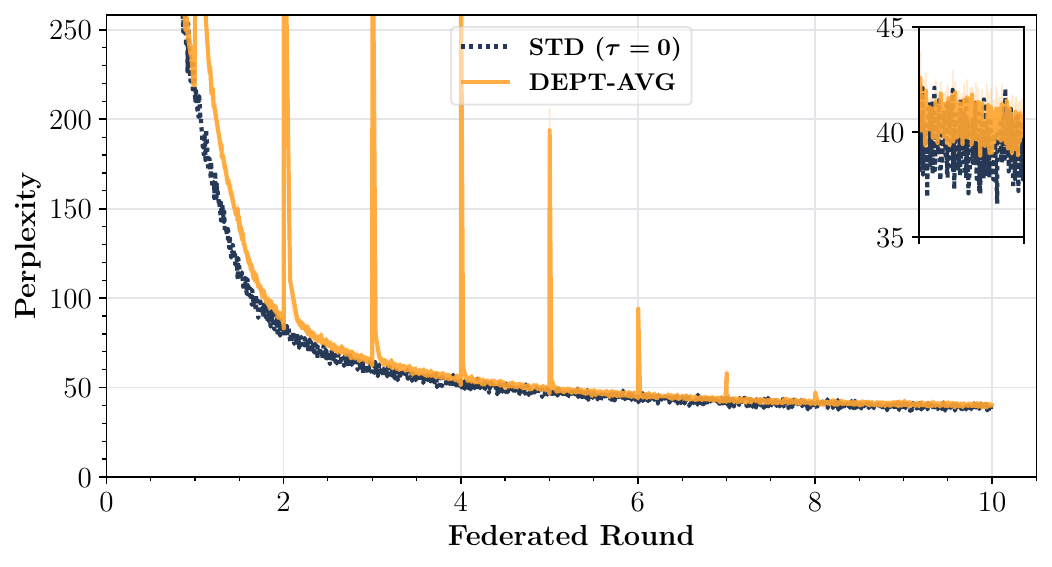}}\hfill
    \subfloat[\texttt{C4}, 12-block, Activations]{\includegraphics[width=0.495\textwidth]{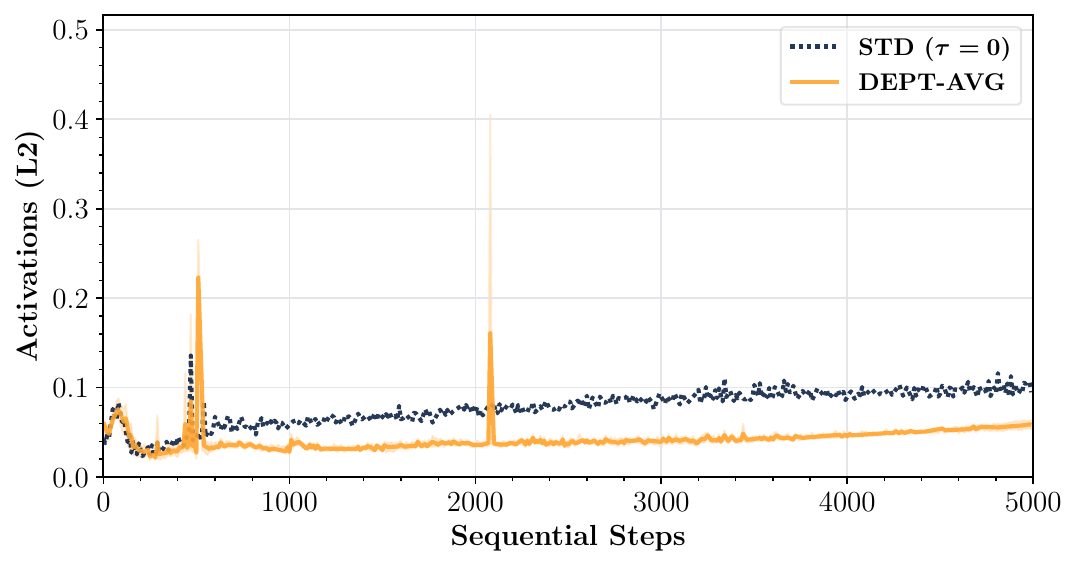}}\hfill
    \caption{Perplexity~(a) and activations~(b) curves for \dept versus uniform sampling on the IID \texttt{C4} dataset. \dept models, outside temporary spikes caused by \texttt{OuterOpt}, perform similarly to standard pre-training regarding training perplexity. However, as seen from the activations, it still provides greater training stability with the potential of extending pre-training.}
    \label{fig:fed:iid}
\end{figure}

In the case of IID data (represented by a random sharding of the \texttt{C4} dataset), \cref{fig:fed:iid} shows that \dept performs similarly to standard pre-training with the benefit of lower activation norms, indicating the potential for longer and more training.
\subsection{One-shot Generalization}
\input{tables/the_pile_350M-r}

\subsection{Transformer Body Generalization}

Table~\ref{tab:the_pile_350M_r} shows the performance of \dept on the \texttt{The Pile} dataset with a $24$-block model trained from randomly initialized embeddings. Here, \dept outperforms all baselines across all subsets, with average improvements of $17.5\%$.  

\FloatBarrier
\subsection{Pre-trained Embedding Matrix Generalization}

\input{tables/the_pile_350M-m}

When continuing pre-training with pre-trained embedding matrices, as shown in Table~\ref{tab:the_pile_350M_m}, \dept secures $10$ out of $17$ wins, with \trim consistently outperforming \glob. A comparison with Tables~\ref{tab:the_pile_350M_r} and~\ref{tab:the_pile_350M_pm} reveals that \dept also outperforms in other scenarios, whether starting from random embeddings or leveraging pre-trained ones. This consistency underscores the robustness of \dept's transformer body across varying embedding initialization and sampling strategies.
\FloatBarrier
\subsection{Scaling Experiments}

\input{tables/the_pile_125M-r}

Here, we train smaller multi-domain models with $12$ blocks to validate the scaling properties of \dept across model sizes. In \cref{tab:the_pile_125M_r}, we observe that, similar to \cref{tab:the_pile_350M_r}, \dept models outperform all baselines significantly when starting from random initialization. Importantly, the embeddings constitute a larger percentage of the model parameters at this model scale. This highlights the robustness of \dept's modifications to the embedding space in enabling the training of a better transformer body.

Interestingly, when using pre-trained embeddings (\cref{tab:mc4_125M_m}), the smaller \dept models perform worse than their larger counterparts. We speculate that the amount of local per-source training performed by \dept prior to \texttt{OuterOpt} should scale with model size. At this scale, the aggregation procedure may be overly harsh on the embedding parameters, particularly for the \glob and \trim configurations. This suggests that careful adjustments to the aggregation procedure may be necessary to maintain \dept's effectiveness at smaller model scales.

\input{tables/the_pile_125M-m}

\subsection{Comparison Against Single-client Models}

\FloatBarrier

\input{tables/the_pile_350M-pr-client-baselines}
\input{tables/mc4_125M-r-cent-baselines}

To study the impact of model averaging, we now compare \dept-based models with models trained on isolated data sources that are never averaged/merged. For fair comparisons, the model of each data source has seen as many tokens as it would have as a component in \dept-based training and has undergone continued pre-training for the same number of steps with access to the full dataset. Additionally, \dept models have undergone continued pre-training from random initialization. If we had compared against such models without the continued pre-training phase, they would have dominated on their respective data source while losing all other comparisons, especially in the case of multilingual data.

\textbf{\cref{tab:the_pile_350M_pr_cent_baselines,tab:mc4_125M_r_cent_baselines}} show how \dept models perform when all participants start from random initialization. Since the models trained on isolated data sources do not get to keep their highly specialized embeddings, this comparison evaluates how generalizable the abstractions learnt by the transformer body are across datasets. In the case of \texttt{The Pile}, shown in \cref{tab:the_pile_350M_pr_cent_baselines}, \dept models outperform the isolated baselines by $7.8\%$ in terms of average perplexity. Crucially, \dept models win all comparisons even though isolated baselines get evaluated on their pre-training dataset, indicating that they have not learned superior abstractions even in this case. For \texttt{MC4}, \cref{tab:mc4_125M_r_cent_baselines} show a very similar trend with a much higher degree of outperformance for \dept, $9.8\%$ on average on in-distribution data and $16.7\%$ for out-of-distribution (OOD) data, likely because the transformer body learned for one language has significant difficulty in adapting to a multilingual context.

\textbf{\cref{tab:the_pile_350M_pm_cent_baselines,tab:mc4_125M_m_cent_baselines}} show the impact of keeping the pre-trained embeddings before continued pre-training. The impact of this change is as expected: embeddings pre-trained on a specific dataset perform well on that dataset; however, they fail to generalize. In the case of \texttt{The Pile}, shown in \cref{tab:the_pile_350M_pm_cent_baselines}, \dept loses most comparisons to the baseline trained on a given dataset; however, it outperforms in terms of average perplexity by a remarkable $27\%$. For \texttt{MC4}, shown in \cref{tab:mc4_125M_m_cent_baselines}, the outperformance in terms of average perplexity is even more significant, $30.6\%$ for in-distribution data and $14.9\%$ of OOD data.

\input{tables/the_pile_350M-pm-client-baselines}

\input{tables/mc4_125M-m-cent-baselines}

\FloatBarrier
\subsection{Comparison Against Pythia}

\input{tables/the_pile_125M-m_pythia-baseline}

The experiments in our work are designed to investigate the outlined research questions instead of producing a state-of-the-art model. However, we believe that providing a comparison against a standard baseline may help better contextualize the performance of a given \dept model. For this purpose, we chose \texttt{Pythia}~\citep{Pythia} as it shares a very similar architecture to the one used in our work, with the only exception being that \texttt{Pythia} uses untied weights for the embedding matrix and thus all its equivalent sizes have more model parameters. \texttt{Pythia} models are trained on one epoch of the entire \texttt{The Pile}, $300$B tokens, regardless of size. Thus, we do not have any OOD dataset for them, and they are expected to perform better on \texttt{Ubuntu IRC}. Since they are trained for many more tokens than \dept models, we do not perform additional continued pre-training when starting from a pre-trained embedding matrix (the extra tokens existed to equalize the amount of work done across baselines); thus, those comparisons show the raw performance of \texttt{Pythia} as published by its authors. When starting from a random initialization we use the standard procedure from above.

\textbf{\cref{tab:the_pile_125M_m_pythia}} shows a comparison between a $160$M \texttt{Pythia} model and the $125$M \dept models when starting from pre-trained embeddings. At this scale, the additional pre-training of \texttt{Pythia} (using $30\times$ the tokens of \dept) does not provide an evident advantage as the model capacity is insufficient to benefit from it. Thus, outside of the expected outperformance on \texttt{Ubuntu IRC (UI)}, \texttt{Pythia-160M} performs similarly to \dept models and is slightly outperformed on average. We also speculate that using the full $22$-dataset version of \texttt{The Pile} during pre-training likely reduced the performance of  \texttt{Pythia-160M} as it had to fit a broader data distribution. We do not provide random initialization results for this model size since we found it impossible to make it behave well during continued pre-training, and we believe the comparison would be unfair.

\input{tables/the_pile_350M-pr-pythia-baselines}
\input{tables/the_pile_350M-pm-pythia-baseline}

\textbf{\cref{tab:the_pile_350M_pr_pythia_baseline,tab:the_pile_350M_pm_pythia_baseline}} show the expected outperformance of the $410$M \texttt{Pythia} model over the \dept models, as this size has sufficient capacity to benefit from the extensive ($10\times$ longer compared to \dept) pre-training. When starting from a random initialization, \cref{tab:the_pile_350M_pr_pythia_baseline}, the best \dept variant is within $1$ average perplexity point of \texttt{Pythia-410M}, indicating that a large portion of the additional token budget is primarily used to obtain better embeddings without providing a significantly improved transformer body. When starting from pre-initialized embeddings, \cref{tab:the_pile_350M_pm_pythia_baseline}, \texttt{Pythia-410M} significantly outperforms \dept achieving an average perplexity $10$ points lower than the best \dept variant. As discussed above, this is driven by its more extensive pre-training and improved embeddings. 

\FloatBarrier\FloatBarrier

%% file: tables/the_pile_350M-r.tex
\begin{table}[ht]
\caption{ Validation perplexity ($\downarrow$) for our $24$-block models trained on \texttt{The Pile} when using continued pre-training with uniform sampling starting from randomly-initialized embeddings. \dept provides a better transformer body for \textbf{all} datasets, outperforming baselines by $17.5\%$ on average.}\label{tab:the_pile_350M_r} 
\resizebox{\textwidth}{!}{
\begin{tabular}{@{}lcccccccccccccccccc@{}}
\toprule
\textbf{\begin{tabular}[c]{@{}l@{}}Name\\ ({\scriptsize UNIGRAM-CE})\end{tabular}} & \textbf{\begin{tabular}[c]{@{}c@{}}DM\\ ($6.9$)\end{tabular}} & \textbf{\begin{tabular}[c]{@{}c@{}}EN\\ ($7.9$)\end{tabular}} & \textbf{\begin{tabular}[c]{@{}c@{}}EP\\ ($10$)\end{tabular}} & \textbf{\begin{tabular}[c]{@{}c@{}}FL\\ ($7.8$)\end{tabular}} & \textbf{\begin{tabular}[c]{@{}c@{}}GH\\ ($7.9$)\end{tabular}} & \textbf{\begin{tabular}[c]{@{}c@{}}CC\\ ($7.9$)\end{tabular}} & \textbf{\begin{tabular}[c]{@{}c@{}}PA\\ ($8.2$)\end{tabular}} & \textbf{\begin{tabular}[c]{@{}c@{}}EN\\ ($7.7$)\end{tabular}} & \textbf{\begin{tabular}[c]{@{}c@{}}PP\\ ($9.1$)\end{tabular}} & \begin{tabular}[c]{@{}c@{}}WK\\ ($8.2$)\end{tabular} & \textbf{\begin{tabular}[c]{@{}c@{}}AX\\ ($7.7$)\end{tabular}} & \textbf{\begin{tabular}[c]{@{}c@{}}UB\\ ($7.8$)\end{tabular}} & \textbf{\begin{tabular}[c]{@{}c@{}}PC\\ ($8$)\end{tabular}} & \textbf{\begin{tabular}[c]{@{}c@{}}NH\\ ($8.1$)\end{tabular}} & \textbf{\begin{tabular}[c]{@{}c@{}}GU\\ ($7.7$)\end{tabular}} & \textbf{\begin{tabular}[c]{@{}c@{}}HN \\ ($7.7$) \end{tabular}} & \textbf{\begin{tabular}[c]{@{}c@{}}UI-OOD\\ ($10$)\end{tabular}} & \textbf{\begin{tabular}[c]{@{}c@{}}AVG\\ ($8.1$)\end{tabular}} \\ \midrule
\textbf{\texttt{STD}} ($\tau=0$) & $4.7$ & $17.5$ & $19.7$ & $22.3$ & $7$ & $64.2$ & $30.7$ & $18.3$ & $48.5$ & $41.6$ & $13$ & $23.9$ & $19$ & $32.1$ & $50.5$ & $42$ & $72.9$ & $31.1$ \\
\textbf{\texttt{STD}} ($\tau=1$) & $5.1$ & $24.1$ & $27.1$ & $31.4$ & $9.2$ & $86.6$ & $43.4$ & $24.6$ & $64.4$ & $58.4$ & $16.5$ & $32.7$ & $25.1$ & $44.7$ & $66.7$ & $55.8$ & $141$ & $44.5$ \\
\textbf{\act} & $-$ & $-$ & $-$ & $-$ & $-$ & $-$ & $-$ & $-$ & $-$ & $-$ & $-$ & $-$ & $-$ & $-$ & $-$ & $-$ & $-$ & $-$ \\ \midrule
\textbf{\glob} & $\mathbf{4.5}$ & $\mathbf{14.2}$ & $16.3$ & $\mathbf{18}$ & $\mathbf{6.1}$ & $\mathbf{53.6}$ & $\mathbf{24.9}$ & $\mathbf{15.5}$ & $\mathbf{40.2}$ & $\mathbf{34}$ & $\mathbf{11.2}$ & $\mathbf{19.7}$ & $\mathbf{16}$ & $\mathbf{26.2}$ & $\mathbf{41.9}$ & $\mathbf{34.6}$ & $\mathbf{58.8}$ & $\mathbf{25.6}$ \\
\textbf{\trim} & $4.5$ & $14.8$ & $16.7$ & $19.1$ & $6.4$ & $56.6$ & $26.4$ & $16.3$ & $42.1$ & $36$ & $11.7$ & $21$ & $16.9$ & $27.7$ & $43.7$ & $36.2$ & $66.1$ & $27.2$ \\
\textbf{\spec} & $4.5$ & $14.5$ & $\mathbf{16.2}$ & $18.8$ & $6.2$ & $55.5$ & $25.8$ & $16$ & $41.1$ & $35.1$ & $11.5$ & $20.5$ & $16.5$ & $27.2$ & $43.1$ & $35.7$ & $63.5$ & $26.6$ \\
\textbf{\texttt{SPEC-OPT}} & $4.6$ & $15.2$ & $16.9$ & $19.4$ & $6.4$ & $57.1$ & $26.5$ & $16.4$ & $42.5$ & $35.9$ & $11.9$ & $21$ & $16.5$ & $27.8$ & $44.9$ & $37.1$ & $60.4$ & $27.1$ \\ \midrule
\textbf{Min Imp (\%)} & $\mathbf{2.5}$ & $\mathbf{13.5}$ & $\mathbf{14.3}$ & $\mathbf{12.9}$ & $\mathbf{8.4}$ & $\mathbf{11.2}$ & $\mathbf{13.9}$ & $\mathbf{10.4}$ & $\mathbf{12.5}$ & $\mathbf{13.7}$ & $\mathbf{8.2}$ & $\mathbf{12.3}$ & $\mathbf{11.5}$ & $\mathbf{13.4}$ & $\mathbf{11}$ & $\mathbf{11.8}$ & $\mathbf{9.2}$ & $\mathbf{12.5}$ \\
\textbf{Max Imp (\%)} & $\mathbf{4.8}$ & $\mathbf{19.1}$ & $\mathbf{18}$ & $\mathbf{19.3}$ & $\mathbf{13}$ & $\mathbf{16.5}$ & $\mathbf{19}$ & $\mathbf{15.3}$ & $\mathbf{17.1}$ & $\mathbf{18.3}$ & $\mathbf{13.5}$ & $\mathbf{17.6}$ & $\mathbf{16}$ & $\mathbf{18.6}$ & $\mathbf{17.1}$ & $\mathbf{17.6}$ & $\mathbf{19.4}$ & $\mathbf{17.5}$ \\ \bottomrule
\end{tabular}
}
\vspace{-0.3cm}
\end{table}

%% file: tables/the_pile_350M-m.tex
\begin{table}[ht]
\caption{Validation perplexity ($\downarrow$)  for our $24$-block models trained on \texttt{The Pile} when performing continued pre-training with uniform sampling starting from a pre-trained embedding matrix. \dept wins $10$ out of $17$ comparisons with \trim always outperforming \glob. When comparing against \cref{tab:the_pile_350M_r,tab:the_pile_350M_pm}, we can observe that \dept wins the complementary comparisons when starting from random embeddings or when using proportional sampling with the pre-trained embedding matrices. This indicates that baselines always have a \textbf{worse} transformer body, with sampling ratios heavily impacting the effectiveness of embeddings for a given dataset.}\label{tab:the_pile_350M_m} 
\resizebox{\textwidth}{!}{%
\begin{tabular}{@{}lcccccccccccccccccc@{}}
\toprule
\textbf{\begin{tabular}[c]{@{}l@{}}Name\\ ({\scriptsize UNIGRAM-CE})\end{tabular}} & \textbf{\begin{tabular}[c]{@{}c@{}}DM\\ ($6.9$)\end{tabular}} & \textbf{\begin{tabular}[c]{@{}c@{}}EE\\ ($7.9$)\end{tabular}} & \textbf{\begin{tabular}[c]{@{}c@{}}EP\\ ($10$)\end{tabular}} & \textbf{\begin{tabular}[c]{@{}c@{}}FL\\ ($7.8$)\end{tabular}} & \textbf{\begin{tabular}[c]{@{}c@{}}GH\\ ($7.9$)\end{tabular}} & \textbf{\begin{tabular}[c]{@{}c@{}}CC\\ ($7.9$)\end{tabular}} & \textbf{\begin{tabular}[c]{@{}c@{}}PA\\ ($8.2$)\end{tabular}} & \textbf{\begin{tabular}[c]{@{}c@{}}SE\\ ($7.7$)\end{tabular}} & \textbf{\begin{tabular}[c]{@{}c@{}}PP\\ ($9.1$)\end{tabular}} & \textbf{\begin{tabular}[c]{@{}c@{}}WK\\ ($8.2$)\end{tabular}} & \textbf{\begin{tabular}[c]{@{}c@{}}AX\\ ($7.7$)\end{tabular}} & \textbf{\begin{tabular}[c]{@{}c@{}}UB\\ ($7.8$)\end{tabular}} & \textbf{\begin{tabular}[c]{@{}c@{}}PC\\ ($8$)\end{tabular}} & \textbf{\begin{tabular}[c]{@{}c@{}}NH\\ ($8.1$)\end{tabular}} & \textbf{\begin{tabular}[c]{@{}c@{}}GU\\ ($7.7$)\end{tabular}} & \textbf{\begin{tabular}[c]{@{}c@{}}HN \\ ($7.7$) \end{tabular}} & \textbf{\begin{tabular}[c]{@{}c@{}}UI-OOD\\ ($10$)\end{tabular}} & \textbf{\begin{tabular}[c]{@{}c@{}}AVG\\ ($8.1$)\end{tabular}} \\ \midrule
\textbf{\texttt{STD}} ($\tau=0$) & $\mathbf{4.3}$ & $11.1$ & $13.4$ & $15$ & $5.5$ & $44.4$ & $20.4$ & $13.2$ & $34$ & $27.1$ & $10.1$ & $16.9$ & $13.6$ & $21.6$ & $35.2$ & $29.1$ & $51.6$ & $21.6$ \\
\textbf{\texttt{STD}} ($\tau=1$) & $4.3$ & $12.6$ & $15.8$ & $\mathbf{13.5}$ & $\mathbf{4.7}$ & $38.8$ & $19.2$ & $\mathbf{11.7}$ & $36.8$ & $24.7$ & $\mathbf{8.8}$ & $16$ & $\mathbf{12}$ & $21.7$ & $34.2$ & $28.6$ & $\mathbf{43.1}$ & $20.4$ \\ \midrule
\textbf{\glob} & $4.5$ & $12.1$ & $13.8$ & $15.6$ & $5.5$ & $41.4$ & $19.4$ & $12.9$ & $33.5$ & $25.6$ & $10.1$ & $15.6$ & $13.3$ & $19.8$ & $38.1$ & $29.2$ & $58.2$ & $21.7$ \\
\textbf{\trim} & $4.4$ & $\mathbf{10}$ & $\mathbf{11.3}$ & $14.8$ & $4.9$ & $\mathbf{37.3}$ & $\mathbf{18.2}$ & $11.8$ & $\mathbf{30.2}$ & $\mathbf{23}$ & $9.8$ & $\mathbf{15.3}$ & $12.8$ & $\mathbf{19}$ & $\mathbf{32.9}$ & $\mathbf{26.7}$ & $47.6$ & $\mathbf{19.4}$ \\ \midrule
\textbf{Min Imp (\%)} & $-4.5$ & $-8.6$ & $-2.9$ & $-15.6$ & $-17.8$ & $-6.8$ & $-0.8$ & $-9.6$ & $\mathbf{1.2}$ & $-3.5$ & $-14$ & $\mathbf{2}$ & $-10.2$ & $\mathbf{8.1}$ & $-11.5$ & $-2.2$ & $-35.1$ & $-6.4$ \\
\textbf{Max Imp (\%)} & $-1.2$ & $\mathbf{10.1}$ & $\mathbf{15.4}$ & $-9.5$ & $-5.1$ & $\mathbf{3.8}$ & $\mathbf{5.1}$ & $-0.3$ & $\mathbf{11.1}$ & $\mathbf{7.1}$ & $-10.9$ & $\mathbf{4.2}$ & $-6$ & $\mathbf{12.1}$ & $\mathbf{3.9}$ & $\mathbf{6.6}$ & $-10.4$ & $\mathbf{4.8}$ \\ \bottomrule
\end{tabular}%
}
\end{table}

%% file: tables/the_pile_125M-r.tex
\begin{table}[t]
\caption{ Validation perplexity ($\downarrow$)  for our $12$-block models trained on \texttt{The Pile} when performing continued pre-training starting from a randomly-initialized embedding matrix. \dept can train a superior transformer body, outperforming all baselines across all subsets by up to $28\%$. }\label{tab:the_pile_125M_r}
\resizebox{\textwidth}{!}{%
\begin{tabular}{@{}lcccccccccccccccccc@{}}
\toprule
\textbf{\begin{tabular}[c]{@{}l@{}}Name\\ ({\scriptsize UNIGRAM-CE})\end{tabular}} & \textbf{\begin{tabular}[c]{@{}c@{}}NH\\ ($8.1$)\end{tabular}} & \textbf{\begin{tabular}[c]{@{}c@{}}GH\\ ($7.9$)\end{tabular}} & \textbf{\begin{tabular}[c]{@{}c@{}}PA\\ ($8.2$)\end{tabular}} & \textbf{\begin{tabular}[c]{@{}c@{}}UB\\ ($7.8$)\end{tabular}} & \textbf{\begin{tabular}[c]{@{}c@{}}FL\\ ($7.8$)\end{tabular}} & \textbf{\begin{tabular}[c]{@{}c@{}}EE\\ ($7.9$)\end{tabular}} & \textbf{\begin{tabular}[c]{@{}c@{}}EP\\ ($10$)\end{tabular}} & \textbf{\begin{tabular}[c]{@{}c@{}}WK\\ ($8.2$)\end{tabular}} & \textbf{\begin{tabular}[c]{@{}c@{}}CC\\ ($7.9$)\end{tabular}}  & \textbf{\begin{tabular}[c]{@{}c@{}}SE\\ ($7.7$)\end{tabular}} & \textbf{\begin{tabular}[c]{@{}c@{}}PC\\ ($8$)\end{tabular}} & \textbf{\begin{tabular}[c]{@{}c@{}}PP\\ ($9.1$)\end{tabular}} & \textbf{\begin{tabular}[c]{@{}c@{}}DM\\ ($6.9$)\end{tabular}} & \textbf{\begin{tabular}[c]{@{}c@{}}AX\\ ($7.7$)\end{tabular}} & \textbf{\begin{tabular}[c]{@{}c@{}}GU\\ ($7.7$)\end{tabular}} & \textbf{\begin{tabular}[c]{@{}c@{}}HN \\ ($7.7$) \end{tabular}} & \textbf{\begin{tabular}[c]{@{}c@{}}UI-OOD\\ ($10$)\end{tabular}} & \textbf{\begin{tabular}[c]{@{}c@{}}AVG\\ ($8.1$)\end{tabular}}\\ \midrule
\textbf{\texttt{STD}} ($\tau=0$) & $63.0$ & $12.1$ & $61.9$ & $44.1$ & $45.2$ & $36.3$ & $47.9$ & $81.8$ & $115.9$ & $33.2$ & $32.4$ & $91.8$ & $5.8$ & $21.6$ & $91.2$ & $75.0$ & $198.1$ & $62.2$ \\
\textbf{\texttt{STD}} ($\tau=1$) & $58.6$ & $11.4$ & $57.2$ & $41.3$ & $42.2$ & $33.6$ & $43.1$ & $75.2$ & $108.2$ & $31.0$ & $30.3$ & $85.4$ & $5.7$ & $20.4$ & $85.6$ & $70.9$ & $168.0$ & $57.0$ \\
\textbf{\act} & $126.7$ & $20$ & $124.8$ & $79.9$ & $82.1$ & $66.3$ & $124.2$ & $147.7$ & $191.6$ & $55.4$ & $61.2$ & $180.1$ & $7.4$ & $33.8$ & $150.8$ & $123.9$ & $377.8$ & $114.9$ \\
\midrule
\textbf{\glob} & $44.5$ & $9.3$ & $43.0$ & $32.0$ & $32.1$ & $25.9$ & $31.4$ & $58.5$ & $83.9$ & $23.8$ & $23.1$ & $64.6$ & $\mathbf{5.1}$ & $16.2$ & $66.4$ & $54.7$ & $114.6$ & $42.9$ \\
\textbf{\trim} & $43.3$ & $8.8$ & $41.8$ & $31.2$ & $30.7$ & $24.6$ & $29.4$ & $56.2$ & $\mathbf{82.3}$ & $\mathbf{23.4}$ & $\mathbf{22.6}$ & $\mathbf{62.7}$ & $5.1$ & $\mathbf{16.0}$ & $\mathbf{64.1}$ & $\mathbf{53.2}$ & $\mathbf{99.0}$ & $\mathbf{40.8}$ \\
\textbf{\spec} & $\mathbf{42.1}$ & $\mathbf{8.7}$ & $\mathbf{40.6}$ & $\mathbf{30.3}$ & $\mathbf{29.8}$ & $\mathbf{23.8}$ & $\mathbf{28.0}$ & $\mathbf{54.8}$ & $87.0$ & $24.9$ & $23.8$ & $67.5$ & $5.2$ & $16.8$ & $69.1$ & $57.1$ & $124.2$ & $43.2$ \\
\midrule
\textbf{Min Imp (\%)} & $\mathbf{24}$ & $\mathbf{19}$ & $\mathbf{25}$ & $\mathbf{23}$ & $\mathbf{24}$ & $\mathbf{23}$ & $\mathbf{27}$ & $\mathbf{22}$ & $\mathbf{20}$ & $\mathbf{20}$ & $\mathbf{21}$ & $\mathbf{21}$ & $\mathbf{8}$ & $\mathbf{18}$ & $\mathbf{19}$ & $\mathbf{20}$ & $\mathbf{26}$ & $\mathbf{24}$ \\
\textbf{Max Imp (\%)} & $\mathbf{28}$ & $\mathbf{24}$ & $\mathbf{29}$ & $\mathbf{27}$ & $\mathbf{29}$ & $\mathbf{29}$ & $\mathbf{35}$ & $\mathbf{27}$ & $\mathbf{24}$ & $\mathbf{25}$ & $\mathbf{25}$ & $\mathbf{27}$ & $\mathbf{11}$ & $\mathbf{22}$ & $\mathbf{25}$ & $\mathbf{25}$ & $\mathbf{41}$ & $\mathbf{28}$ \\
\bottomrule
\end{tabular}%
}
\end{table}

%% file: tables/the_pile_125M-m.tex
\begin{table}[t]
\caption{ Validation perplexity ($\downarrow$)  for our $12$-block models trained on \texttt{The Pile} when performing continued pre-training starting from a pre-trained embedding matrix. \dept performs worse than for the $24$-block trained on \texttt{The Pile} and than for our $MC4$ models. However, when considering \cref{tab:the_pile_125M_r}, we can observe it wins all comparisons by wide margins when starting from a randomly initialized embedding matrix, indicating that this gap is driven by the embedding space being fitted to the high-resource languages despite the baselines having a \textbf{worse} transformer body.}\label{tab:the_pile_125M_m}
\resizebox{\textwidth}{!}{%
\begin{tabular}{lcccccccccccccccccc}
\toprule
\textbf{\begin{tabular}[c]{@{}l@{}}Name\\ ({\scriptsize UNIGRAM-CE})\end{tabular}} & \textbf{\begin{tabular}[c]{@{}c@{}}NH\\ ($8.1$)\end{tabular}} & \textbf{\begin{tabular}[c]{@{}c@{}}GH\\ ($7.9$)\end{tabular}} & \textbf{\begin{tabular}[c]{@{}c@{}}PA\\ ($8.2$)\end{tabular}} & \textbf{\begin{tabular}[c]{@{}c@{}}UB\\ ($7.8$)\end{tabular}} & \textbf{\begin{tabular}[c]{@{}c@{}}FL\\ ($7.8$)\end{tabular}} & \textbf{\begin{tabular}[c]{@{}c@{}}EE\\ ($7.9$)\end{tabular}} & \textbf{\begin{tabular}[c]{@{}c@{}}EP\\ ($10$)\end{tabular}} & \textbf{\begin{tabular}[c]{@{}c@{}}WK\\ ($8.2$)\end{tabular}} & \textbf{\begin{tabular}[c]{@{}c@{}}CC\\ ($7.9$)\end{tabular}}  & \textbf{\begin{tabular}[c]{@{}c@{}}SE\\ ($7.7$)\end{tabular}} & \textbf{\begin{tabular}[c]{@{}c@{}}PC\\ ($8$)\end{tabular}} & \textbf{\begin{tabular}[c]{@{}c@{}}PP\\ ($9.1$)\end{tabular}} & \textbf{\begin{tabular}[c]{@{}c@{}}DM\\ ($6.9$)\end{tabular}} & \textbf{\begin{tabular}[c]{@{}c@{}}AX\\ ($7.7$)\end{tabular}} & \textbf{\begin{tabular}[c]{@{}c@{}}GU\\ ($7.7$)\end{tabular}} & \textbf{\begin{tabular}[c]{@{}c@{}}HN \\ ($7.7$) \end{tabular}} & \textbf{\begin{tabular}[c]{@{}c@{}}UI-OOD\\ ($10$)\end{tabular}} & \textbf{\begin{tabular}[c]{@{}c@{}}AVG\\ ($8.1$)\end{tabular}} \\ \midrule
\textbf{\texttt{STD}} ($\tau=0$) & $31.9$ & $7.6$ & $30.5$ & $24.2$ & $23.2$ & $\mathbf{18.1}$ & $\mathbf{20.5}$ & $41.2$ & $64.6$ & $18.9$ & $19.2$ & $49.4$ & $\mathbf{4.9}$ & $13.6$ & $51.7$ & $42.6$ & $68.4$ & $31.2$ \\
\textbf{\texttt{STD}} ($\tau=1$) & $32.7$ & $\mathbf{6.6}$ & $29.5$ & $23.4$ & $\mathbf{21.3}$ & $20.8$ & $27.8$ & $\mathbf{38.4}$ & $\mathbf{57.9}$ & $\mathbf{17.0}$ & $\mathbf{17.2}$ & $56.8$ & $5.0$ & $\mathbf{12.0}$ & $\mathbf{51.2}$ & $42.8$ & $81.8$ & $31.9$ \\
\midrule
\textbf{\glob} & $30.2$ & $7.1$ & $29.8$ & $22.9$ & $23.7$ & $20.0$ & $21.9$ & $41.6$ & $61.8$ & $17.9$ & $19.6$ & $48.0$ & $5.2$ & $13.7$ & $54.1$ & $42.2$ & $90.7$ & $32.4$ \\
\textbf{\trim} & $\mathbf{29.5}$ & $6.9$ & $\mathbf{29.2}$ & $\mathbf{22.4}$ & $23.0$ & $19.4$ & $21.1$ & $40.8$ & $60.6$ & $17.4$ & $19.2$ & $\mathbf{46.7}$ & $5.1$ & $13.4$ & $52.4$ & $\mathbf{41.0}$ & $\mathbf{81.1}$ & $\mathbf{31.1}$ \\
\midrule
\textbf{Min Imp (\%)} & $\mathbf{5.4}$ & $\mathbf{-7.0}$ & $-1.2$ & $2.0$ & $-11.2$ & $-10.5$ & $-6.7$ & $-8.3$ & $-6.8$ & $-5.1$ & $-13.8$ & $\mathbf{3.0}$ & $-5.5$ & $-14.3$ & $-5.6$ & $\mathbf{0.9}$ & $-32.7$ & $-3.7$ \\
\textbf{Max Imp (\%)} & $\mathbf{7.5}$ & $\mathbf{-4.0}$ & $1.0$ & $4.3$ & $-7.9$ & $-7.5$ & $-2.9$ & $-6.2$ & $-4.7$ & $-2.6$ & $-11.4$ & $\mathbf{5.6}$ & $-4.2$ & $-11.5$ & $-2.3$ & $\mathbf{3.8}$ & $-18.6$ & $\mathbf{0.3}$ \\
\bottomrule
\end{tabular}%
}
\end{table}

%% file: tables/the_pile_350M-pr-client-baselines.tex
\begin{table}[ht]
\caption{ Validation perplexity ($\downarrow$) for $24$-block models trained on \texttt{The Pile} after \textbf{continued pre-training} with \textbf{proportional} sampling from \textbf{randomly-initialized} embeddings, compared to models which had been pre-trained on a single data source for the same total number of tokens as \dept has seen from their distributions. \dept outperforms all baselines. \dept outperforms all baselines. Baselines whose pre-training dataset matches the evaluation dataset are highlighted in olive.}\label{tab:the_pile_350M_pr_cent_baselines} 
\centering
\resizebox{\textwidth}{!}{%
\begin{tabular}{@{}lcccccccccccccccccc@{}}
\toprule
\textbf{\begin{tabular}[c]{@{}l@{}}Name\\ ({\scriptsize UNIGRAM-CE})\end{tabular}} & \textbf{\begin{tabular}[c]{@{}c@{}}DM\\ ($6.9$)\end{tabular}} & \textbf{\begin{tabular}[c]{@{}c@{}}EE\\ ($7.9$)\end{tabular}} & \textbf{\begin{tabular}[c]{@{}c@{}}EP\\ ($10$)\end{tabular}} & \textbf{\begin{tabular}[c]{@{}c@{}}FL\\ ($7.8$)\end{tabular}} & \textbf{\begin{tabular}[c]{@{}c@{}}GH\\ ($7.9$)\end{tabular}} & \textbf{\begin{tabular}[c]{@{}c@{}}CC\\ ($7.9$)\end{tabular}} & \textbf{\begin{tabular}[c]{@{}c@{}}PA\\ ($8.2$)\end{tabular}} & \textbf{\begin{tabular}[c]{@{}c@{}}SE\\ ($7.7$)\end{tabular}} & \textbf{\begin{tabular}[c]{@{}c@{}}PP\\ ($9.1$)\end{tabular}} & \textbf{\begin{tabular}[c]{@{}c@{}}WK\\ ($8.2$)\end{tabular}} & \textbf{\begin{tabular}[c]{@{}c@{}}AX\\ ($7.7$)\end{tabular}} & \textbf{\begin{tabular}[c]{@{}c@{}}UB\\ ($7.8$)\end{tabular}} & \textbf{\begin{tabular}[c]{@{}c@{}}PC\\ ($8$)\end{tabular}} & \textbf{\begin{tabular}[c]{@{}c@{}}NH\\ ($8.1$)\end{tabular}} & \textbf{\begin{tabular}[c]{@{}c@{}}GU\\ ($7.7$)\end{tabular}} & \textbf{\begin{tabular}[c]{@{}c@{}}HN \\ ($7.7$)\end{tabular}} & \textbf{\begin{tabular}[c]{@{}c@{}}UI-OOD\\ ($10$)\end{tabular}} & \textbf{\begin{tabular}[c]{@{}c@{}}AVG\\ ($8.1$)\end{tabular}} \\ \midrule
\textbf{\texttt{CC}} & $4.8$ & $29.3$ & $44.4$ & $20$ & $5.9$ & {\color{olive}$54.4$} & $30$ & $16.4$ & $77.3$ & $37.7$ & $10.8$ & $23.2$ & $15.8$ & $38$ & $52$ & $44.7$ & $109.2$ & $36.1$ \\
\textbf{\texttt{PC}} & $4.8$ & $28.1$ & $40.3$ & $19.1$ & $5.7$ & $52.1$ & $27.9$ & $15.7$ & $72.8$ & $35.8$ & $10.4$ & $21.9$ & {\color{olive}$14.8$} & $35.5$ & $50.3$ & $43.4$ & $110.6$ & $34.7$ \\
\textbf{\texttt{AX}} & $4.9$ & $28.9$ & $41.7$ & $19.8$ & $5.7$ & $53.5$ & $29.1$ & $15.9$ & $74.4$ & $36.8$ & {\color{olive}$10.5$} & $22.5$ & $15.3$ & $36.8$ & $52.1$ & $44.8$ & $97.4$ & $34.7$ \\
\textbf{\texttt{GH}} & $4.9$ & $30.1$ & $43.6$ & $20.8$ & {\color{olive}$5.8$} & $55.9$ & $30.7$ & $16.5$ & $78.2$ & $38.5$ & $10.9$ & $23.8$ & $16$ & $38.9$ & $54.5$ & $46.7$ & $120.7$ & $37.4$ \\
\textbf{\texttt{FL}} & $4.9$ & $31.8$ & $49.8$ & {\color{olive}$21.5$} & $6.3$ & $58.5$ & $32.4$ & $17.4$ & $84.2$ & $40.9$ & $11.4$ & $24.8$ & $16.8$ & $41$ & $55.8$ & $47.9$ & $122.1$ & $39.3$ \\
\textbf{\texttt{SE}} & $4.8$ & $28.2$ & $42.2$ & $19.4$ & $5.6$ & $52.9$ & $29$ & {\color{olive}$15.5$} & $75$ & $36.6$ & $10.5$ & $22.4$ & $15.3$ & $36.8$ & $50.8$ & $43.4$ & $100$ & $34.6$ \\
\textbf{\texttt{WK}} & $4.8$ & $28.1$ & $42.1$ & $18.9$ & $5.7$ & $51.6$ & $28.4$ & $15.8$ & $74$ & {\color{olive}$34.8$} & $10.5$ & $21.9$ & $15.1$ & $35.7$ & $49.8$ & $43.2$ & $95.4$ & $33.9$ \\
\textbf{\texttt{DM}} & {\color{olive} $7.3$} & $140.4$ & $559.4$ & $100.6$ & $28$ & $239.5$ & $184$ & $71$ & $543.9$ & $213.9$ & $34.8$ & $121.7$ & $69.3$ & $213.9$ & $193$ & $170$ & $966.8$ & $226.9$ \\ \midrule
\textbf{\glob} & $4.8$ & $\mathbf{25.7}$ & $38.2$ & $\mathbf{17.3}$ & $5.4$ & $\mathbf{47.7}$ & $\mathbf{25.7}$ & $\mathbf{14.7}$ & $68.3$ & $\mathbf{32.7}$ & $\mathbf{9.9}$ & $\mathbf{20}$ & $\mathbf{14}$ & $\mathbf{32.2}$ & $\mathbf{46.5}$ & $\mathbf{39.8}$ & $94.8$ & $31.6$ \\
\textbf{\trim} & $4.8$ & $27.3$ & $39.5$ & $18.5$ & $5.6$ & $51.2$ & $27.8$ & $15.4$ & $71.8$ & $35.1$ & $10.3$ & $21.7$ & $14.8$ & $35.1$ & $49.1$ & $42.2$ & $95.7$ & $33.3$ \\
\textbf{\spec} & $4.8$ & $26.7$ & $36.8$ & $18.2$ & $5.5$ & $50.1$ & $27.1$ & $15.1$ & $69.1$ & $34.2$ & $10.1$ & $21.1$ & $14.5$ & $34.3$ & $48.5$ & $41.7$ & $97.6$ & $32.7$ \\
\textbf{\texttt{SPEC-OPT}} & $\mathbf{4.7}$ & $25.9$ & $\mathbf{35}$ & $17.5$ & $\mathbf{5.4}$ & $48.3$ & $26.1$ & $14.7$ & $\mathbf{66.6}$ & $32.8$ & $9.9$ & $20.4$ & $14.1$ & $32.9$ & $47.3$ & $40.5$ & $\mathbf{88.6}$ & $\mathbf{31.2}$ \\ \midrule
\textbf{Min Imp (\%)} & $\mathbf{0.7}$ & $\mathbf{2.7}$ & $\mathbf{2}$ & $\mathbf{1.6}$ & $\mathbf{0.5}$ & $\mathbf{0.8}$ & $\mathbf{0.5}$ & $\mathbf{0.3}$ & $\mathbf{1.4}$ & $-0.9$ & $\mathbf{1.2}$ & $\mathbf{1}$ & $\mathbf{0}$ & $\mathbf{1}$ & $\mathbf{1.3}$ & $\mathbf{2.1}$ & $-2.3$ & $\mathbf{1.7}$ \\
\textbf{Max Imp (\%)} & $\mathbf{1.2}$ & $\mathbf{8.3}$ & $\mathbf{13.2}$ & $\mathbf{8.4}$ & $\mathbf{4.3}$ & $\mathbf{7.6}$ & $\mathbf{7.9}$ & $\mathbf{5.3}$ & $\mathbf{8.5}$ & $\mathbf{6}$ & $\mathbf{5}$ & $\mathbf{8.8}$ & $\mathbf{5.6}$ & $\mathbf{9.2}$ & $\mathbf{6.5}$ & $\mathbf{7.8}$ & $\mathbf{7.1}$ & $\mathbf{7.8}$ \\ \bottomrule
\end{tabular}%
}
\end{table}

%% file: tables/mc4_125M-r-cent-baselines.tex
\begin{table}[t]
\caption{Validation perplexity ($\downarrow$) for $12$-block models trained on \texttt{MC4} after \textbf{continued pre-training} with \textbf{unfiorm} sampling from \textbf{randomly-initialized} embeddings, compared to models which had been pre-trained on a single data source for the same total number of tokens as \dept has seen from their distributions. Baselines whose pre-training dataset matches the evaluation dataset are highlighted in olive.}\label{tab:mc4_125M_r_cent_baselines}
\resizebox{\textwidth}{!}{%
\begin{tabular}{@{}lccccccccccccc@{}}
\toprule
 & \multicolumn{9}{c}{\textbf{In-Distribution}} & \multicolumn{4}{c}{\textbf{Out-of-Distribution}} \\ \cmidrule(l){2-10}  \cmidrule(l){11-14} 
\toprule
 & \multicolumn{9}{c}{\textbf{In-Distribution}} & \multicolumn{4}{c}{\textbf{Out-of-Distribution}} \\ \cmidrule(l){2-14} 
\textbf{\begin{tabular}[c]{@{}l@{}}Name\\ ({\scriptsize UNIGRAM-CE})\end{tabular}} & \textbf{\begin{tabular}[c]{@{}c@{}}ZH\\ ($9.8$)\end{tabular}} & \textbf{\begin{tabular}[c]{@{}c@{}}UR\\ ($10.5$)\end{tabular}} & \textbf{\begin{tabular}[c]{@{}c@{}}MS\\ ($9.2$)\end{tabular}} & \textbf{\begin{tabular}[c]{@{}c@{}}IT\\ ($7.7$)\end{tabular}} & \textbf{\begin{tabular}[c]{@{}c@{}}SR\\ ($10.5$)\end{tabular}} & \textbf{\begin{tabular}[c]{@{}c@{}}LA\\ ($9$)\end{tabular}} & \textbf{\begin{tabular}[c]{@{}c@{}}EN\\ ($7.5$)\end{tabular}} & \textbf{\begin{tabular}[c]{@{}c@{}}SW\\ ($10$)\end{tabular}} & \textbf{\begin{tabular}[c]{@{}c@{}}Avg (In-D)\\ ($9.3$)\end{tabular}} & \textbf{\begin{tabular}[c]{@{}c@{}}EL\\ ($14.4$)\end{tabular}} & \textbf{\begin{tabular}[c]{@{}c@{}}HI\\ ($13.9$)\end{tabular}} & \textbf{\begin{tabular}[c]{@{}c@{}}DE\\ ($9.7$)\end{tabular}} & \textbf{\begin{tabular}[c]{@{}c@{}}Avg (OOD)\\ ($12.6$)\end{tabular}} \\ \midrule
\textbf{\texttt{ZH}} & { \color{olive}$187.8$} & $44.6$ & $113.9$ & $98.6$ & $89.1$ & $73.9$ & $128.8$ & $76$ & $101.6$ & $5744.8$ & $6476.5$ & $1448$ & $4556.4$ \\
\textbf{\texttt{UR}} & $94.8$ & { \color{olive}$27.5$} & $66$ & $56.9$ & $48.7$ & $42.5$ & $79.3$ & $44.1$ & $57.5$ & $2596.5$ & $2371.1$ & $690.5$ & $1886$ \\
\textbf{\texttt{MS}} & $78.8$ & $24.8$ & { \color{olive}$58$} & $50$ & $42.7$ & $37.3$ & $70$ & $38.9$ & $50.1$ & $2673.3$ & $2329.2$ & $599.4$ & $1867.3$ \\
\textbf{\texttt{IT}} & $81.3$ & $25.1$ & $59.7$ & { \color{olive}$51$} & $43.8$ & $38.2$ & $71.8$ & $39.8$ & $51.3$ & $2617.2$ & $2256.9$ & $615.3$ & $1829.8$ \\
\textbf{\texttt{SR}} & $85.4$ & $25.7$ & $61$ & $52.4$ & { \color{olive}$44.9$} & $39.4$ & $73.7$ & $40.9$ & $52.9$ & $2992.4$ & $2648.3$ & $657.2$ & $2099.3$ \\
\textbf{\texttt{LA}} & $104.8$ & $29.6$ & $71.7$ & $60.6$ & $53.5$ & { \color{olive}$46$} & $85.2$ & $47.6$ & $62.4$ & $2838.7$ & $2824.8$ & $746.6$ & $2136.7$ \\
\textbf{\texttt{EN}} & $104.4$ & $30.1$ & $71.9$ & $61.2$ & $54.3$ & $46.2$ & { \color{olive}$85$} & $47.8$ & $62.6$ & $3344.4$ & $3360.6$ & $834.8$ & $2513.3$ \\
\textbf{\texttt{SW}} & $79.3$ & $24.7$ & $58.1$ & $49.9$ & $43$ & $37.3$ & $69.9$ & { \color{olive}$39$} & $50.1$ & $2552.3$ & $2067.5$ & $608.3$ & $1742.7$ \\ \midrule
\textbf{\glob} & $67.7$ & $\mathbf{22.4}$ & $\mathbf{53.7}$ & $\mathbf{46}$ & $\mathbf{38.6}$ & $\mathbf{33.9}$ & $\mathbf{65.4}$ & $\mathbf{35.2}$ & $\mathbf{45.4}$ & $2308.3$ & $1676.5$ & $559.5$ & $1514.7$ \\
\textbf{\trim} & $\mathbf{67.7}$ & $22.8$ & $55.2$ & $47.5$ & $39.7$ & $35.1$ & $67.2$ & $36.3$ & $46.4$ & $2547.7$ & $1911$ & $567.4$ & $1675.4$ \\
\textbf{\spec} & $69.5$ & $23$ & $55.4$ & $47.8$ & $40.3$ & $34.7$ & $68.1$ & $36.3$ & $46.9$ & $\mathbf{2232.1}$ & $\mathbf{1578.8}$ & $\mathbf{544.7}$ & $\mathbf{1451.9}$ \\ \midrule
\textbf{Min Imp (\%)} & $\mathbf{11.8}$ & $\mathbf{6.6}$ & $\mathbf{4.4}$ & $\mathbf{4.2}$ & $\mathbf{5.7}$ & $\mathbf{6}$ & $\mathbf{2.7}$ & $\mathbf{6.8}$ & $\mathbf{6.4}$ & $\mathbf{0.2}$ & $\mathbf{7.6}$ & $\mathbf{5.3}$ & $\mathbf{3.9}$ \\
\textbf{Max Imp (\%)} & $\mathbf{14.1}$ & $\mathbf{9.3}$ & $\mathbf{7.4}$ & $\mathbf{7.7}$ & $\mathbf{9.6}$ & $\mathbf{9.1}$ & $\mathbf{6.5}$ & $\mathbf{9.6}$ & $\mathbf{9.4}$ & $\mathbf{12.5}$ & $\mathbf{23.6}$ & $\mathbf{9.1}$ & $\mathbf{16.7}$ \\ \bottomrule
\end{tabular}%
}
\end{table}

%% file: tables/the_pile_350M-pm-client-baselines.tex
\begin{table}[ht]
\caption{ Validation perplexity ($\downarrow$) for $24$-block models trained on \texttt{The Pile} after \textbf{continued pre-training} with \textbf{proportional} sampling from \textbf{pre-trained} embeddings, compared to models which had been pre-trained on a single data source for the same total number of tokens as \dept has seen from their distributions. \dept significantly outperforms in terms of average perplexity but gets beaten by specialized models on their respective data source. Baselines whose pre-training dataset matches the evaluation dataset are highlighted in olive.}\label{tab:the_pile_350M_pm_cent_baselines} 
\centering
\resizebox{\textwidth}{!}{%
\begin{tabular}{@{}lcccccccccccccccccc@{}}
\toprule
\textbf{\begin{tabular}[c]{@{}l@{}}Name\\ ({\scriptsize UNIGRAM-CE})\end{tabular}} & \textbf{\begin{tabular}[c]{@{}c@{}}DM\\ ($6.9$)\end{tabular}} & \textbf{\begin{tabular}[c]{@{}c@{}}EE\\ ($7.9$)\end{tabular}} & \textbf{\begin{tabular}[c]{@{}c@{}}EP\\ ($10$)\end{tabular}} & \textbf{\begin{tabular}[c]{@{}c@{}}FL\\ ($7.8$)\end{tabular}} & \textbf{\begin{tabular}[c]{@{}c@{}}GH\\ ($7.9$)\end{tabular}} & \textbf{\begin{tabular}[c]{@{}c@{}}CC\\ ($7.9$)\end{tabular}} & \textbf{\begin{tabular}[c]{@{}c@{}}PA\\ ($8.2$)\end{tabular}} & \textbf{\begin{tabular}[c]{@{}c@{}}SE\\ ($7.7$)\end{tabular}} & \textbf{\begin{tabular}[c]{@{}c@{}}PP\\ ($9.1$)\end{tabular}} & \begin{tabular}[c]{@{}c@{}}WK\\ ($8.2$)\end{tabular} & \textbf{\begin{tabular}[c]{@{}c@{}}AX\\ ($7.7$)\end{tabular}} & \textbf{\begin{tabular}[c]{@{}c@{}}UB\\ ($7.8$)\end{tabular}} & \textbf{\begin{tabular}[c]{@{}c@{}}PC\\ ($8$)\end{tabular}} & \textbf{\begin{tabular}[c]{@{}c@{}}NH\\ ($8.1$)\end{tabular}} & \textbf{\begin{tabular}[c]{@{}c@{}}GU\\ ($7.7$)\end{tabular}} & \textbf{\begin{tabular}[c]{@{}c@{}}HN \\ ($7.7$)\end{tabular}} & \textbf{\begin{tabular}[c]{@{}c@{}}UI-OOD\\ ($10$)\end{tabular}} & \textbf{\begin{tabular}[c]{@{}c@{}}AVG\\ ($8.1$)\end{tabular}} \\ \midrule
\textbf{\texttt{CC}} & $19.6$ & $57.6$ & $294.4$ & $35.1$ & $32.2$ & {\color{olive}$\mathbf{30.8}$} & $41.1$ & $47.9$ & $133.1$ & $35.8$ & $44.2$ & $26$ & $47$ & $43.7$ & $52.8$ & $38.2$ & $79.2$ & $62.3$ \\
\textbf{\texttt{PC}} & $4.8$ & $25.5$ & $37.5$ & $17.3$ & $5.5$ & $45.5$ & $\mathbf{17.7}$ & $14.8$ & $63$ & $30.7$ & $9.4$ & $17.6$ & {\color{olive}$\mathbf{10.1}$} & $23.2$ & $46.4$ & $39$ & $100.8$ & $29.9$ \\
\textbf{\texttt{AX}} & $4.8$ & $27$ & $38.5$ & $18.7$ & $5.5$ & $49.7$ & $25.9$ & $14.7$ & $64.8$ & $33.8$ & {\color{olive}$\mathbf{7.4}$} & $19.5$ & $13.8$ & $32.9$ & $49.5$ & $41.4$ & $110.3$ & $32.8$ \\
\textbf{\texttt{GH}} & $5$ & $31.1$ & $47.3$ & $24.4$ & {\color{olive}$\mathbf{3.9}$} & $62.6$ & $37.1$ & $13.3$ & $80.6$ & $44.5$ & $11.6$ & $26.4$ & $17.9$ & $47.6$ & $60.8$ & $47.1$ & $70.1$ & $37.1$ \\
\textbf{\texttt{FL}} & $4.9$ & $23.8$ & $49.6$ & {\color{olive}$\mathbf{10.1}$} & $5.9$ & $45$ & $27.3$ & $15.8$ & $73.4$ & $31.6$ & $10.6$ & $20.4$ & $14.7$ & $33.6$ & $42.5$ & $38.4$ & $99.5$ & $32.2$ \\
\textbf{\texttt{SE}} & $4.7$ & $24.3$ & $38.3$ & $17.9$ & $4.4$ & $45.7$ & $27.1$ & {\color{olive}$\mathbf{9.3}$} & $62.1$ & $33$ & $9.5$ & $19.8$ & $14.3$ & $34.3$ & $46.1$ & $34.7$ & $70.2$ & $29.2$ \\
\textbf{\texttt{WK}} & $4.9$ & $23.8$ & $33.8$ & $16.4$ & $5.7$ & $40$ & $25.1$ & $15.2$ & $57.4$ & {\color{olive}$\mathbf{18.6}$} & $10.2$ & $19.4$ & $13.9$ & $31.3$ & $39.5$ & $37.4$ & $98.7$ & $28.9$ \\
\textbf{\texttt{DM}} & {\color{olive} $\mathbf{4.4}$} & $81.6$ & $210.9$ & $59.3$ & $14.4$ & $143.8$ & $99.7$ & $41.2$ & $248.4$ & $116.5$ & $22.6$ & $67.9$ & $40.8$ & $124.1$ & $126.1$ & $108.1$ & $424.8$ & $113.8$ \\ \midrule
\textbf{\glob} & $4.5$ & $\mathbf{17}$ & $\mathbf{16.1}$ & $13.2$ & $4.5$ & $34.5$ & $17.9$ & $11.2$ & $\mathbf{37.8}$ & $22.4$ & $8.4$ & $\mathbf{14.4}$ & $11$ & $\mathbf{20.6}$ & $\mathbf{35.5}$ & $\mathbf{28.3}$ & $61.2$ & $\mathbf{21.1}$ \\
\textbf{\trim} & $4.6$ & $20.5$ & $23$ & $13.9$ & $4.6$ & $38$ & $20.2$ & $12$ & $49.9$ & $25.1$ & $8.7$ & $16.6$ & $11.8$ & $25.7$ & $38$ & $32.9$ & $\mathbf{56.8}$ & $23.7$ \\ \midrule
\textbf{Min Imp (\%)} & $-2.9$ & $\mathbf{13.9}$ & $\mathbf{32}$ & $-37.4$ & $-19.8$ & $-23.5$ & $-14.1$ & $-28.2$ & $\mathbf{13.1}$ & $-34.8$ & $-18.7$ & $\mathbf{5.5}$ & $-16.9$ & $-10.9$ & $\mathbf{3.8}$ & $\mathbf{5.3}$ & $\mathbf{12.7}$ & $\mathbf{18.1}$ \\
\textbf{Max Imp (\%)} & $-1$ & $\mathbf{28.5}$ & $\mathbf{52.3}$ & $-31.1$ & $-16.6$ & $-12$ & $-1.1$ & $-19.8$ & $\mathbf{34.2}$ & $-20.4$ & $-14.3$ & $\mathbf{18.1}$ & $-9.3$ & $\mathbf{11.2}$ & $\mathbf{10.1}$ & $\mathbf{18.5}$ & $\mathbf{19}$ & $\mathbf{27}$ \\ \bottomrule
\end{tabular}%
}
\end{table}

%% file: tables/mc4_125M-m-cent-baselines.tex
\begin{table}[ht]
    \caption{Validation perplexity ($\downarrow$) for $12$-block models trained on \texttt{MC4} after \textbf{continued pre-training} with \textbf{uniform} sampling from \textbf{pre-trained} embeddings, compared to models which had been pre-trained on a single data source for the same total number of tokens as \dept has seen from their distributions. \dept significantly outperforms in terms of average perplexity but gets beaten by specialized models on their respective data source. Baselines whose pre-training dataset matches the evaluation dataset are highlighted in olive.} \label{tab:mc4_125M_m_cent_baselines} 
\resizebox{\textwidth}{!}{%
\begin{tabular}{@{}lccccccccccccc@{}}
\toprule
 & \multicolumn{9}{c}{\textbf{In-Distribution}} & \multicolumn{4}{c}{\textbf{Out-of-Distribution}} \\ \cmidrule(l){2-14} 
\textbf{\begin{tabular}[c]{@{}l@{}}Name\\ ({\scriptsize UNIGRAM-CE})\end{tabular}} & \textbf{\begin{tabular}[c]{@{}c@{}}ZH\\ ($9.8$)\end{tabular}} & \textbf{\begin{tabular}[c]{@{}c@{}}UR\\ ($10.5$)\end{tabular}} & \textbf{\begin{tabular}[c]{@{}c@{}}MS\\ ($9.2$)\end{tabular}} & \textbf{\begin{tabular}[c]{@{}c@{}}IT\\ ($7.7$)\end{tabular}} & \textbf{\begin{tabular}[c]{@{}c@{}}SR\\ ($10.5$)\end{tabular}} & \textbf{\begin{tabular}[c]{@{}c@{}}LA\\ ($9$)\end{tabular}} & \textbf{\begin{tabular}[c]{@{}c@{}}EN\\ ($7.5$)\end{tabular}} & \textbf{\begin{tabular}[c]{@{}c@{}}SW\\ ($10$)\end{tabular}} & \textbf{\begin{tabular}[c]{@{}c@{}}Avg (In-D)\\ ($9.3$)\end{tabular}} & \textbf{\begin{tabular}[c]{@{}c@{}}EL\\ ($14.4$)\end{tabular}} & \textbf{\begin{tabular}[c]{@{}c@{}}HI\\ ($13.9$)\end{tabular}} & \textbf{\begin{tabular}[c]{@{}c@{}}DE\\ ($9.7$)\end{tabular}} & \textbf{\begin{tabular}[c]{@{}c@{}}Avg (OOD)\\ ($12.6$)\end{tabular}} \\ \midrule
\textbf{\texttt{ZH}} & {\color{olive} $33.3$} & $36.6$ & $87.1$ & $71.3$ & $68.7$ & $55.9$ & $90.2$ & $58.1$ & $62.6$ & 5351.5127 & 3197.97314 & 936.293457 & 3161.92643 \\
\textbf{\texttt{UR}} & $124.7$ & {\color{olive}$12.7$} & $70.8$ & $62.9$ & $57.4$ & $49.5$ & $76.8$ & $49.8$ & $63.1$ & 3189.1106 & 774.645325 & 802.290588 & 1588.68217 \\
\textbf{\texttt{MS}} & $89.5$ & $26$ & {\color{olive}$27.8$} & $48.1$ & $47.1$ & $38.2$ & $58.9$ & $38.3$ & $46.7$ & 2983.83643 & 2491.64209 & 620.154907 & 2031.87781 \\
\textbf{\texttt{IT}} & $91.4$ & $27.4$ & $59.6$ & {\color{olive}$24.4$} & $46.9$ & $35.8$ & $59.6$ & $40.7$ & $48.2$ & 2353.07642 & 3057.45215 & 344.762726 & 1918.43043 \\
\textbf{\texttt{SR}} & $99.3$ & $28.1$ & $65.1$ & $52.5$ & {\color{olive}$20.2$} & $41.7$ & $70.1$ & $43.6$ & $52.6$ & 2174.83203 & 3398.59985 & 643.610352 & 2072.34741 \\
\textbf{\texttt{LA}} & $100.2$ & $30.2$ & $66.9$ & $48.1$ & $51$ & {\color{olive}$20.1$} & $68.1$ & $45.1$ & $53.7$ & 716.628967 & 2479.87817 & 240.702454 & 1145.73653 \\
\textbf{\texttt{EN}} & $1142.1$ & $150.7$ & $276.5$ & $211.1$ & $408.8$ & $147.1$ & {\color{olive}$86.2$} & $169.1$ & $323.9$ & 1315877.75 & 679658.688 & 11445.9727 & 668994.137 \\
\textbf{\texttt{SW}} & $91.6$ & $26.2$ & $55.9$ & $48.3$ & $47$ & $38.1$ & $58.3$ & {\color{olive}$17.8$} & $47.9$ & 2782.6792 & 2673.07813 & 557.471863 & 2004.40973 \\ \midrule
\textbf{\glob} & $\mathbf{40.1}$ & $\mathbf{15.5}$ & $\mathbf{30.1}$ & $39.6$ & $39$ & $29.7$ & $40.5$ & $\mathbf{24.6}$ & $\mathbf{32.4}$ & $1737.3$ & $\mathbf{823.4}$ & $335.1$ & $965.3$ \\
\textbf{\trim} & $41.9$ & $16.2$ & $31.3$ & $41.3$ & $40.8$ & $30.8$ & $42$ & $25.6$ & $33.7$ & $1725$ & $855.2$ & $345.6$ & $975.3$ \\ \midrule
\textbf{Min Imp (\%)} & $-26.1$ & $-28.1$ & $-12.8$ & $-68.9$ & $-101.9$ & $-53$ & $\mathbf{28}$ & $-43.9$ & $\mathbf{27.8}$ & $-142.4$ & $-10.4$ & $-43.6$ & $\mathbf{14.9}$ \\
\textbf{Max Imp (\%)} & $-20.6$ & $-22.8$ & $-8.5$ & $-62.1$ & $-92.7$ & $-47.4$ & $\mathbf{30.6}$ & $-38.6$ & $\mathbf{30.7}$ & $-142.4$ & $-10.4$ & $-43.6$ & $\mathbf{14.9}$ \\ \bottomrule
\end{tabular}%
}
\vspace{-0.4cm}
\end{table}

%% file: tables/the_pile_125M-m_pythia-baseline.tex
\begin{table}[t]
\caption{Validation perplexity ($\downarrow$)  for our $12$-block models trained on \texttt{The Pile} when performing continued pre-training using \textbf{uniform} sampling starting from a \textbf{pre-trained} embedding matrix. \dept slightly outperforms \texttt{Pythia-160M} at this small scale as its $30\times$ greater number of tokens is not beneficial with insufficient model capacity. \texttt{Pythia-160M} was trained on \texttt{Ubuntu IRC~(UI)}, thus its outperformance is expected as it is not an OOD dataset for this model.}\label{tab:the_pile_125M_m_pythia}
\resizebox{\textwidth}{!}{%
\begin{tabular}{@{}lcccccccccccccccccc@{}}
\toprule
\textbf{\begin{tabular}[c]{@{}l@{}}Name\\ ({\scriptsize UNIGRAM-CE})\end{tabular}} & \textbf{\begin{tabular}[c]{@{}c@{}}NH\\ ($8.1$)\end{tabular}} & \textbf{\begin{tabular}[c]{@{}c@{}}GH\\ ($7.9$)\end{tabular}} & \textbf{\begin{tabular}[c]{@{}c@{}}PA\\ ($8.2$)\end{tabular}} & \textbf{\begin{tabular}[c]{@{}c@{}}UB\\ ($7.8$)\end{tabular}} & \textbf{\begin{tabular}[c]{@{}c@{}}FL\\ ($7.8$)\end{tabular}} & \textbf{\begin{tabular}[c]{@{}c@{}}EE\\ ($7.9$)\end{tabular}} & \textbf{\begin{tabular}[c]{@{}c@{}}EP\\ ($10$)\end{tabular}} & \textbf{\begin{tabular}[c]{@{}c@{}}WK\\ ($8.2$)\end{tabular}} & \textbf{\begin{tabular}[c]{@{}c@{}}CC\\ ($7.9$)\end{tabular}}  & \textbf{\begin{tabular}[c]{@{}c@{}}SE\\ ($7.7$)\end{tabular}} & \textbf{\begin{tabular}[c]{@{}c@{}}PC\\ ($8$)\end{tabular}} & \textbf{\begin{tabular}[c]{@{}c@{}}PP\\ ($9.1$)\end{tabular}} & \textbf{\begin{tabular}[c]{@{}c@{}}DM\\ ($6.9$)\end{tabular}} & \textbf{\begin{tabular}[c]{@{}c@{}}AX\\ ($7.7$)\end{tabular}} & \textbf{\begin{tabular}[c]{@{}c@{}}GU\\ ($7.7$)\end{tabular}} & \textbf{\begin{tabular}[c]{@{}c@{}}HN \\ ($7.7$) \end{tabular}} & \textbf{\begin{tabular}[c]{@{}c@{}}UI-OOD\\ ($10$)\end{tabular}} & \textbf{\begin{tabular}[c]{@{}c@{}}AVG\\ ($8.1$)\end{tabular}} \\ \midrule
\textbf{\texttt{PYTHIA-160M}} & $47.3$ & $8.2$ & $36.8$ & $31.4$ & $24.1$ & $34$ & $32.8$ & $\mathbf{40.2}$ & $64.1$ & $22.4$ & $21.35$ & $74.5$ & $6.8$ & $16.3$ & $55$ & $54.9$ & $\mathbf{24.31}$ & $33.1$ \\ \midrule
\textbf{\glob} & $30.2$ & $7.1$ & $29.8$ & $22.9$ & $23.7$ & $20.0$ & $21.9$ & $41.6$ & $61.8$ & $17.9$ & $19.6$ & $48.0$ & $5.2$ & $13.7$ & $54.1$ & $42.2$ & $90.7$ & $32.4$ \\
\textbf{\trim} & $\mathbf{29.5}$ & $\mathbf{6.9}$ & $\mathbf{29.2}$ & $\mathbf{22.4}$ & $\mathbf{23.0}$ & $\mathbf{19.4}$ & $\mathbf{21.1}$ & $40.8$ & $\mathbf{60.6}$ & $\mathbf{17.4}$ & $\mathbf{19.2}$ & $\mathbf{46.7}$ & $\mathbf{5.1}$ & $\mathbf{13.4}$ & $\mathbf{52.4}$ & $\mathbf{41.0}$ & $81.1$ & $\mathbf{31.1}$ \\ \bottomrule
\end{tabular}%
}
\end{table}

%% file: tables/the_pile_350M-pr-pythia-baselines.tex
\begin{table}[ht]
\caption{ Validation perplexity ($\downarrow$) for $24$-block models trained on \texttt{The Pile} after \textbf{continued pre-training} with \textbf{proportional} sampling from \textbf{randomly-initialized} embeddings, compared to \texttt{Pythia-410M}. \dept models come close to \texttt{Pythia-410M} despite  the latter being trained on $10 \times$ more tokens, indicating a comparable if slightly worse transformer body. \texttt{Pythia-410M} was trained on \texttt{Ubuntu IRC~(UI)}, thus its outperformance is expected as it is not an OOD dataset for this model.}\label{tab:the_pile_350M_pr_pythia_baseline} 
\centering
\resizebox{\textwidth}{!}{%
\begin{tabular}{lcccccccccccccccccc}
\toprule
\textbf{\begin{tabular}[c]{@{}l@{}}Name\\ ({\scriptsize UNIGRAM-CE})\end{tabular}} & \textbf{\begin{tabular}[c]{@{}c@{}}DM\\ ($6.9$)\end{tabular}} & \textbf{\begin{tabular}[c]{@{}c@{}}EE\\ ($7.9$)\end{tabular}} & \textbf{\begin{tabular}[c]{@{}c@{}}EP\\ ($10$)\end{tabular}} & \textbf{\begin{tabular}[c]{@{}c@{}}FL\\ ($7.8$)\end{tabular}} & \textbf{\begin{tabular}[c]{@{}c@{}}GH\\ ($7.9$)\end{tabular}} & \textbf{\begin{tabular}[c]{@{}c@{}}CC\\ ($7.9$)\end{tabular}} & \textbf{\begin{tabular}[c]{@{}c@{}}PA\\ ($8.2$)\end{tabular}} & \textbf{\begin{tabular}[c]{@{}c@{}}SE\\ ($7.7$)\end{tabular}} & \textbf{\begin{tabular}[c]{@{}c@{}}PP\\ ($9.1$)\end{tabular}} & \begin{tabular}[c]{@{}c@{}}WK\\ ($8.2$)\end{tabular} & \textbf{\begin{tabular}[c]{@{}c@{}}AX\\ ($7.7$)\end{tabular}} & \textbf{\begin{tabular}[c]{@{}c@{}}UB\\ ($7.8$)\end{tabular}} & \textbf{\begin{tabular}[c]{@{}c@{}}PC\\ ($8$)\end{tabular}} & \textbf{\begin{tabular}[c]{@{}c@{}}NH\\ ($8.1$)\end{tabular}} & \textbf{\begin{tabular}[c]{@{}c@{}}GU\\ ($7.7$)\end{tabular}} & \textbf{\begin{tabular}[c]{@{}c@{}}HN \\ ($7.7$)\end{tabular}} & \textbf{\begin{tabular}[c]{@{}c@{}}UI-OOD\\ ($10$)\end{tabular}} & \textbf{\begin{tabular}[c]{@{}c@{}}AVG\\ ($8.1$)\end{tabular}} \\ \midrule
\textbf{\texttt{PYTHIA-410M}} & \multicolumn{1}{l}{$4.9$} & \multicolumn{1}{l}{$25.9$} & \multicolumn{1}{l}{$43.3$} & \multicolumn{1}{l}{$17.4$} & \multicolumn{1}{l}{$\mathbf{5.1}$} & \multicolumn{1}{l}{$\mathbf{45.6}$} & \multicolumn{1}{l}{$\mathbf{24.7}$} & \multicolumn{1}{l}{$\mathbf{13.9}$} & \multicolumn{1}{l}{$\mathbf{65.7}$} & \multicolumn{1}{l}{$\mathbf{31.7}$} & \multicolumn{1}{l}{$\mathbf{9.6}$} & \multicolumn{1}{l}{$\mathbf{18.8}$} & \multicolumn{1}{l}{$\mathbf{13.5}$} & \multicolumn{1}{l}{$\mathbf{31.3}$} & \multicolumn{1}{l}{$\mathbf{44.5}$} & \multicolumn{1}{l}{$\mathbf{38.3}$} & \multicolumn{1}{l}{$\mathbf{81.2}$} & \multicolumn{1}{l}{$\mathbf{30.3}$} \\ \midrule
\textbf{\glob} & $4.8$ & $\mathbf{25.7}$ & $38.2$ & $\mathbf{17.3}$ & $5.4$ & $47.7$ & $25.7$ & $14.7$ & $68.3$ & $32.7$ & $9.9$ & $20$ & $14$ & $32.2$ & $46.5$ & $39.8$ & $94.8$ & $31.6$ \\
\textbf{\trim} & $4.8$ & $27.3$ & $39.5$ & $18.5$ & $5.6$ & $51.2$ & $27.8$ & $15.4$ & $71.8$ & $35.1$ & $10.3$ & $21.7$ & $14.8$ & $35.1$ & $49.1$ & $42.2$ & $95.7$ & $33.3$ \\
\textbf{\spec} & $4.8$ & $26.7$ & $36.8$ & $18.2$ & $5.5$ & $50.1$ & $27.1$ & $15.1$ & $69.1$ & $34.2$ & $10.1$ & $21.1$ & $14.5$ & $34.3$ & $48.5$ & $41.7$ & $97.6$ & $32.7$ \\
\textbf{\texttt{SPEC-OPT}} & $\mathbf{4.7}$ & $25.9$ & $\mathbf{35}$ & $17.5$ & $5.4$ & $48.3$ & $26.1$ & $14.7$ & $66.6$ & $32.8$ & $9.9$ & $20.4$ & $14.1$ & $32.9$ & $47.3$ & $40.5$ & $88.6$ & $31.2$ \\ \bottomrule
\end{tabular}%
}
\end{table}

%% file: tables/the_pile_350M-pm-pythia-baseline.tex
\begin{table}[ht]
\caption{ Validation perplexity ($\downarrow$) for $24$-block models trained on \texttt{The Pile} after \textbf{continued pre-training} with \textbf{proportional} sampling from \textbf{randomly-initialized} embeddings, compared to \texttt{Pythia-410M}. \texttt{Pythia-410M} significantly outperforms \dept as its $30\times$ larger number of training tokens allow it to train much better embeddings. \texttt{Pythia-410M} was trained on \texttt{Ubuntu IRC~(UI)}, thus its outperformance is expected as it is not an OOD dataset for this model.}\label{tab:the_pile_350M_pm_pythia_baseline} 
\centering
\resizebox{\textwidth}{!}{%
\begin{tabular}{lcccccccccccccccccc}
\toprule
\textbf{\begin{tabular}[c]{@{}l@{}}Name\\ ({\scriptsize UNIGRAM-CE})\end{tabular}} & \textbf{\begin{tabular}[c]{@{}c@{}}DM\\ ($6.9$)\end{tabular}} & \textbf{\begin{tabular}[c]{@{}c@{}}EE\\ ($7.9$)\end{tabular}} & \textbf{\begin{tabular}[c]{@{}c@{}}EP\\ ($10$)\end{tabular}} & \textbf{\begin{tabular}[c]{@{}c@{}}FL\\ ($7.8$)\end{tabular}} & \textbf{\begin{tabular}[c]{@{}c@{}}GH\\ ($7.9$)\end{tabular}} & \textbf{\begin{tabular}[c]{@{}c@{}}CC\\ ($7.9$)\end{tabular}} & \textbf{\begin{tabular}[c]{@{}c@{}}PA\\ ($8.2$)\end{tabular}} & \textbf{\begin{tabular}[c]{@{}c@{}}SE\\ ($7.7$)\end{tabular}} & \textbf{\begin{tabular}[c]{@{}c@{}}PP\\ ($9.1$)\end{tabular}} & \begin{tabular}[c]{@{}c@{}}WK\\ ($8.2$)\end{tabular} & \textbf{\begin{tabular}[c]{@{}c@{}}AX\\ ($7.7$)\end{tabular}} & \textbf{\begin{tabular}[c]{@{}c@{}}UB\\ ($7.8$)\end{tabular}} & \textbf{\begin{tabular}[c]{@{}c@{}}PC\\ ($8$)\end{tabular}} & \textbf{\begin{tabular}[c]{@{}c@{}}NH\\ ($8.1$)\end{tabular}} & \textbf{\begin{tabular}[c]{@{}c@{}}GU\\ ($7.7$)\end{tabular}} & \textbf{\begin{tabular}[c]{@{}c@{}}HN \\ ($7.7$)\end{tabular}} & \textbf{\begin{tabular}[c]{@{}c@{}}UI-OOD\\ ($10$)\end{tabular}} & \textbf{\begin{tabular}[c]{@{}c@{}}AVG\\ ($8.1$)\end{tabular}} \\ \midrule
\textbf{\texttt{PYTHIA-410M}} & $\mathbf{3.8}$ & $\mathbf{9.7}$ & $\mathbf{8.9}$ & $\mathbf{7.7}$ & $\mathbf{3}$ & $\mathbf{19}$ & $\mathbf{11.8}$ & $\mathbf{7.2}$ & $\mathbf{21.3}$ & $\mathbf{12.5}$ & $\mathbf{5.9}$ & $\mathbf{10.3}$ & $\mathbf{7.6}$ & $\mathbf{15}$ & $\mathbf{17.6}$ & $\mathbf{16.9}$ & $\mathbf{7.8}$ & $\mathbf{10.9}$ \\ \midrule
\textbf{\glob} & $4.5$ & $17$ & $16.1$ & $13.2$ & $4.5$ & $34.5$ & $17.9$ & $11.2$ & $37.8$ & $22.4$ & $8.4$ & $14.4$ & $11$ & $20.6$ & $35.5$ & $28.3$ & $61.2$ & $21.1$ \\
\textbf{\trim} & $4.6$ & $20.5$ & $23$ & $13.9$ & $4.6$ & $38$ & $20.2$ & $12$ & $49.9$ & $25.1$ & $8.7$ & $16.6$ & $11.8$ & $25.7$ & $38$ & $32.9$ & $56.8$ & $23.7$ \\ \bottomrule

\end{tabular}%
}
\end{table}

%% file: appendix/applications.tex
\section{Applications}

% Remember to discuss positional embedding. Potential new section.
% 

\subsection{Federated Pre-Training of LLMs on Multilingual Population}
The challenges of training under data heterogeneity have come back into focus with recent forays into federated pre-training~\citep{DiLoCo,LLMFL,DatasetGrouper,Distro}, triggered in equal parts by privacy concerns, compute sharing and the search for more data in previously untapped reservoirs.

The way in which datasets are curated, filtered, and combined has a significant impact on the performance of LLMs \citep{long2024llms}. Determining the best methods for data curation, filtering, and mixing from various sources requires extensive experimentation to identify configurations that optimize performance on target evaluation metrics \citep{meta2024introducing}. Consequently, the specific details of these processes are often closely guarded by leading LLM developers. Despite careful dataset preparation, data heterogeneity remains inevitable due to the inherent imbalance in data sources. One of the most prominent imbalances is in language representation. For instance, only about 5\% of the pre-training data for Llama3 is non-English, covering over 30 languages, which results in lower expected performance in non-English contexts \citep{meta2024introducing}. A similar performance disparity across languages has also been observed with GPT-4 \citep{achiam2023gpt}.

Current datasets used for pre-training are highly geographically concentrated to a few areas of the globe~\citep{DatasetGeography}, providing the so-called high-resource languages, with high-quality domain-specific data being available predominantly in such languages~\citep{LowResourceLanguagesSurvey}. Such datasets are collected from internet sources and then curated~\citep{gpt3,llama3}. However, bottlenecks in the rate of high-quality data generation~\citep{WillWeRunOutOfData} and copyright concerns~\citep{NytVsOpenAi} have led to large organizations making deals with private data providers such as publishers~\citep{OpenAISpringerDeal,OpenAINewsPublishersDeal} in order to meet the demand of ever-growing models.

Federated pre-training as a methodology allows the model to be taken directly to the training data, potentially enabling training under privacy concerns or legislation that limits data movement~\citep{positionpaper}. While this has obvious applications for collaborative training of LMs, it can also be applied by a single organization as a drop-in replacement for mini-batch SGD during pre-training~\citep{DiLoCo}, which eliminates dataset movement while massively lowering the communication frequency of model training compared to Data-parallel algorithms~\citep{FSDP_ZeRO} which need to synchronize gradients every batch. The version of the algorithm used in a centralized setting, mathematically equivalent to Federated Averaging~\citep{fedavg}, has alternatively been known as: (a) communication-efficient SGD~\citep{LinearSpeedupSGD}, (b) Local SGD~\citep{LocalSGD,LocalSGD_Trade_Offs_At_Scale},  or (c) as a specific variant of the REPTILE~\citep{REPTILE} meta-learning algorithm. Under these various methodologies, it has been shown to (a) confer a linear speedup to convergence similar to increasing batch size, (b) provide better generalization to models compared to standard large-batch training, (c) enable meta-learning across various tasks.

While the current centralized pre-training recipe may be stabilized with great effort, such measures are largely impractical in federated training scenarios where the participants refuse to offer full control over their data to a third party or where the underlying training distribution may shift as new participants enter a federation or old ones exit. Furthermore, the complexity of the current pipeline is impractical to all except the best-funded organizations, even in a centralized training context.

The inability to directly inspect data sources in a federated context makes it impossible to construct a dedicated vocabulary for a data mixture, ensure a standard curation pipeline on a per-sample basis, or strongly control data sampling rates across all sources. Motivated by this extreme setting, we aim to construct a pre-training procedure that is capable of learning from multiple highly heterogeneous data sources without model divergence while providing a foundation model with greater \textbf{generalization} and more \textbf{plasticity} in adapting to new data.

% \subsection{Distributed Pre-Training of LLMs on Multilingual Data Silos}

%% file: appendix/narrative_sections.tex
\section{Training Under Data Heterogeneity}\label{app:sec:training_under_data_heterogeneity}

Training LLMs such as Llama 3~\citep{llama3} requires extensive manual tuning, heuristics, and model-based data selection procedures. This effort aims to achieve the desired mix of categories, such as general knowledge, mathematics, coding, and multilingual data.

This complexity arises due to the wide range of capabilities required by LMs and the risk of negative interference across domains and languages. Current pre-training methodologies are prone to divergence unless data sampling ratios can be meticulously curated based on the characteristics of the data and its fit to the model's distribution at any given time~\citep{llama3}. Multi-domain ratios are manually curated for downstream performance, requiring extensive and expensive tuning,  while multilingual pre-training often employs temperature-weighted sampling~\citep{BERT,CurseOfMultilingualityUnsupervisedCrossLingual,mC4} due to the vast number of languages involved, 

As illustrated in \cref{fig:motivation:pre_train_Activations}, pre-training on heterogeneous data can result in model activation divergence~\citep{TrainingComputeOptimalLLMs}, even with a sampling temperature of $\tau=1.0$, which corresponds to proportional sampling based on dataset size. Activation divergence is a precursor to significant, often irrecoverable, increases in loss, and necessitating model re-starts from earlier checkpoints with lower learning rates~\citep{meta_opt}. Longer training durations could be achieved by disproportionately sampling from larger, lower-quality datasets like \texttt{C4} or high-resource languages like English in multilingual pre-training. Alternatively, methods like active forgetting via embedding resetting \citep{ActiveForgetting}, \texttt{ACT}, may artificially extend the training duration past the natural divergence point.

Previous studies show that this \textit{Curse of Multilinguality} and/or \textit{Negative Interference} can be attributed to vocabulary dilution and capacity contention~\citep{CurseOfMultilingualityUnsupervisedCrossLingual}, language-specific parameter emergence~\citep{NegativeInterferenceMetaLearning}, and suboptimal tokenization~\citep{HowGoodIsYourTokenizer}. Increasing model and vocabulary size helps capacity contention~\citep{CurseOfMultilingualityUnsupervisedCrossLingual,NegativeInterferenceMetaLearning}, but this requires immense hardware resources~\citep{llama3} to shard the model across multiple GPUS. Addressing vocabulary dilution in highly multilingual models is even more challenging, as providing enough tokens for all languages would result in impractically large models~\citep{HowGoodIsYourTokenizer}. These limitations drive us to find scalable methods to incorporate broader data mixtures without significantly increasing the in-memory model size during training.

\section{Fine-tuning \dept Models}\label{app:fine_tuning}
We evaluate fine-tuning performance on three downstream tasks: RACE, MNLI, and STSB. All models are fine-tuned using the recipes provided by \citet{gpt-1} for each task using the AdamW optimizer with a linear learning rate scheduler. For RACE, the model is trained for 5 epochs with a learning rate of 6e-5 and a batch size of 16. For MNLI, fine-tuning is performed over 2 epochs with a learning rate of 4e-5 and a batch size of 32. Finally, STSB is fine-tuned for 5 epochs using a learning rate of 2e-5 and a batch size of 32. The results are reported in \cref{tab:downstreamfull}.

\begin{table}[h]
\caption{The performance on downstream tasks ($\boldsymbol{\uparrow}$), following continued pre-training, shows that \dept models achieve $3\%-7.5\%$ relative improvements over the baselines, with \trim delivering the best results. \dept consistently outperforms baselines, even with pre-trained embedding initialization, underscoring the importance of an effective transformer body.} 
\label{tab:downstreamfull}
\centering
\resizebox{0.75\textwidth}{!}{
\begin{tabular}{@{}lccccccccc@{}}
\toprule
 & \multicolumn{4}{c}{\textbf{Random Init}} & \multicolumn{4}{c}{\textbf{Pre-trained Init}} \\ \cmidrule(r){2-5} \cmidrule(l){6-9}
\textbf{Name} & \textbf{\begin{tabular}[c]{@{}c@{}}RACE \\ (ACC)\end{tabular}} & \textbf{\begin{tabular}[c]{@{}c@{}}MNLI \\ (ACC)\end{tabular}} & \textbf{\begin{tabular}[c]{@{}c@{}}STSB \\ (PC)\end{tabular}} & \textbf{\begin{tabular}[c]{@{}c@{}}SST2 \\ (ACC)\end{tabular}} & \textbf{\begin{tabular}[c]{@{}c@{}}RACE \\ (ACC)\end{tabular}} & \textbf{\begin{tabular}[c]{@{}c@{}}MNLI \\ (ACC)\end{tabular}} & \textbf{\begin{tabular}[c]{@{}c@{}}STSB \\ (PC)\end{tabular}} & \textbf{\begin{tabular}[c]{@{}c@{}}SST2 \\ (ACC)\end{tabular}} \\ \cmidrule(r){1-5} \cmidrule(l){6-9}
\textbf{\texttt{STD}} ($\tau=0$) & $0.5$ & $0.6$ & $0.66$ & $0.79$ & $0.5$ & $0.71$ & $0.74$ & $0.81$ \\
\textbf{\texttt{STD}} ($\tau=1$) & $0.46$ & $0.68$ & $0.73$ & $0.81$ & $0.53$ & $0.7$ & $0.76$ & $0.83$ \\
\textbf{\act} & $0.45$ & $0.66$ & $0.73$ & $0.8$ & $-$ & $-$ & $-$ & $-$ \\ \cmidrule(r){1-5} \cmidrule(l){6-9}
\textbf{\glob} & $0.51$ & $\mathbf{0.72}$ & $0.78$ & $0.83$ & $0.51$ & $0.69$ & $0.76$ & $0.82$ \\
\textbf{\trim} & $\mathbf{0.53}$ & $0.71$ & $0.78$ & $0.83$ & $\mathbf{0.55}$ & $\mathbf{0.73}$ & $\mathbf{0.81}$ & $\mathbf{0.86}$ \\
\textbf{\spec} & $0.52$ & $0.71$ & $\mathbf{0.79}$ & $0.81$ & $-$ & $-$ & $-$ & $-$ \\
\textbf{\texttt{SPEC-OPT}} & $0.51$ & $0.69$ & $0.77$ & $\mathbf{0.85}$ & $-$ & $-$ & $-$ & $-$ \\ \cmidrule(r){1-5} \cmidrule(l){6-9}
\textbf{Min Imp (\%)} & $\mathbf{2.9\%}$ & $\mathbf{4.6\%}$ & $\mathbf{5.9\%}$ & $-0.7\%$ & $-3.7\%$ & $-3.2\%$ & $\mathbf{0.5\%}$ & $-1.8\%$ \\
\textbf{Max Imp (\%)} & $\mathbf{5.8\%}$ & $\mathbf{6.1\%}$ & $\mathbf{7.5\%}$ & $\mathbf{4.1\%}$ & $\mathbf{3.2\%}$ & $\mathbf{3\%}$ & $\mathbf{6.6\%}$ & $\mathbf{3.2\%}$ \\ \bottomrule
\end{tabular}%
}
\end{table}

\section{Using \spec Models for Inference}\label{app:extending_spec}

As discussed in \cref{subsec:variant_assumptions_and_benefits,subsec:limitations,subsec:continued_pre_training}, \spec models do not inherently support inference on a broad corpus after initial pre-training. Suppose local vocabularies and embedding matrices are available without privacy concerns. In that case, inference can be performed using the embedding matrix of the broadest data source or the one closest to the target application. For instance, targeting English text would utilize \texttt{EN} embeddings for \texttt{MC4} or \texttt{CC} embeddings for \texttt{The Pile}. While effective, this limits generalization beyond the broadest dataset in the pre-training distribution.

To handle a corpus resembling a mixture of all pre-training data sources or unseen ones, \spec models require a broader embedding matrix for good performance. This can be achieved through multi-phased adaptive or continued pre-training, starting with a random embedding matrix or the broadest pre-training one, as demonstrated in this work. Alternatives include vocabulary/embedding transfer~\citep{TransTokenization} or vocabulary matching~\citep{VocabMatching}. If these methods fail to reach the desired performance, additional optimization may be necessary to align the embeddings with the transformer body.